%% file: main.tex
\documentclass[twoside,11pt]{article}

\usepackage{blindtext}

\usepackage[abbrvbib]{jmlr2e}

\usepackage[utf8]{inputenc} 
\usepackage[T1]{fontenc}    
\usepackage{hyperref}       
\usepackage{url}            
\usepackage{booktabs}       
\usepackage{amsfonts}       
\usepackage{nicefrac}       
\usepackage{microtype}      
\usepackage{xcolor}         

\usepackage{graphicx}
\usepackage{pifont}
\usepackage{subcaption}
\usepackage{placeins}
\usepackage{multirow}
\usepackage{multicol}
\usepackage{amsmath} 
\usepackage{cleveref}
\usepackage{algorithm}
\usepackage{algorithmic}
\usepackage{xcolor}
\usepackage{makecell}
\usepackage{comment}
\usepackage{lineno}

\crefname{table}{Figure}{Figures}
\crefname{figure}{Figure}{Figures}
\crefname{appendix}{Appendix}{}
\crefname{section}{Section}{}

\makeatletter
\let\c@table\c@figure     
\makeatother

\usepackage{xcolor}

\usepackage{lastpage}
\jmlrheading{}{2026}{1-\pageref{LastPage}}{1/26}{}{}{Yihang She, Andrew Blake, Clement Atzberger, Adriano Gualandi, Srinivasan Keshav}

\usepackage{lastpage}
\jmlrheading{27}{2026}{1-\pageref{LastPage}}{12/25; Revised
8/26}{11/26}{25-3154}{Yihang She, Andrew Blake, Clement Atzberger, Adriano Gualandi, Srinivasan Keshav}

\ShortHeadings{PILA for Interpretable EO}{She, Blake, Atzberger, Gualandi and Keshav}
\firstpageno{1}

\begin{document}

\title{PILA: Physics-Informed Low Rank Augmentation for Interpretable Earth Observation}

\author{\name Yihang She \email ys611@cam.ac.uk \\
       \addr Department of Computer Science and Technology, University of Cambridge\\
       15 JJ Thomson Avenue, Cambridge CB3 0FD, United Kingdom
       \AND
       \name Andrew Blake \email ab@ablake.ai \\
       \addr Clare Hall, University of Cambridge\\
       Herschel Road, Cambridge CB3 9AL, United Kingdom
       \AND
       \name Clement Atzberger \email clement@cyclops.ai \\
       \addr Cyclops AI
       \AND
       \name Adriano Gualandi \email ag2347@cam.ac.uk \\
       \addr Department of Earth Sciences, University of Cambridge\\
       Bullard Laboratories, Madingley Road, Cambridge CB3 0EZ, United Kingdom
       \AND
       \name Srinivasan Keshav \email sk818@cam.ac.uk \\
       \addr Department of Computer Science and Technology, University of Cambridge\\
       15 JJ Thomson Avenue, Cambridge CB3 0FD, United Kingdom}

\editor{Kilian Weinberger}

\maketitle

\begin{abstract}
\input{0_abstract}

\end{abstract}


\input{1_introduction}

\input{2_related_work}
\input{3_methods}

\input{4_results}
\input{5_conclusion}


\acks{This work was supported by the UKRI Centre for Doctoral Training in Application of Artificial Intelligence to the study of Environmental Risks (reference EP/S022961/1) and Cambridge Centre for Carbon Credits. We would also like to thank Markus Immitzer from Cyclops AI for sharing the Sentinel-2 data with us.}

\vskip 0.2in
\bibliography{references}

\newpage
\appendix
\input{6_appendix}


\end{document}

%% file: 0_abstract.tex
Physically meaningful representations are essential for Earth Observation (EO), yet existing physical models are often simplified and incomplete, leading to discrepancies between simulation and observations that hinder reliable forward model inversion. 
Common approaches to EO inversion either ignored this incompleteness or relied on case-specific preprocessing. 
Recent methods use physics-informed autoencoders but depend on auxiliary variables that are difficult to interpret and multiple regularizers that are difficult to balance.
We propose Physics-Informed Low-Rank Augmentation (PILA), a framework that augments incomplete physical models using a learnable low-rank residual to improve flexibility, while remaining close to governing physics. 
We evaluate PILA on two EO inverse problems involving diverse physical processes: forest radiative transfer inversion from optical remote sensing; and volcanic deformation inversion from GNSS displacement data.
Across domains, PILA yields more accurate and interpretable physical variables. 
For forest spectral inversion, it improves species separation and, compared to ground measurements, reduces prediction errors by 40-71\% relative to the state-of-the-art. 
For volcanic deformation, PILA's recovery of variables captures a major inflation event at the Akutan volcano in 2008, and estimates source depth, volume change, and displacement patterns consistent with prior studies that required substantial preprocessing.
Finally, we analyse the effects of model rank, observability, and physical priors. 
Using synthetic data in a controlled setting, we further demonstrate the effectiveness of PILA's low-rank augmentation design, suggesting PILA may offer a general pathway for inverting incomplete physical models beyond Earth Observation.
The code is available at \url{https://github.com/yihshe/PILA.git}.

%% file: 1_introduction.tex
\section{Introduction}
Earth Observation (EO) can be viewed as obtaining observations $X$ derived from underlying physical variables $Z_{\mathrm{phy}}$ operated on by a physical process of the Earth system $\mathcal{F}^*$:
\begin{equation}\label{eq:forward_observation_ideal}
    X = \mathcal{F}^*(Z_{\mathrm{phy}}).
\end{equation}
Inverting $\mathcal{F}^*$ to derive $Z_{\mathrm{phy}}$ from $X$ is widely used for scientific discovery in different domains \citep{hao2022physics}, and is an important problem in EO applications ranging from land-surface monitoring \citep{lanconelli2025fifth} to geohazard assessment \citep{tarantola2005inverse}. 

The Earth system is intrinsically complex and noisy, so physical models $\mathcal{F}$ are often \textit{incomplete}, making this inversion challenging. Incompleteness may arise because $\mathcal{F}$ is too rigid and simplified to approximate $\mathcal{F}^*$, trading precision for computational efficiency \citep{atzberger2000development}, or because it focusses only on a subprocess of $\mathcal{F}^*$ and omits other processes that contribute to $X$ \citep{kiyoo1958relations}. Overlooking model incompleteness can lead to inaccurate retrieval of physical representations \citep{weiss2020s2toolbox,zerah2024physics}. 
Alternatively, handling $\mathcal{F}$'s incompleteness can require assumptions and case-specific preprocessing to separate modelled observation from raw observation, limiting generality \citep{ji2011transient,walwer2016data,shi2025fine}. 

\begin{figure}[htbp]
    \centering
    \includegraphics[width=\textwidth]{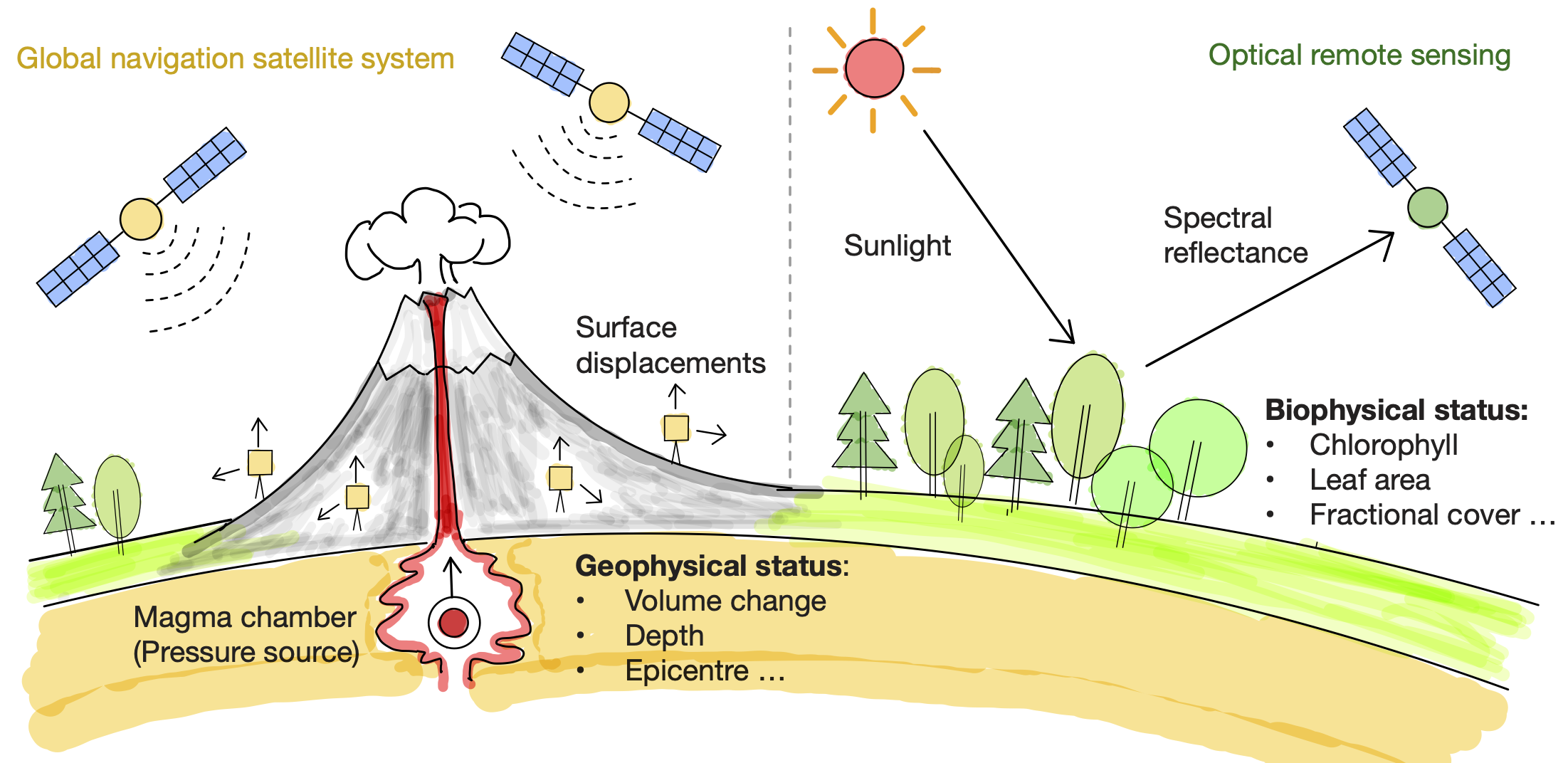}
    \caption{\textbf{We study inverse problems arising from Earth observations of disparate physical processes, to understand the planet, from the surface to the subsurface.}
    \textit{Study cases include: the inversion of a forest radiative transfer model---a planet renderer---to estimate biophysical status; and the inversion of a volcanic deformation model---representative of a broad family of geophysical inverse problems---to infer subsurface geophysical activity.}
    }
    \label{fig:intro_teaser_figure}
\end{figure}

In recent work on physics-informed machine learning \citep{kato2020differentiable,hao2022physics}, an auto-encoder architecture, with the decoder replaced by $\mathcal{F}$, is used to reconstruct observations. In this approach, the encoder learns an approximate inverse of $\mathcal{F}$ and the latent representations are disentangled to recover physical variables. 
Inverse graphics is a notable example, where a differentiable renderer serves as $\mathcal{F}$ and latent codes represent scene parameters \citep{loper2014opendr}. This view extends naturally to the radiative transfer process of EO, which acts as a ``planetary renderer'' across spectral bands, making its inversion similar to inverse graphics when $\mathcal{F}$ is made differentiable.
Another example is the inversion of dynamical systems, where data-driven auxiliary variables are learned to augment the incomplete $\mathcal{F}$ \citep{yin2021augmenting,takeishi2021physics}. These variables increase the flexibility of $\mathcal{F}$ to better reconstruct $X$. 
However, the minimized reconstruction error may mainly reflect the flexibility of the data-driven component and risk distorting physical causality; careful regularisation is needed to balance data-driven augmentation with reconstruction from physical variables.
Existing methods struggle to balance the regularisation terms representing different modelling assumptions \citep{yin2021augmenting, takeishi2021physics, takeishi2023deep}. They also tend to focus entirely on idealised systems or simple datasets, unlike EO settings where the systems are complex and the data is real.

To address these research gaps, inspired by the highly successful Low-Rank Adaptation (LoRA) technique \citep{hu2022lora} originally proposed for fine-tuning large language models (LLMs),
we refine an incomplete $\mathcal{F}$ by augmenting it with a \textit{low-rank residual}. LoRA reduces storage and compute costs for extremely dense neural network layers by adapting pretrained weights using an intrinsically low-rank matrix, in which the rank acts as a `tuning knob.' This yields a conceptually simple yet expressive mechanism for handling the incompleteness of $\mathcal{F}$ in inverse problems. 

We evaluate PILA on two representative physical processes in EO inverse problems (\cref{fig:intro_teaser_figure}): (1) the radiative transfer of the Earth's surface in optical remote sensing; and (2) the magmatic inflation or deflation driving the volcanic surface deformation observed by Global Navigation Satellite Systems (GNSS). 
For the former, we use a forest radiative transfer model (RTM) \citep{atzberger2000development} with Sentinel-2 observations. This RTM simplifies canopy structure using aggregated indicators such as leaf area and fractional coverage, trading accuracy for computational efficiency. It performs well on homogeneous surfaces like croplands but shows discrepancies between simulated and measured spectra  due to the complex structure of forests.
For the latter, we use a Mogi source \citep{kiyoo1958relations}, a deformation model widely used in volcano geodesy that computes surface displacements from a pressurized subsurface point source—though real magma chambers are typically more complex. It is a key example of geophysical inverse problems \citep{tarantola2005inverse}. The observed displacements may also contain tectonic trends and seasonal movements caused by hydrological loadings (\cref{appx:gnss_data_components}), so the observations contain unmodeled sources beyond the response of the simple model.

Comparisons with the state of the art approach Hybrid VAE (HVAE) \citep{takeishi2021physics} show that PILA recovers physical variables more accurately and with better disentanglement. Inverting the RTM of forest spectra over mixed-species Austrian forests, spanning both coniferous and broadleaf stands,  PILA learned variables that separate species better. Compared with ground measurements, PILA reduces the prediction errors by 40-71\% relative to the baseline. For GNSS time series at the Akutan volcano in the Aleutian arc, PILA successfully identifies a major volcanic inflation in early 2008 due to a Mogi source; the inferred volume change and displacement magnitudes agree with previous estimates that required numerical optimization on preprocessed GNSS data. Although we focus here on EO, PILA may also apply more broadly to inverse problems involving incomplete physical models.

In summary, our contributions are as follows. 
\begin{itemize}
    \item A novel physics-informed machine learning framework that augments physical models of varying incompleteness, using a data-driven low rank residual to invert them. 
    \item Applications to inverse problems arising from diverse physical processes in the Earth system, enabling interpretable EO insights into both the biophysical state of the surface and the geophysical processes beneath it. 
    \item Experiments showing that PILA enables more accurate, better-disentangled retrieval of physical variables. The experiments reveal that recovery of physical variables depends not only on the inversion algorithm, but also: how observable these variables are in the modelled physical process; the choice of physical priors; and how rank selection serves as a tuning knob for reconstruction accuracy and physical interpretability.
\end{itemize}

%% file: 2_related_work.tex
\section{Related Work}

\subsection{Physics-informed machine learning}
Physics-informed machine learning (PIML) tasks can be grouped according to how physics is used \citep{hao2022physics}: (1) \textbf{neural simulators}, which predict system states under physical constraints, and
(2) \textbf{inverse problems}, which infer physical parameters using a forward physical operator as prior.
Earth-surface radiative transfer belongs to the latter, and follows the physics of electromagnetic radiation. Its inversion closely parallels inverse graphics where differentiable rendering enables autoencoder-based inversion \citep{kato2020differentiable}.
A seminal work is OpenDR \citep{loper2014opendr}, which introduced this idea by making a graphics renderer fully differentiable, which now has enabled many vision and graphics applications \citep{pavlakos2018learning, kato2019learning, kato2020differentiable}.
Incomplete physical models are rarely addressed in inverse graphics, but are studied in PIML for dynamical systems (e.g. pendulums, diffusion, waves) \citep{hao2022physics}.
\citet{yin2021augmenting} proposed augmenting incomplete ODE/PDE dynamics with neural auxiliary variables. 
\citet{takeishi2021physics} extended this idea to introduce Hybrid Variational Auto-Encoders (HVAE) to invert incomplete dynamical systems and applied it to inverting a simple static operator of galaxy distributions. 
Follow-up work explores expert enhancement \citep{wehenkel2022robust} and distribution alignment \citep{singh2025hybrid}, but targets neural simulators rather than inverse problems.
A persistent challenge with incomplete physical priors is selecting regularisation and hyperparameters to balance competing training objectives \citep{takeishi2023deep}.

\subsection{Representation learning and low-rank structures}
Representation learning aims to discover compact and informative features from raw data. 
Our auto-encoder approach is closely linked to disentangled representation learning, which seeks semantically meaningful latent variables \citep{locatello2019challenging}. 
Variational Autoencoders (VAE) \citep{kingma2013auto,higgins2017beta} and GAN-based models such as InfoGAN \citep{chen2016infogan} and StyleGAN \citep{karras2019style} allow disentangled variables to emerge during learning. However, they lack clearly defined physical meanings. 
A parallel line of work focusses on self-supervised representation learning for downstream tasks \citep{gui2024survey}. In EO, this has led to numerous \textit{Earth foundation models} (see \citep{xiao2025foundation} for a list), which compress raw observations into rich feature spaces using large EO datasets and modern compute. However, these representations are usually intermediate and require fine-tuning with ground-truth data to gain physical meaning. 
For physical models, the situation is simplified because the forward model directly defines an inference mechanism with a fixed set of interpretable variables, namely the physical input parameters. 
Beyond compressing representations, imposing a low-rank gradient matrix enables controlled adaptation from the pretrained representation to that required for the downstream task. Low-Rank Adaptation (LoRA) \citep{hu2022lora} demonstrates this by updating pretrained model weights with a low-rank parameterisation, cutting storage and computation while preserving downstream performance.

\subsection{Inverse problems arising from Earth Observation}
Many inverse problems in EO arise from two distinct physical processes: (1) radiative transfer, which governs electromagnetic radiation in optical remote sensing and is simulated by RTMs \citep{lanconelli2025fifth}; and (2) geophysical processes that govern surface displacements observed by GNSS, modeled to study events such as earthquakes and volcanic eruptions \citep{tarantola2005inverse}. 
The forest RTM we use \citep{atzberger2000development} is a variant of PROSAIL RTM \citep{verhoef1984light, rosema1992new, feret2017prospect}, a 1-D RTM that approximates canopy structure statistically. While less accurate than recent 3-D RTMs \citep{qi2019less, Eradiate, cao2025physically}, it remains widely used for its computational efficiency \citep{ishaq2023systematic} and operational role in satellite biophysical products \citep{weiss2002validation, weiss2020s2toolbox}. 
Classical inversion approaches include Bayesian inference, numerical optimisation, lookup tables \citep{groetsch1993inverse,goel1988models, combal2003retrieval}, and neural network regressions \citep{gong1999inverting, schlerf2006inversion, weiss2020s2toolbox}; recent autoencoder-based inversion \citep{zerah2024physics} targets mainly for homogeneous croplands and assumes minimal model incompleteness.
The Mogi model \citep{kiyoo1958relations, segall2013volcano} represents a broader class of inverse problems in geophysics \citep{zhdanov2015inverse}. Geodetic position time series often combine multiple deformation mechanisms, so preprocessing methods such as trajectory modelling \citep{bevis2014trajectory}, Independent Component Analysis \citep{gualandi2016blind,ebmeier2016application,amoruso2024initial}, and Singular Spectrum Analysis \citep{walwer2016data} are commonly applied to separate their contributions. Recent supervised machine learning approaches bypass this preprocessing \citep{costantino2024denoising}, but require prior knowledge of the target signal. In contrast, PILA, as proposed here, performs inversion in a self-supervised, end-to-end setting.

%% file: 3_methods.tex
\section{Methods}
\label{sec:methods}

\subsection{Forward process}
Given the physical model $\mathcal{F}$, the forward process that produces a real observation $X$ can be written as follows:
\begin{equation}\label{eq:forward_observation}
    X = X_{\mathcal{F}} + \Delta + \epsilon,
\end{equation}
where $X_{\mathcal{F}}$ is the observation modelled by $\mathcal{F}$ from the physical variables $Z_{\mathrm{phy}}$:
\begin{equation}\label{eq:decode_phy}
    X_{\mathcal{F}} = \mathcal{F}(Z_{\mathrm{phy}}) .
\end{equation}
Variable $\Delta$ represents the discrepancy between $X_{\mathcal{F}}$ and $X$ resulting from unmodeled physical processes $\mathcal{C}$, which are controlled by auxiliary variables $Z_{\mathrm{aux}}$ and can be conditioned on the modelled observation $X_{\mathcal{F}}$:
\begin{equation}\label{eq:decode_aux}
    \Delta = \mathcal{C}(Z_{\mathrm{aux}})\mid X_{\mathcal{F}},
\end{equation}
and $\epsilon$ denotes random noise. 
Ideally, when $\mathcal{F}$ is perfectly constructed to account for all physical processes, $\Delta$ is not necessary in \cref{eq:forward_observation}. In that case, the retrieval of $Z_{\mathrm{phy}}$ depends primarily on the observability of $Z_{\mathrm{phy}}$ from $X$. 

\subsection{Learning the inverse}
The goal of the inverse problem is to infer physical variables $Z_{\mathrm{phy}}$ from raw observations $X$ given the physical model $\mathcal{F}$. Regardless of the degree of incompleteness of $\mathcal{F}$, ignoring discrepancies between $X$ and $X_{\mathcal{F}}$ can lead to implausible retrieval of $Z_{\mathrm{phy}}$ \citep{zerah2024physics}. Therefore, we seek estimates of $Z_{\mathrm{phy}}$ and a data-driven component $\Delta$ such that:
\begin{equation}\label{eq:forward_observation_approx}
    X \approx X_{\mathcal{F}} + \Delta.
\end{equation}

\begin{figure}[htbp]
    \centering
    \includegraphics[width=\textwidth]{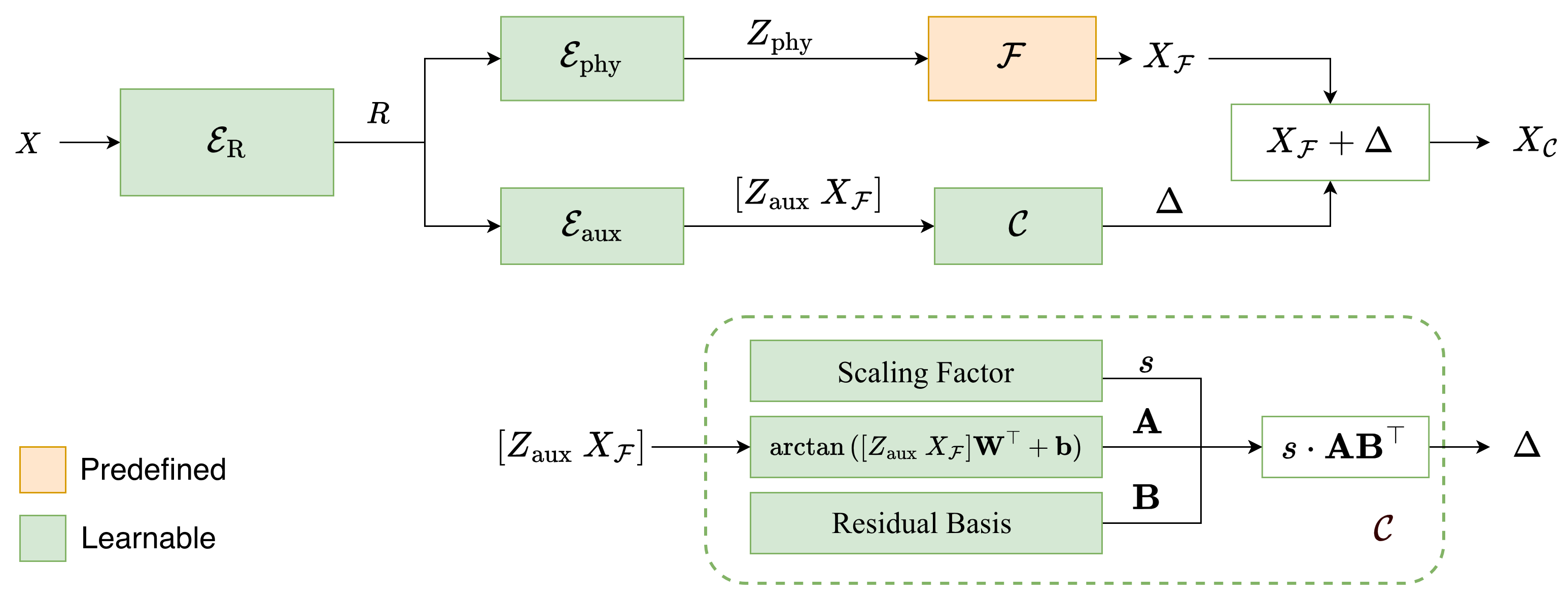}
    \caption{\textbf{PILA inverts physical models of varying incompleteness using a residual of intrinsically low rank.}
    \textit{Given an observation $X$, $\mathcal{E}_{\mathrm{R}}$ maps it to a high-dimensional feature $R$, which is then encoded by $\mathcal{E}_{\mathrm{phy}}$ and $\mathcal{E}_{\mathrm{aux}}$ into physical variables $Z_{\mathrm{phy}}$ and auxiliary variables $Z_{\mathrm{aux}}$. $Z_{\mathrm{phy}}$ is decoded by $\mathcal{F}$ to produce a physical reconstruction $X_{\mathcal{F}}$, which is refined by a low-rank residual $\Delta$ to yield the final reconstruction $X_{\mathcal{C}}$. The residual $\Delta$ is computed by a mapping $\mathcal{C}$ as the product of a scaling factor $s$, a coefficient matrix $\mathbf{A}\in\mathbb{R}^{n\times r}$, and a residual basis matrix $\mathbf{B}\in\mathbb{R}^{d\times r}$, with $\operatorname{rank}(\Delta)= r\ll d$. Here, $s$ and $\mathbf{B}$ are shared parameters applied to all samples, while $\mathbf{A}$ is obtained by linearly mapping the concatenation of $Z_{\mathrm{aux}}$ and the physical output $X_{\mathcal{F}}$, with a stop-gradient operation applied to $X_{\mathcal{F}}$ during backpropagation.}
    }
    \label{fig:methods_phylora}
\end{figure}

We use an auto-encoder architecture that includes a physical pathway for $X_{\mathcal{F}}$ and an auxiliary pathway for $\Delta$ (see \cref{fig:methods_phylora}). Specifically, the raw observation $X$ is passed to a shared feature extractor $\mathcal{E}_{\mathrm{R}}$ which projects $X$ into a high-dimensional feature representation $R = \mathcal{E}_{\mathrm{R}}(X)$. Two lightweight encoders map $R$ into physical variables $Z_{\mathrm{phy}} = \mathcal{E}_{\mathrm{phy}}(R)$ and auxiliary variables $Z_{\mathrm{aux}} = \mathcal{E}_{\mathrm{aux}}(R)$.

The physical pathway uses $\mathcal{F}$ as its decoder to reconstruct $X_{\mathcal{F}}$ as in \cref{eq:decode_phy}. The auxiliary pathway learns a mapping $\mathcal{C}$ from $Z_{\mathrm{aux}}$ to the residual $\Delta$ that can be conditioned on $X_{\mathcal{F}}$ as in \cref{eq:decode_aux}. A stop-gradient operation is applied to the condition $X_{\mathcal{F}}$ during backpropagation to block gradients from flowing back into $\mathcal{F}$, equivalent to calling \texttt{.detach()} on $X_{\mathcal{F}}$ in PyTorch. 
This supplies physical context to the auxiliary pathway while ensuring that its gradients do not update the physical pathway, keeping the two pathways separate.

The augmented reconstruction $X_{\mathcal{C}}$ is obtained by adding the residual to the physical reconstruction:
\begin{equation}\label{eq:decode_compensate}
    X_{\mathcal{C}} = X_{\mathcal{F}} + \Delta.
\end{equation}

Here, $\Delta$ represents the discrepancy between $X$ and $X_{\mathcal{F}}$, formulated as an additive component in the observation space, which is semantically clear. Importantly, this does not imply that $\Delta$ is a constant offset to $X_{\mathcal{F}}$. $\Delta$ can be nonlinear and state-dependent, as $Z_{\mathrm{aux}}$ is encoded through a nonlinear mapping of $X$, and $X_{\mathcal{F}}$ is provided as physical context in \cref{eq:decode_aux}.

A straightforward learning objective minimizes the mean squared error (MSE) between the augmented reconstruction and the raw observation:
\begin{equation}\label{eq:loss_recontruction}
    \mathcal{L}_{\mathrm{rec}} =\lVert X - X_{\mathcal{C}} \rVert_2^2 .
\end{equation}

However, optimising only $\mathcal{L}_{\mathrm{rec}}$ does not ensure a plausible retrieval of $Z_{\mathrm{phy}}$. Although $\Delta$ can augment the incomplete $\mathcal{F}$ and reduce reconstruction error, too much flexibility in the mapping function $\mathcal{C}$ can cause $Z_{\mathrm{phy}}$ to drift away from physical interpretations. The key challenge is therefore to balance the complexity of $\Delta$ with the interpretability of $Z_{\mathrm{phy}}$.

\paragraph{Low-rank augmentation to physical reconstruction.}

Instead of optimising the pretrained weight $\mathbf{W}_0\in\mathbb{R}^{d\times k}$ of dense neural network layers, LoRA learns a low-rank matrix $\Delta\mathbf{W}\in\mathbb{R}^{d\times k}$:
\begin{equation}\label{eq:lora_llm}
    \Delta\mathbf{W} = \mathbf{B}\mathbf{A},
\end{equation}
where $\mathbf{B}\in\mathbb{R}^{d\times r}$ and $\mathbf{A}\in\mathbb{R}^{r\times k}$ are the decomposition matrices, and $\operatorname{rank}(\Delta\mathbf{W}) = r \ll \min(d,k)$. For example, when fine-tuning a 178B-parameter GPT-3 model, $d$ and $r$ can be 12{,}888 and 2, respectively, making the adaptation highly efficient. If this gain in efficiency does not compromise downstream performance, it indicates that \textbf{only certain aspects} of a pretrained weight need modification and that a low-rank parameterisation is sufficient to do that. 

Analogously, we propose learning $\Delta\in\mathbb{R}^{n\times d}$ as a low-rank decomposition that augments $X_{\mathcal{F}}\in\mathbb{R}^{n\times d}$, where $n$ is the batch size and $d$ is the dimensionality of each observation:
\begin{equation}\label{eq:lora_delta}
    \Delta = s \cdot\mathbf{A}\mathbf{B}^\top,
\end{equation}
where $\mathbf{A}\in\mathbb{R}^{n\times r}$, $\mathbf{B}\in\mathbb{R}^{d\times r}$, $s\in\mathbb{R}$, and $\operatorname{rank}(\Delta) = r \ll d$. We learn a scaling factor $s$ and a basis matrix $\mathbf{B}$ shared between all samples and a sample-specific coefficient matrix $\mathbf{A}$ obtained from a linear mapping of $Z_{\mathrm{aux}}\in\mathbb{R}^{n\times r}$ and $X_{\mathcal{F}}\in\mathbb{R}^{n\times d}$, followed by an activation $\arctan$ in the bound entries of $\mathbf{A}$ in $(-1,1)$:
\begin{equation}\label{eq:lora_delta_coefficient}
    \mathbf{A} = \arctan\!\left([Z_{\mathrm{aux}}\ X_{\mathcal{F}}]\mathbf{W}^\top + \mathbf{b}\right).
\end{equation}

We regularise the residual by encouraging orthonormality of the basis columns in $\mathbf{B}$ through a Frobenius penalty:
\begin{equation}\label{eq:loss_residual_basis}
    \mathcal{L}_{\mathrm{res}} = \lVert \mathbf{B}^\top\mathbf{B} - \mathbf{I} \rVert_F^2.
\end{equation}

This encourages the columns of the basis matrix $\mathbf{B}$ to be orthonormal,
decoupling the directions of variation of the low-rank residual, and therefore ensures the most efficient and expressive use of the rank. 
Since both $\mathbf{A}$ and $\mathbf{B}$ are constrained, we also learn a single scaling factor $s$ that controls the global magnitude of $\Delta$ for all samples. The value of $s$ is implicitly constrained by the reconstruction loss in \cref{eq:loss_recontruction}. 
At the start of training, $s$ is initialised as 1, elements of $\mathbf{A}$ near 0, and $\mathbf{B}$ as an orthonormal matrix. Thus, $\Delta$ initially assumes $\mathcal{F}$ as almost correct, and then increases to accommodate the incompleteness of $\mathcal{F}$, with complexity constrained by rank $r$. The number of auxiliary variables is set equal to $r$.

\paragraph{Prior to regularise the physical space.}

So far, the inferred $Z_{\mathrm{phy}}$ is not guaranteed to be physically meaningful. When reconstruction loss dominates the objective, the model can use $Z_{\mathrm{phy}}$ to explain observations that should instead be absorbed by $Z_{\mathrm{aux}}$. Imposing a prior on $Z_{\mathrm{phy}}$ reduces this ambiguity and, together with the regularisation of $\Delta$, they help to keep $Z_{\mathrm{phy}}$ close to meaningful physical values.

Given sufficient physical knowledge, an informative prior on $Z_{\mathrm{phy}}\in\mathbb{R}^k$ could be formulated as:
\begin{equation}\label{eq:informative_prior}
    p(Z_{\mathrm{phy}}) = \mathcal{N}\!\left(Z_{\mathrm{phy}}\,\middle|\, \boldsymbol{\mu}_{\mathrm{phy}},\ \boldsymbol{\Sigma}_{\mathrm{phy}} \right),
\end{equation}
where $\boldsymbol{\mu}_{\mathrm{phy}}\in\mathbb{R}^k$ and $\boldsymbol{\Sigma}_{\mathrm{phy}}\in\mathbb{R}^{k\times k}$ are the prior mean and covariance. We might apply a variational auto-encoder with a Kullback–Leibler (KL) divergence to integrate this prior:
\begin{equation}\label{eq:loss_prior_KL}
    \mathcal{L}_{\mathrm{prior}} = \mathrm{KL}\!\left(q(Z_{\mathrm{phy}}\mid X)\,\|\,p(Z_{\mathrm{phy}})\right).
\end{equation}
However, in reality, an informative prior is often unavailable. For the inverse problems considered in this paper, we know only the physical range and extreme boundary values are largely implausible, leading to an end-stop loss:
\begin{equation}\label{eq:loss_prior_end_stop}
    \mathcal{L}_{\mathrm{prior}} = \log(Z_{\mathrm{phy}}) + \log(1 - Z_{\mathrm{phy}}) .
\end{equation}
It penalises values near the extremes of the physical range but is almost uniform elsewhere. Without the earlier KL term, the architecture is no longer a variational auto-encoder, but a deterministic one.

\paragraph{Training objective.}
The full training objective is therefore the following:
\begin{equation}\label{eq:pila_loss}
    \mathcal{L} = \mathcal{L}_{\mathrm{rec}} + \beta\,\mathcal{L}_{\mathrm{prior}} + \lambda\,\mathcal{L}_{\mathrm{res}},
\end{equation}
where the reconstruction loss $\mathcal{L}_{\mathrm{rec}}$, the residual loss $\mathcal{L}_{\mathrm{res}}$, and the prior loss $\mathcal{L}_{\mathrm{prior}}$ are defined in \cref{eq:loss_recontruction}, \cref{eq:loss_residual_basis}, and \cref{eq:loss_prior_end_stop}, respectively. 
Hyperparameters $\lambda$ and $\beta$ are chosen to balance the loss terms ---  details of the implementation can be found in \cref{appx:implementation_detals}. 

\subsection{Case studies for Earth observation and physical models}\label{sec:physical-models}

\textbf{Radiative transfer model:} The INFORM radiative transfer model of optical remote sensing (denoted $\mathcal{F}_{\text{RTM}}$), developed by \citet{atzberger2000development}, combines radiative transfer processes of the Earth surface at different levels to simulate the reflectance of the forest canopy. The building blocks of INFORM include the PROSPECT model \citep{feret2017prospect} that simulates leaf optical properties from biochemical parameters such as chlorophyll content $Z_{\mathrm{cab}}$ and dry matter $Z_{\mathrm{cm}}$, returning per-wavelength leaf reflectance and transmittance. On top of PROSPECT is the SAIL model \citep{verhoef1985earth} that simulates how light is scattered and transmitted through a canopy. Its main input is the canopy leaf area index ($Z_{\mathrm{LAI}}$), which controls the probability that light is scattered through the canopy. Based on canopy reflectance, INFORM includes the FLIM model \citep{rosema1992new} to introduce geometric mixing rules at the forest-stand level using parameters such as fractional cover ($Z_{\mathrm{fc}}$) and the understory leaf area index ($Z_{\mathrm{LAIu}}$) (\cref{appx:rtm_modelling}).

Forest structural complexity causes discrepancies between simulated and measured spectra, such as overestimation in visible bands at lower ranges and underestimation in near-infrared bands at higher ranges (\cref{fig:appx_rtm_inversion_austria_ablations_spectra_no_adaptation_no_prior} in \cref{appx:ablations_residual_prior}). 
We aim to estimate a subset of biophysical variables whose plausible ranges are known (\cref{tab:inform_para_list}). Other variables are fixed at typical values for the study area or regressed using allometric equations (see \cref{appx:rtm_vars}). 
$\mathcal{F}_{\text{RTM}}$ was originally implemented in NumPy and was not explicitly differentiable, which prevents gradient-based training. We therefore reimplemented it in PyTorch, replacing non-differentiable operations with stable approximations (\cref{appx:making_diff_rtm}) and handling numerical instabilities during backpropagation (\cref{appx:bypassing_instability}). The final implementation provides a \textbf{fully differentiable} RTM suitable for end-to-end learning.

\begin{table*}[htbp]
\caption{\textbf{Learned biophysical variables in $\mathcal{F}_{\text{RTM}}$.} \textit{Other canopy variables, such as canopy height and crown diameter, are inferred using allometric equations \citep{jucker2017allometric} to mitigate the ill-posed nature of the inverse problem (see \cref{appx:rtm_vars}).}}
\label{tab:inform_para_list}
\centering
\resizebox{\textwidth}{!}{
    \begin{tabular}{lccccccc}
    \toprule
    Sub-model & \multicolumn{4}{|c|}{Leaf Model} & \multicolumn{1}{c|}{Canopy Model} & \multicolumn{2}{c}{Forest Model} \\
    \midrule
    Variable & \makecell{Structure\\ Parameter} & \makecell{Chlorophyll\\ A+B} & \makecell{Water\\ Content} & \makecell{Dry\\ Matter} & \makecell{Leaf Area\\ Index} & \makecell{Undergrowth\\ LAI} & \makecell{Fractional\\ Coverage} \\
    \midrule
    Acronym & N & cab & cw & cm & LAI & LAIu & fc \\
    \midrule
    Min & 1 & 10 & 0.001 & 0.005 & 0.01 & 0.01 & 0.1 \\
    Max & 3 & 80 & 0.02 & 0.05 & 5 & 3 & 1 \\
    \bottomrule
    \end{tabular}
}
\end{table*}

\textbf{Volcanic deformation model:} The Mogi model (denoted $\mathcal{F}_{\text{Mogi}}$) describes the displacement field on the Earth’s surface resulting from a spherical pressure source, typically a magma chamber, at depth \citep{kiyoo1958relations}. The inverse problem for $\mathcal{F}_{\text{Mogi}}$ involves estimating the location (horizontal coordinates $x_m$ and $y_m$, and depth $d$) and volume change ($\Delta V$) through time of the magma chamber from observed surface displacements (\cref{tab:mogi_para_list} and \cref{appx:mogi_descriptions}). Compared to $\mathcal{F}_{\text{RTM}}$, $\mathcal{F}_{\text{Mogi}}$ relies on simple assumptions, approximating the Earth’s crust as an elastic half-space. Despite its simplicity, the Mogi model captures first-order patterns of surface deformation associated with volcanic sources. It is intended to model only transient displacements, whereas GNSS-observed surface displacements can be significantly influenced by other physical processes, such as long-term tectonic plate motion, seasonal hydrological loading, and coloured noise. The recorded surface displacement can be small (usually at millimetre level) compared to other sources (e.g., hydrological loading at millimetre to centimetre levels). Accurate retrieval of geophysical variables from raw observations is possible therefore only by carefully allowing for the relatively high incompleteness of the model. See \cref{appx:gnss_data_components} for details on the assumed components of surface displacements around a volcano.

\begin{table*}[htbp]
\caption{\textbf{Learned geophysical variables in $\mathcal{F}_{\text{Mogi}}$.} \textit{The source's horizontal location $(x_m, y_m)$ is assumed to lie within the GNSS stations' boundaries, with depth and volume ranges matching those used in \citep{walwer2016data}.}}
\label{tab:mogi_para_list}
\centering
\resizebox{0.9\textwidth}{!}{
    \begin{tabular}{lcccc}
    \toprule
    Variable & Location-X ($\mathrm{km}$) & Location-Y ($\mathrm{km}$) & Depth ($\mathrm{km}$) & Volume Change ($10^6~\mathrm{m^3}$)\\
    \midrule
    Acronym & $x_m$ & $y_m$ & $d$ & $\Delta V$ \\
    \midrule
    Min & -9.33 & -5.80 & 2 & -10 \\
    Max & 14.35 & 7.62 & 20 & 10 \\
    \bottomrule
    \end{tabular}
}
\end{table*}

\subsection{Datasets}\label{sec:datasets}
For RTM inversion, we use spectra observed by Sentinel-2 \citep{drusch2012sentinel} ($X_{\text{S2}}$), one of the most widely used optical remote sensing platforms, at $20~\mathrm{m} \times 20~\mathrm{m}$ spatial resolution. Sentinel-2 spectra comprise 11 bands that cover a wide electromagnetic spectrum, including visible light, near infrared, and shortwave infrared (see \cref{appx:s2_data}). Specifically, we use two datasets:  
(1) The \textbf{Austrian} dataset \citep{immitzer2019optimal} consists of 17,692 individual spectra extracted from multiple sites across the Wienerwald Biosphere Reserve, Austria observed by Sentinel-2. Spectra cover the period from April to October 2018. This dataset contains coniferous and broadleaf forests, with detailed species information for each spectrum (12 species in total). These species labels are used to validate the plausibility of the biophysical patterns of the retrieved variables.  
(2) The \textbf{Wytham} dataset consists of spectra extracted from Wytham Woods, UK, from available observations in June and July 2018, covering the period (3--5 July 2018) when ground measurements of several biophysical variables (chlorophyll, fractional coverage, canopy LAI and understory LAI) were collected \citep{origo2020fiducial}. These ground measurements facilitate validation of the accuracy of the retrieved variables.

For the Mogi inversion, the dataset comes from 12 GNSS stations around the Akutan Volcano, one of the most active volcanoes in the Aleutian Islands (Alaska, USA). Each day, every GNSS station records its position on the surface in the east, north, and vertical directions (denoted $X_{\text{GNSS}}$). In total, we use observations from 6,525 consecutive days from 2006 until early 2024 ({\texttt{http://geodesy.unr.edu}}, \citep{blewitt2018harnessing}) (see \cref{appx:gnss_data}). In particular, a major volcanic eruption at Akutan in 2008 has been widely studied \citep{ji2011transient, walwer2016data, shi2025fine}. The authors applied various pre-processing strategies aimed at extracting the volcanic deformation at the surface. This allowed them to invert the volcanic deformation signal and estimate the source's depth, horizontal location, and volume change. We use this information to validate the retrieved geophysical variables --- specifically whether they capture the 2008 eruption and whether the estimated magnitude agrees with existing studies.

\subsection{Baseline}\label{sec:baseline_HVAE}
For comparison, we use the Hybrid Variational Autoencoder (HVAE) \citep{takeishi2021physics}, which represents the state of the art in learning the inverse of incomplete physical models. Although HVAE was developed primarily for idealised dynamical systems such as the forced damped pendulum and advection–diffusion, the authors also demonstrated a static example: inverting galaxy images, where a parametric galaxy distribution acts as the physical forward model. All applications follow the same principles and share many hyperparameters, but HVAE provides separate implementations for each study case depending on the physical prior and data modality. We therefore adapt, as our baseline, the HVAE implementation that was used with galaxy images, since this setting aligns most closely with our static Earth-observation operators. More details of the implementation can be found in \cref{appx:implementation_detals}. 

\begin{figure}[htbp]
    \centering
    \includegraphics[width=\textwidth]{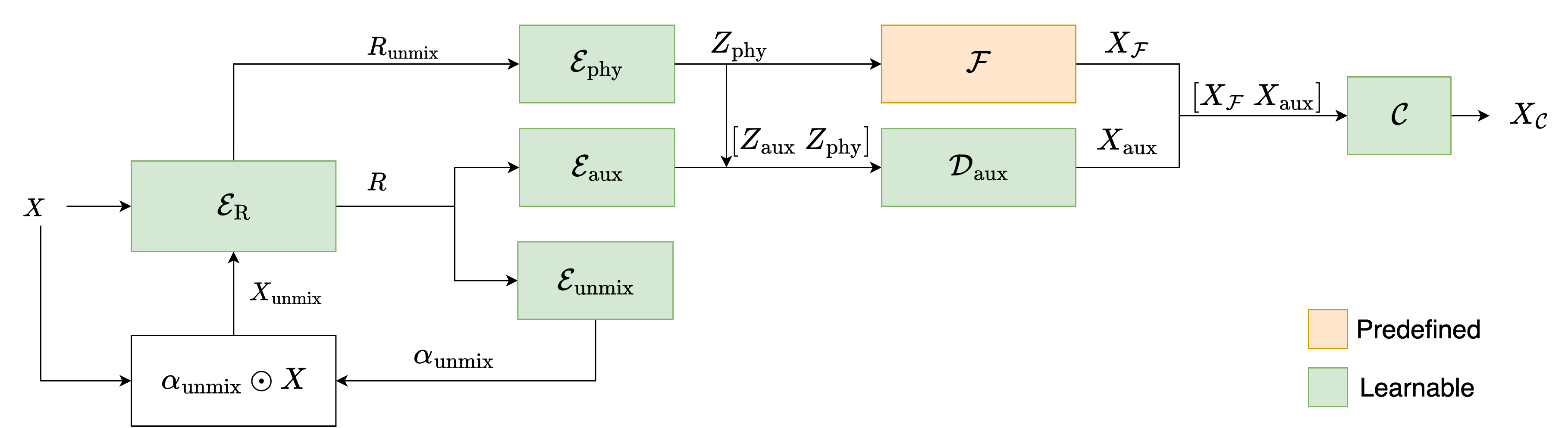}
    \caption{\textbf{HVAE contains multiple regularisation terms that can be difficult to tune in practice.}
    \textit{Given an observation $X$, as in PILA, $\mathcal{E}_{\mathrm{R}}$ maps it to a high-dimensional feature $R$. A physical pathway predicts $Z_{\mathrm{phy}}$ and reconstructs $X_{\mathcal{F}}$, while an auxiliary pathway predicts $Z_{\mathrm{aux}}$ and augments $X_{\mathcal{F}}$ to produce the refined reconstruction $X_{\mathcal{C}}$. The architecture also includes an ``unmixing'' step to separate the modeled observation from the raw observation.: $R$ is passed to $\mathcal{E}_{\mathrm{unmix}}$ to obtain coefficients $\alpha_{\mathrm{unmix}}$, which are multiplied element-wise with $X$ to yield the unmixed observation $X_{\mathrm{unmix}}$. This is fed again to $\mathcal{E}_{\mathrm{R}}$ to obtain the unmixed feature $R_{\mathrm{unmix}}$, followed by $\mathcal{E}_{\mathrm{phy}}$ to produce physical variables $Z_{\mathrm{phy}}$. Then $Z_{\mathrm{phy}}$ is concatenated with $Z_{\mathrm{aux}}$ and passed through the decoder $\mathcal{D}_{\mathrm{aux}}$ to obtain the auxiliary feature $X_{\mathrm{aux}}$. Finally, $X_{\mathrm{aux}}$ is concatenated with $X_{\mathcal{F}}$ and mapped by a non-linear function $\mathcal{C}$ to produce $X_{\mathcal{C}}$. Several regularisation terms are added to enforce desired behavior.}
    }
    \label{fig:methods_physvae}
\end{figure}

We briefly summarise the HVAE algorithm using the notation of our method to clarify similarities and differences (\cref{fig:methods_physvae}). Using an auto-encoder architecture, HVAE includes a physical pathway to invert the physical model and an auxiliary pathway to learn the unmodelled component. Specifically, during encoding, an observation $X$ is passed through a shared feature extractor $\mathcal{E}_{\mathrm{R}}$ that produces an intermediate representation $R=\mathcal{E}_{\mathrm{R}}(X)$. Before applying $\mathcal{E}_{\mathrm{phy}}$, there is an additional “unmixing” loop with $\mathcal{E}_{\mathrm{unmix}}$ that generates unmixing coefficients $\alpha_{\mathrm{unmix}}=\mathcal{E}_{\mathrm{unmix}}(R)$. The original observation is then “cleansed” to produce $X_{\mathrm{unmix}}=\alpha_{\mathrm{unmix}}\odot X$, where $\odot$ denotes element-wise multiplication, and the physical encoder maps this to $Z_{\mathrm{phy}}=\mathcal{E}_{\mathrm{phy}}(\mathcal{E}_{\mathrm{R}}(X_{\mathrm{unmix}}))$. Another lightweight encoder $\mathcal{E}_{\mathrm{aux}}$ is used to obtain the auxiliary variables $Z_{\mathrm{aux}}=\mathcal{E}_{\mathrm{aux}}(R)$. 

During decoding, $Z_{\mathrm{phy}}$ is passed through the physical model for the physical reconstruction $X_{\mathcal{F}}=\mathcal{F}(Z_{\mathrm{phy}})$. For the auxiliary pathway, $[Z_{\mathrm{aux}}\ Z_{\mathrm{phy}}]$ is passed to a multi-layer perceptron (MLP) decoder $\mathcal{D}_{\mathrm{aux}}$ to compute auxiliary features $X_{\mathrm{aux}}=\mathcal{D}_{\mathrm{aux}}([Z_{\mathrm{aux}}\ Z_{\mathrm{phy}}])$. Together with the physical reconstruction, $[X_{\mathcal{F}}\ X_{\mathrm{aux}}]$ is passed to another decoder $\mathcal{C}$ to produce the final reconstruction $X_{\mathcal{C}}=\mathcal{C}([X_{\mathcal{F}}\ X_{\mathrm{aux}}])$. In the original galaxy example, $\mathcal{C}$ is implemented as a small U-Net, since it processes 2D images; here we use a lightweight MLP because each data sample is a 1D vector.

In addition to the reconstruction loss $\mathcal{L}_{\mathrm{rec}}(X_{\mathcal{C}}, X)$ defined as in \cref{eq:loss_recontruction}, HVAE uses three regularisation terms with the motivation of limiting the flexibility of the auxiliary pathway and anchoring $Z_{\mathrm{phy}}$ to the causal structure defined by $\mathcal{F}$:

Regularisation of the unmixing loop $\mathcal{E}_{\mathrm{unmix}}$, encouraging $X_{\mathrm{unmix}}$ (the input for physical encoding) to align with $X_{\mathcal{F}}$:
\begin{equation}\label{eq:pvae_loss_unmix}
    \mathcal{L}_{\mathrm{unmix}} = \lVert X_{\mathrm{unmix}} - X_{\mathcal{F}} \rVert_2^2.
\end{equation}

Regularisation of the physics encoder $\mathcal{E}_{\mathrm{phy}}$ using synthetic data $(Z'_{\mathrm{phy}}, X'_{\mathcal{F}})$ generated by $\mathcal{F}$, ensuring that the encoding process will maintain its physical meaning:
\begin{equation}\label{eq:pvae_loss_syn}
    \mathcal{L}_{\mathrm{syn}} = \lVert Z'_{\mathrm{phy}} - \mathcal{E}_{\mathrm{phy}}(\mathcal{E}_{\mathrm{R}}(X'_{\mathcal{F}})) \rVert_2^2.
\end{equation}

Regularisation of the mapping of $\mathcal{C}$ to minimise reliance on the residual $\Delta=X_{\mathcal{C}}-X_{\mathcal{F}}$:
\begin{equation}\label{eq:pvae_loss_least}
    \mathcal{L}_{\mathrm{res}} = \lVert \Delta \rVert_2^2.
\end{equation}

HVAE also imposes priors on the latent space. For physical variables $Z_{\mathrm{phy}}$, the prior $p(Z_{\mathrm{phy}})$ is claimed to be based on physical knowledge, as formulated in \cref{eq:informative_prior}. For $Z_{\mathrm{aux}}$, a standard normal prior is applied. Using a variational autoencoder formulation, the regularisation is:
\begin{equation}\label{eq:pvae_loss_prior}
    \mathcal{L}_{\mathrm{prior}} = \mathrm{KL}\!\left(q(Z_{\mathrm{phy}} \mid X)\,\|\,p(Z_{\mathrm{phy}})\right) + \mathrm{KL}\!\left(q(Z_{\mathrm{aux}} \mid X)\,\|\,\mathcal{N}(0,I)\right).
\end{equation}

The training objective is:
\begin{equation}\label{eq:pvae_loss}
    \mathcal{L} = \mathcal{L}_{\mathrm{rec}} + \beta\,\mathcal{L}_{\mathrm{prior}} + \lambda_1\,\mathcal{L}_{\mathrm{unmix}}
                 + \lambda_2\,\mathcal{L}_{\mathrm{syn}} + \lambda_3\,\mathcal{L}_{\mathrm{res}},
\end{equation}
where $\mathcal{L}_{\mathrm{rec}}$, $\mathcal{L}_{\mathrm{prior}}$, $\mathcal{L}_{\mathrm{unmix}}$, $\mathcal{L}_{\mathrm{syn}}$, and $\mathcal{L}_{\mathrm{res}}$ are defined in \cref{eq:loss_recontruction}, \cref{eq:pvae_loss_prior}, \cref{eq:pvae_loss_unmix}, \cref{eq:pvae_loss_syn}, and \cref{eq:pvae_loss_least}, respectively.

HVAE shares our motivation that the incomplete $\mathcal{F}$ should be augmented just enough to allow a meaningful retrieval of $Z_{\mathrm{phy}}$. Each assumption appears reasonable in isolation. However, the behaviour of each regularisation term in constraining the complexity of the data-driven component is opaque. For example, how “unmixed” should $X$ be? If $\mathcal{F}$ is incomplete, how much should we trust the synthetic data $(Z'_{\mathrm{phy}}, X'_{\mathcal{F}})$? How should residuals be minimised? As noted by \citep{takeishi2023deep}, it is difficult to justify whether to rely on a particular regulariser or a combination of them, so simply minimising these loss terms does not necessarily yield an accurate retrieval of physical variables. In contrast, in PILA the low-rank residual has a clear role: its rank and associated regularisation (\cref{eq:loss_residual_basis}) directly control the complexity of $\Delta$, making its effect easier to interpret and to tune. 

Additionally, although HVAE proposes that the prior should reflect physical knowledge (\cref{eq:informative_prior}), in practice the same generic prior is used for all cases (pendulum, diffusion, galaxy, etc.), namely $p(Z_{\mathrm{phy}})=\mathcal{N}(0.5,\ (0.866)^2\,\mathbf{I})$, after scaling all variable ranges to $(0,1)$ with  mean centred in the physical range and arbitrary variance. The KL weight $\beta$ is set small to limit the influence of the prior (see \cref{appx:HVAE_formulations}). 
In contrast, when an informative prior is unavailable, PILA uses a simple end-stop prior to enforce known physical bounds without imposing an arbitrary distribution within them, thereby minimising unnecessary interference with the learning process.

%% file: 4_results.tex
\section{Results} \label{sec:results}
\subsection{Inversion of Sentinel-2 spectra for forest monitoring}\label{sec:results_rtm}

\paragraph{Austrian dataset:}

\begin{figure}[htbp]
    \centering
    \includegraphics[width=0.9\textwidth]{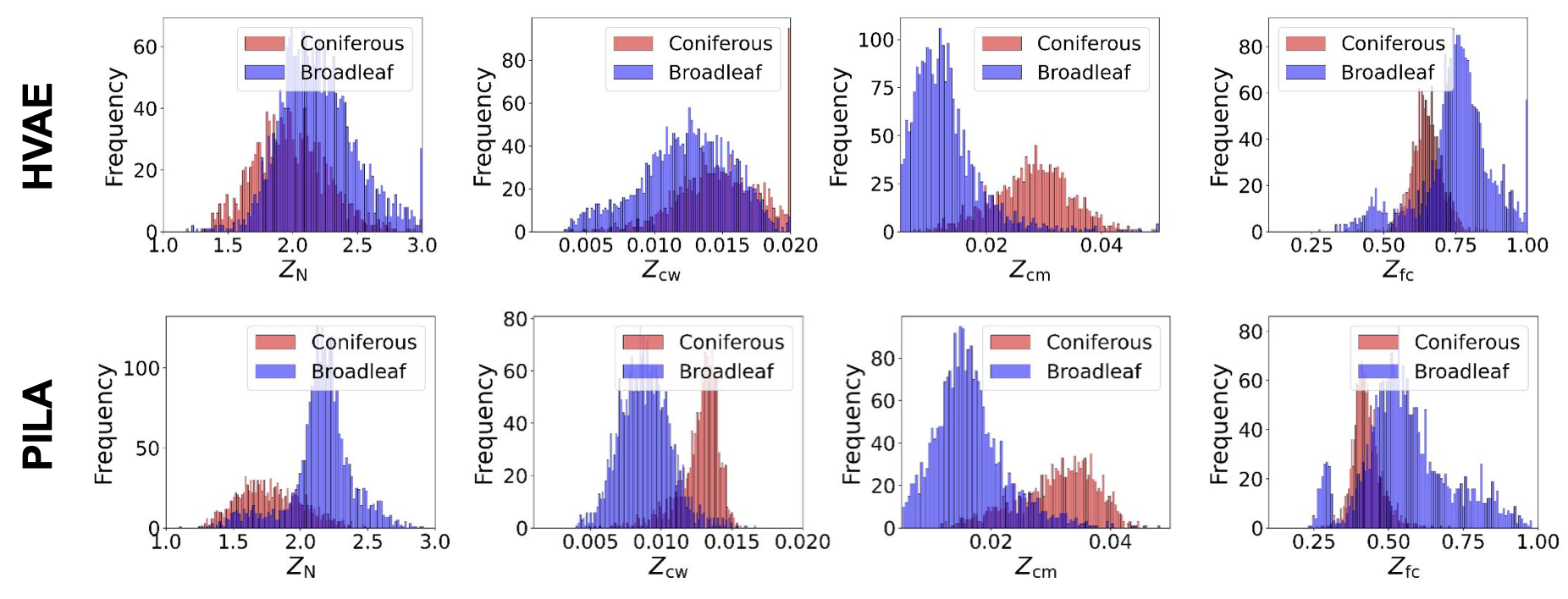}
    \caption{\textbf{PILA (with $r=2$) yields more plausible distributions of physical variables.}
    \textit{In contrast, variables recovered by HVAE frequently fall at boundary values. For example, $Z_{\mathrm{cm}}=0$ means no dry matter --- but there is foliage present.}}
    \label{fig:rtm_inversion_austria_variables}
\end{figure}

\begin{figure}[t]
    \centering
    \includegraphics[width=0.9\textwidth]{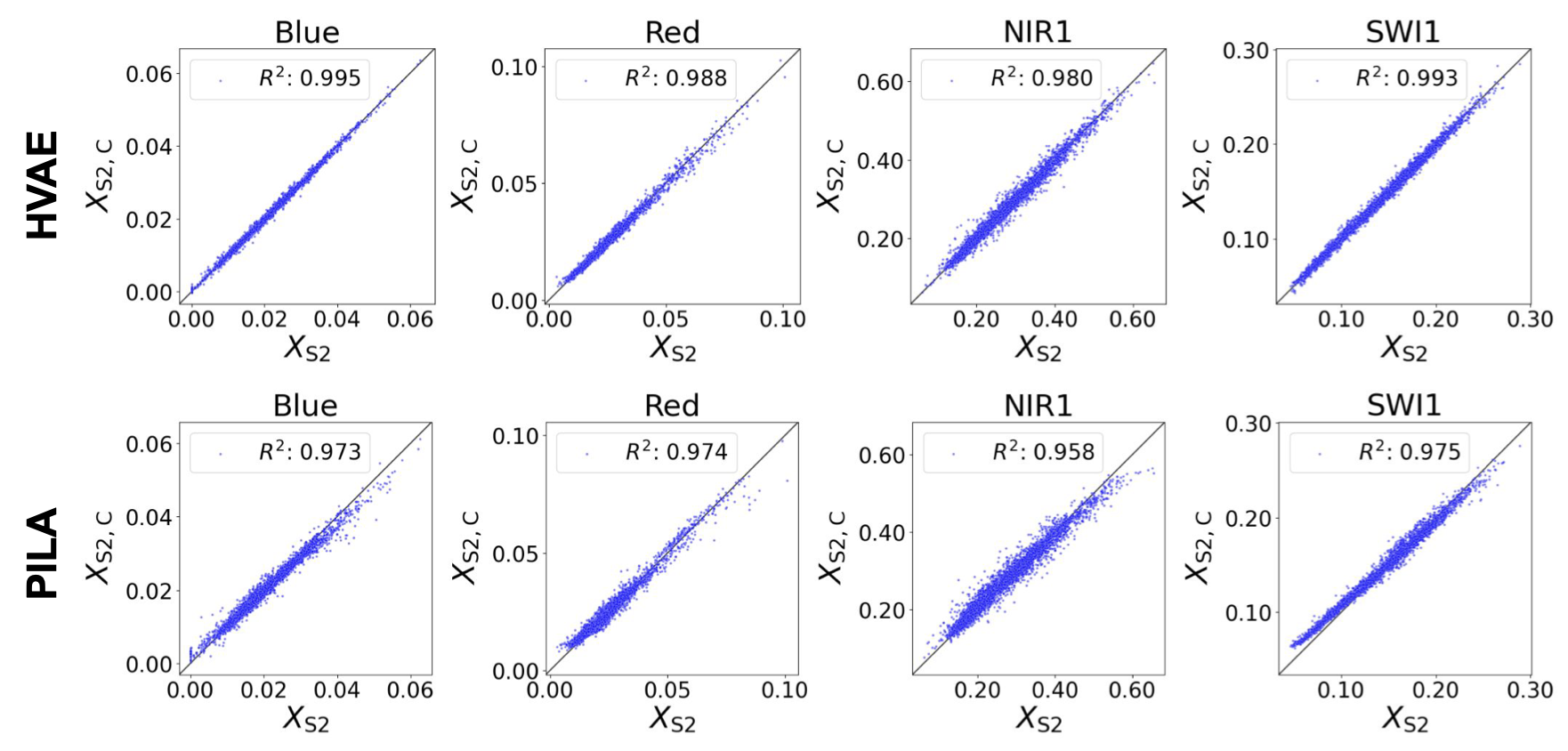}
    \caption{\textbf{PILA achieves less perfect reconstruction of spectral bands ($r=2$).}
    \textit{HVAE attains slightly higher reconstruction accuracy than PILA but the trade-off is that PILA frequently recovers more plausible values of physical variables (see \cref{fig:rtm_inversion_austria_variables}).}}
    \label{fig:rtm_inversion_austria_spectra}
\end{figure}

\begin{figure}[htbp]
    \centering
    \includegraphics[width=0.9\textwidth]{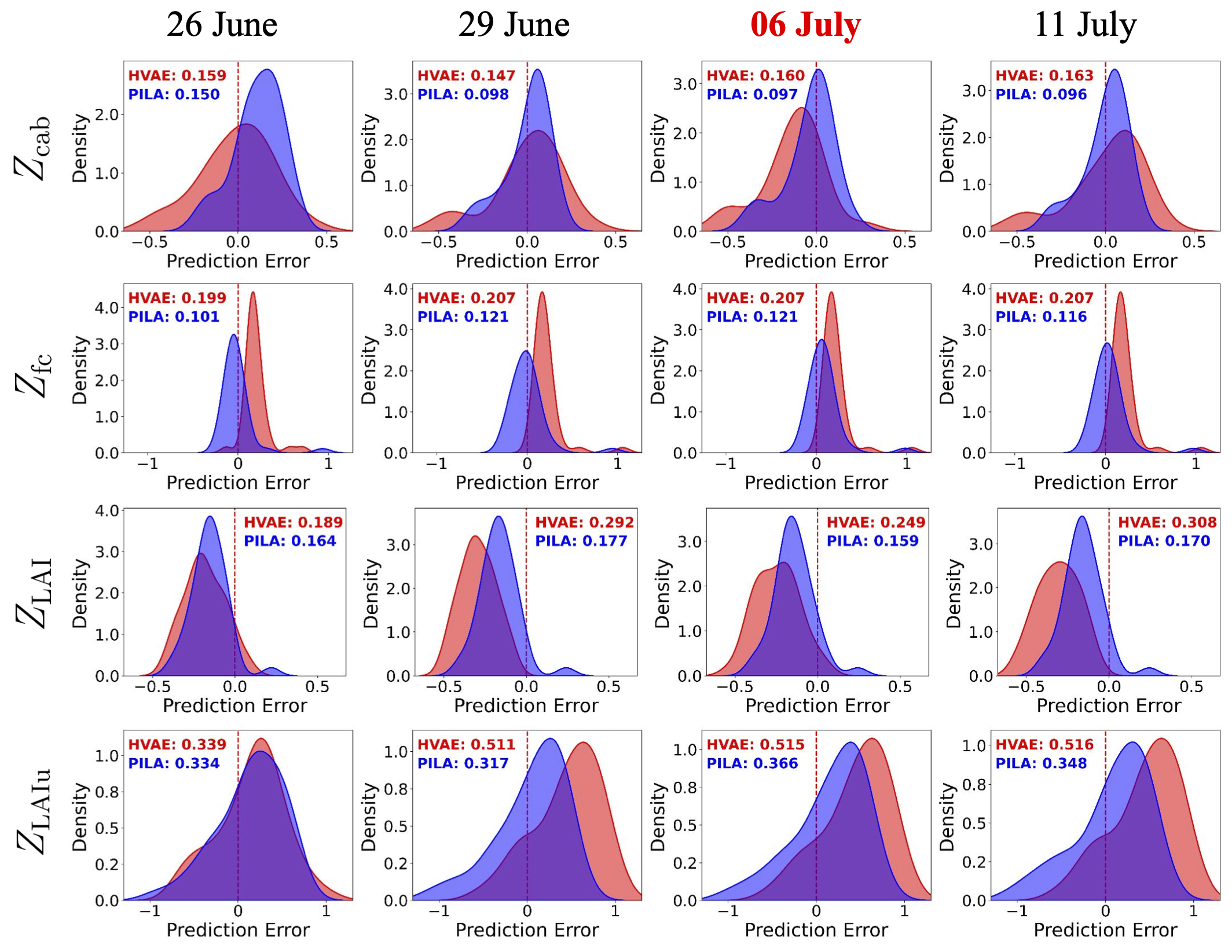}
    \caption{\textbf{PILA recovers biophysical forest variables more accurately than HVAE does.} 
    \textit{For each of the 42 ground measurement sites in Wytham Woods, we compute the prediction error as the difference between predictions and ground truth (both normalised) and estimate its kernel density. We also calculate the Mean Absolute Error (MAE) across all sites and indicate it on each plot. Ground measurements were conducted between 3–5 July 2018. PILA achieves consistently smaller MAE across all dates and variables. For both $Z_{\mathrm{cab}}$ and $Z_{\mathrm{fc}}$,  PILA exhibits less prediction bias when compared with HVAE, particularly on 6 July—the date closest to the ground measurement period. Both models exhibit larger errors for $Z_{\mathrm{LAI}}$ and $Z_{\mathrm{LAIu}}$ due to limits of observability and the physical model (\cref{sec:sensitivity}), yet our predictions remain closer to the ground truth values.}
    }
    \label{fig:rtm_inversion_wytham_variables_prediction_error_density}
\end{figure}

To invert the RTM, we use minimal rank $r=2$ in PILA to learn $\Delta_{\mathrm{S2}}$  and augment the incomplete model $\mathcal{F}_{\mathrm{RTM}}$. 
In terms of variable retrieval, both HVAE and PILA learn distributions that are consistent with expectations based on forest types (\cref{fig:rtm_inversion_austria_variables}). For example, broadleaf forests typically exhibit more complex canopy structures (higher $Z_{\mathrm{N}}$) and a larger fractional coverage ($Z_{\mathrm{fc}}$) than coniferous forests. HVAE \citep{takeishi2021physics} is a more flexible neural model and so achieves a higher reconstruction accuracy of the spectra (\cref{fig:rtm_inversion_austria_spectra}). However the physical variables are more spread out and saturate significantly at boundary values. These saturated retrievals are generally physically implausible. Full validation on the Wytham dataset will further confirm this limitation (\cref{fig:rtm_inversion_wytham_variables_prediction_error_density}). More details can be found in \cref{appx:results_rtm_austria}. 

\begin{figure}[htbp]
    \centering
    \includegraphics[width=0.9\textwidth]{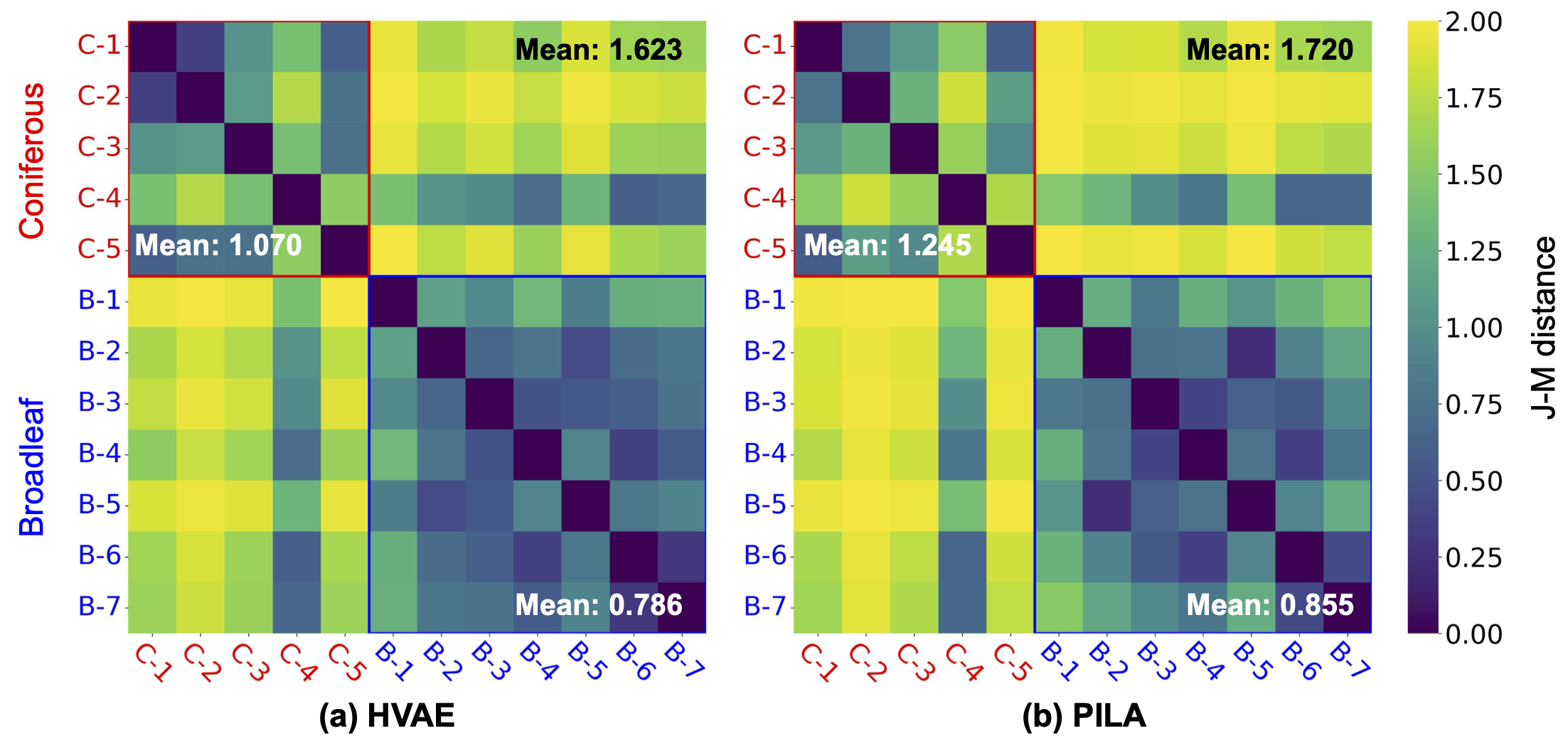}
    \caption{\textbf{PILA learns more separated representations that align with the biophysical patterns of tree species.} 
    \textit{This figure shows the pairwise Jeffreys–Matusita distance between species, each represented by a vector of retrieved biophysical variables. J-M distance of 2 indicates complete separability; 0 indicates none. C-4 (Larix decidua) appears close to the broadleaf group, as it is one of the few coniferous species that shed their needles in the fall, resulting in similar spectral characteristics.}
    }
    \label{fig:rtm_inversion_austria_variables_JM_distance}
\end{figure}
\vspace{-2mm}

PILA learns biophysical representations that have better separability between tree species (\cref{fig:rtm_inversion_austria_variables_JM_distance}). We represent each tree species by a vector of standardised values of seven inferred variables and compute the Jeffries–Matusita (J-M) distance between species using these vectors. The J-M distance varies between 0 and 2. Complete separation would yield a J–M distance of 0 on the diagonal (the same species) and 2 (complete separation) between different species, analogous to a confusion matrix. In reality, species within the same forest type (coniferous/broadleaf) usually have more similar biophysical traits and thus are less distinct than species from different forest types. Compared to HVAE, PILA shows clearer separation between forest types (mean coniferous–broadleaf distance: 1.720 vs. 1.623) and better within-type separability (coniferous: 1.245 vs. 1.070; broadleaf: 0.855 vs. 0.786), indicating stronger class separability in the inferred variables. 
More details can be found in \cref{appx:robustness_species_separability_JM_distance}.

Interestingly, the coniferous species C-4 appears close to the broadleaf group. This pattern is reasonable because C-4 is European Larch, and as its Latin name \texttt{Larix Decidua} suggests, it is one of the few conifers that shed their needles in the fall. This phenological behaviour produces spectral signals similar to those of broadleaf species when observed by optical satellite sensors. 

\paragraph{Wytham dataset:}
The Wytham dataset contains spectra only from broadleaf forests, and the region also has ground measurements of physical variables ($Z_{\mathrm{cab}}$, $Z_{\mathrm{fc}}$, $Z_{\mathrm{LAI}}$, $Z_{\mathrm{LAIu}}$) \citep{origo2020fiducial}, which we use to validate the predictions of the model. These ground measurements were conducted between 3 and 5 July 2018 and cover 42 sites in Wytham Woods, UK. More details can be found in \cref{appx:results_rtm_wytham}. 

In \cref{fig:rtm_inversion_wytham_variables_prediction_error_density}, we compare the prediction errors of HVAE and PILA against the ground measurements.
Both models predict biophysical variables at the locations of ground measurement sites using Sentinel-2 spectra acquired on four dates around the field campaign.
Taking the 6 July prediction as an example --- the closest date to the ground measurement campaign (\cref{tab:rtm_inversion_wytham_variables_insitu_july6}) --- the mean absolute error (MAE) from HVAE is between 40\% and 71\% higher for the four biophysical variables than from PILA ($Z_{\mathrm{LAIu}}$: 0.515 vs.\ 0.366; $Z_{\mathrm{fc}}$: 0.207 vs.\ 0.121). 

\begin{table}[htbp]
\centering
\caption{\textbf{PILA prediction matches the ground measurements more closely.}
\textit{Using spectra of Wytham Wood from 6 July (closest to the 3–5 July field campaign), PILA reduces the standardized MAE by 40-71\% compared to HVAE. Even for $Z_{\mathrm{LAI}}$ and $Z_{\mathrm{LAIu}}$, where retrieval is more difficult due to sensitivity issues (\cref{sec:sensitivity}), the predicted means are closer to the ground measurement values.}
}
\resizebox{0.85\textwidth}{!}{
\begin{tabular}{l *{4}{cc}} 
\toprule
\multirow{2}{*}{\textbf{Model}} & \multicolumn{2}{c}{$\mathbf{Z_{\mathrm{cab}}}$} & \multicolumn{2}{c}{$\mathbf{Z_{\mathrm{fc}}}$} & \multicolumn{2}{c}{$\mathbf{Z_{\mathrm{LAI}}}$} & \multicolumn{2}{c}{$\mathbf{Z_{\mathrm{LAIu}}}$} \\
\cmidrule(lr){2-3} \cmidrule(lr){4-5} \cmidrule(lr){6-7} \cmidrule(lr){8-9}
& \textbf{Mean} & \textbf{MAE} & \textbf{Mean} & \textbf{MAE} & \textbf{Mean} & \textbf{MAE} & \textbf{Mean} & \textbf{MAE} \\
\midrule
HVAE & 27.784 & 0.160 & 1.000 & 0.207 & 2.200 & 0.249 & 2.949 & 0.515 \\
\textbf{PILA} & 34.354 & 0.097 & 0.888 & 0.121 & 2.718 & 0.159 & 2.069 & 0.366 \\
Ground Truth & 36.640 & N/A & 0.814 & N/A & 3.421 & N/A & 1.614 & N/A \\
\bottomrule
\end{tabular}
}
\label{tab:rtm_inversion_wytham_variables_insitu_july6}
\end{table}

For both chlorophyll content ($Z_{\mathrm{cab}}$) and fractional canopy coverage ($Z_{\mathrm{fc}}$), PILA's predictions align more closely with the ground measurements, and its prediction errors are centred around zero. In contrast, HVAE prediction errors are more spread for $Z_{\mathrm{cab}}$, and its predictions for $Z_{\mathrm{fc}}$ saturate at 1, which is visualised as the over-estimation of $Z_{\mathrm{fc}}$ for all four dates. 

Both models struggle to retrieve the other two canopy variables: canopy leaf area index ($Z_{\mathrm{LAI}}$) and understory leaf area index ($Z_{\mathrm{LAIu}}$). These two variables represent the same metric—leaf area—but in different layers of the canopy. The predicted $Z_{\mathrm{LAI}}$ values cluster around 3, while ground measurements apparently exceed this value, which visually is an underestimate of $Z_{\mathrm{LAI}}$ in \cref{fig:rtm_inversion_wytham_variables_prediction_error_density}, indicating very dense canopy layers. Similarly, both models show a limited ability to predict $Z_{\mathrm{LAIu}}$, despite the large variability in the ground measurements. However, the mean predictions of PILA remain closer to the ground measurement means for both variables (\cref{tab:rtm_inversion_wytham_variables_insitu_july6}).

This limitation in retrieving $Z_{\mathrm{LAI}}$ and $Z_{\mathrm{LAIu}}$ is related to the observability of the physical variables and the sensitivity of the physical model ($\mathcal{F}_{\mathrm{RTM}}$) under different conditions—an important aspect of the inverse problem. We discuss this further in \cref{sec:sensitivity}. 

In addition to the two Sentinel-2 datasets presented above for forest vegetation, we further evaluate PILA and the HVAE baseline on the GreenHyperSpectra dataset \citep{cherif2026greenhyperspectra}, which covers a wider range of land-cover types and involves a slightly different RTM. PILA continues to achieve more accurate retrieval results than HVAE. More details can be found in \cref{appx:rtm_evaluation_greenhyperspectra}.

\subsection{Inversion of GNSS signals for volcanic deformation}\label{sec:results_mogi}
Beyond optical sensing for retrieving the biophysical status of the Earth's surface, PILA also performs well with GNSS signals to understand the geophysical processes beneath the surface. Since the Mogi model ($\mathcal{F}_{\mathrm{Mogi}}$) only models the transient displacements of the Earth's surface caused by volcanic activity, it is relatively incomplete compared to the raw GNSS observations ($X_{\mathrm{GNSS}}$). We set the rank as $r=4$ to accommodate this residual complexity. 
The effects of rank selection on residual learning and variable retrieval are discussed in more detail in \cref{sec:rank}. More details can be found in \cref{appx:results_mogi}. 

For the Mogi model considered here, a mix of techniques has previously been applied~\citep{ji2011transient,walwer2016data,shi2025fine}, including the classic trajectory model and multivariate statistical approaches, to isolate the volcanic signal. Studying the time period 2006–2009, all these studies have revealed a transient deformation event caused by the inflation of a magma source in early 2008. The event lasted about half a year and was characterised by surface uplift and radial extension around the Mogi source. 

\begin{figure}[htbp]
    \centering
    \includegraphics[width=0.9\textwidth]{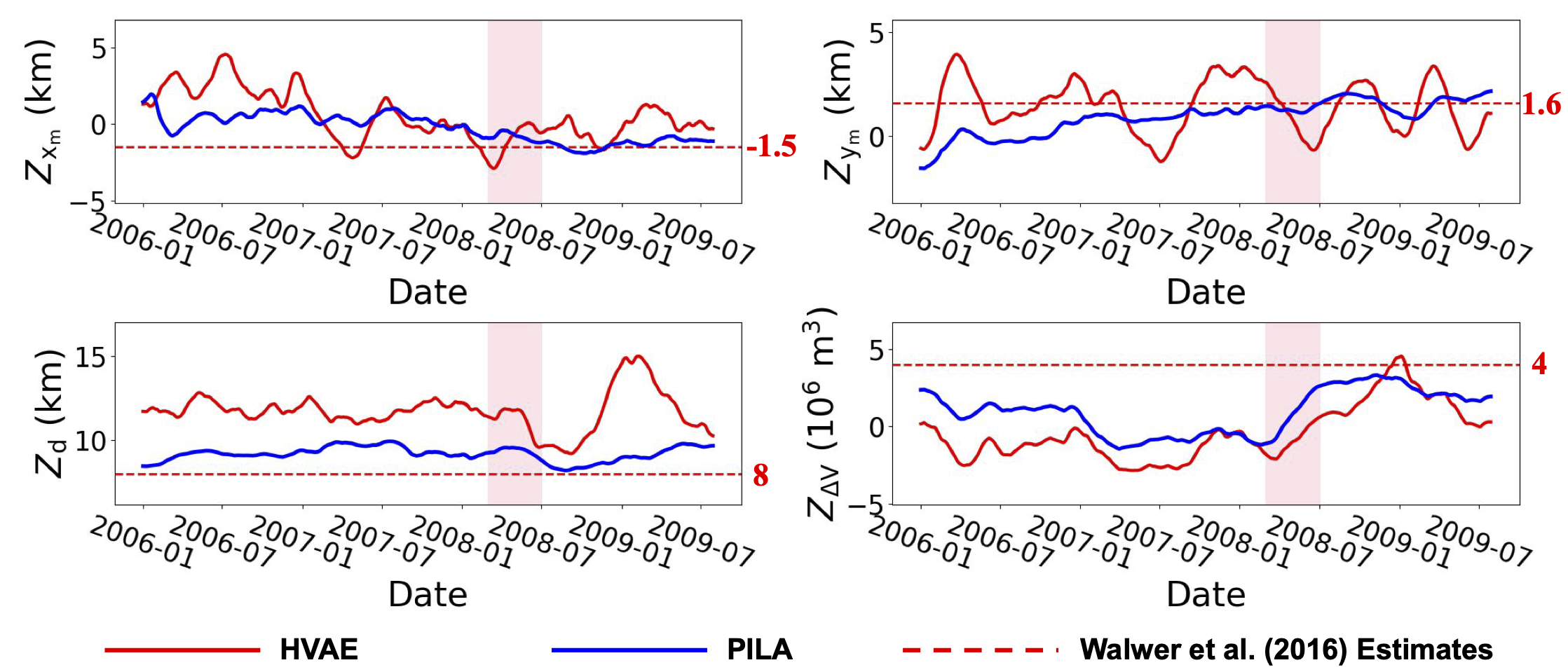}
    \caption{\textbf{PILA retrieves geophysical variables of volcanic deformation more accurately than HVAE does.} 
    \textit{PILA ($r=4$) recovers geophysical variables that capture the 2008 volcanic inflation event, with a relatively steady Mogi source located roughly beneath the volcano at a mean depth of 9.35 km; the volume change is about $3.7 \cdot 10^6~\text{m}^3$ over the inflation period, consistent with previous findings \citep{walwer2016data}. In contrast, the HVAE-estimated Mogi source location fluctuates significantly, and it estimates volume change at $2.7 \cdot 10^6~\text{m}^3$, only weakly capturing the 2008 event.}
    }
    \label{fig:mogi_inversion_variables}
\end{figure}

\begin{figure}[htbp]
    \centering
    \includegraphics[width=0.9\textwidth]{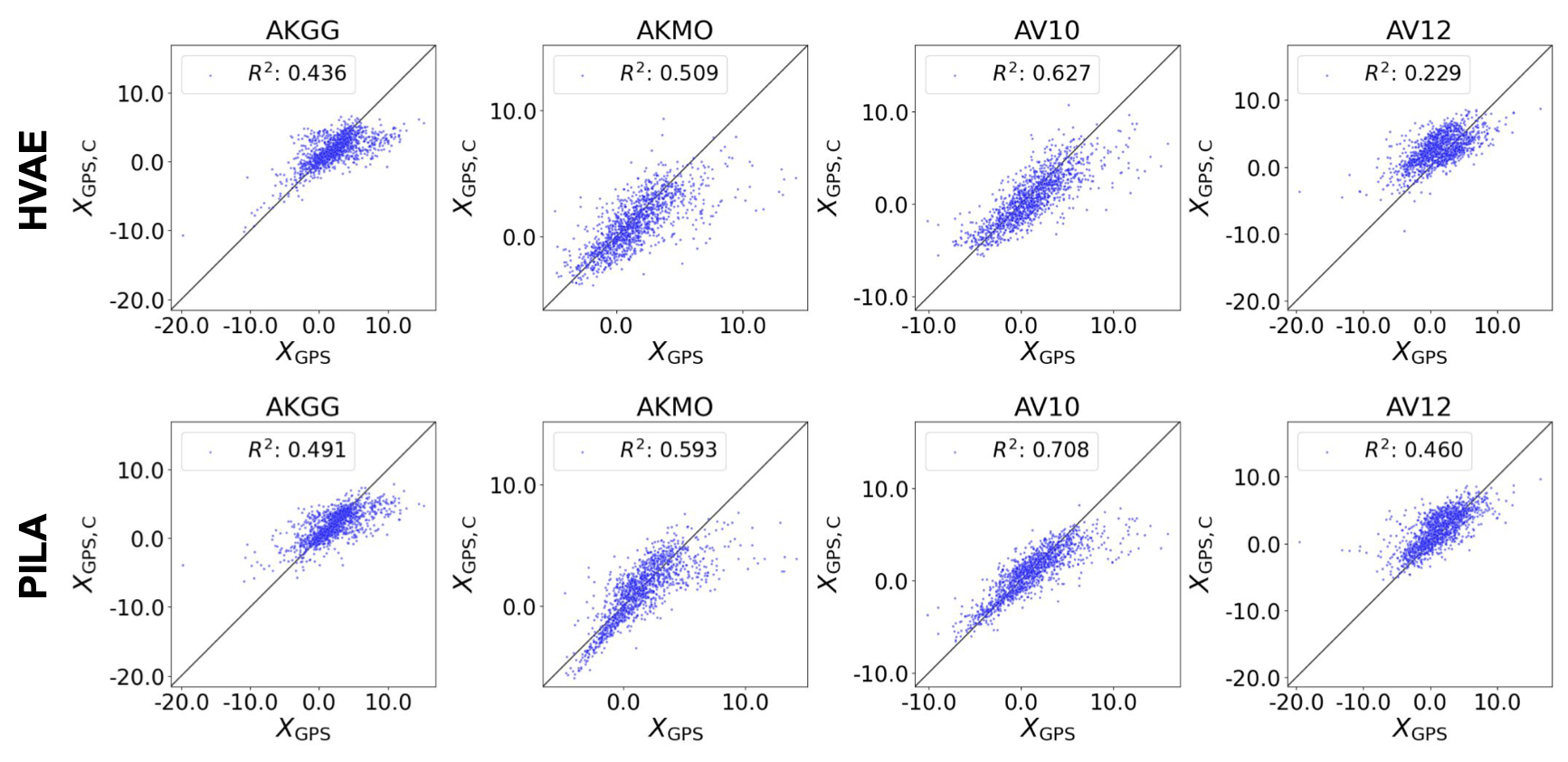}
    \caption{\textbf{PILA reconstruction accuracy of GNSS signals is maintained ($r=4$).} 
    \textit{Reconstruction accuracy is better than HVAE, despite the expected trade-off with more accurate recovery of physical variables (see \cref{fig:mogi_inversion_variables}).}
    }
    \label{fig:mogi_inversion_gnss}
\end{figure}

The temporal variation of the volume change ($Z_{\mathrm{\Delta V}}$) inferred by PILA (\cref{fig:mogi_inversion_variables}) successfully reproduces this major volcanic inflation event, with a total volume change of $3.7\cdot10^6~\text{m}^3$ at a mean depth ($Z_{\mathrm{d}}$) of 9.35 km, followed by a slow decrease of $Z_{\mathrm{\Delta V}}$ after this event. In addition, the horizontal location of the source, represented by $(x_{\mathrm{m}}, y_{\mathrm{m}})$, is roughly beneath the Akutan Volcano and remains relatively stable over time. These estimates are consistent with previous studies \citep{walwer2016data}, which used grid search optimisation to fit Mogi parameters to processed GNSS signals. 
In contrast, the horizontal position and source depth estimated by HVAE fluctuate significantly over time. The instability is likely due to the model incorrectly attributing seasonal components of $X_{\mathrm{GNSS}}$ to physical variables. Its predicted volume change (about $2.7\cdot10^6~\text{m}^3$) only weakly captures the volume increase of the magma chamber associated with the 2008 event (\cref{fig:mogi_inversion_variables}). 
In this case, PILA also achieves higher reconstruction accuracy than HVAE, though still imperfect compared to raw observations (\cref{fig:mogi_inversion_gnss}).

\begin{figure}[htbp]
    \centering
    \includegraphics[width=0.9\textwidth]{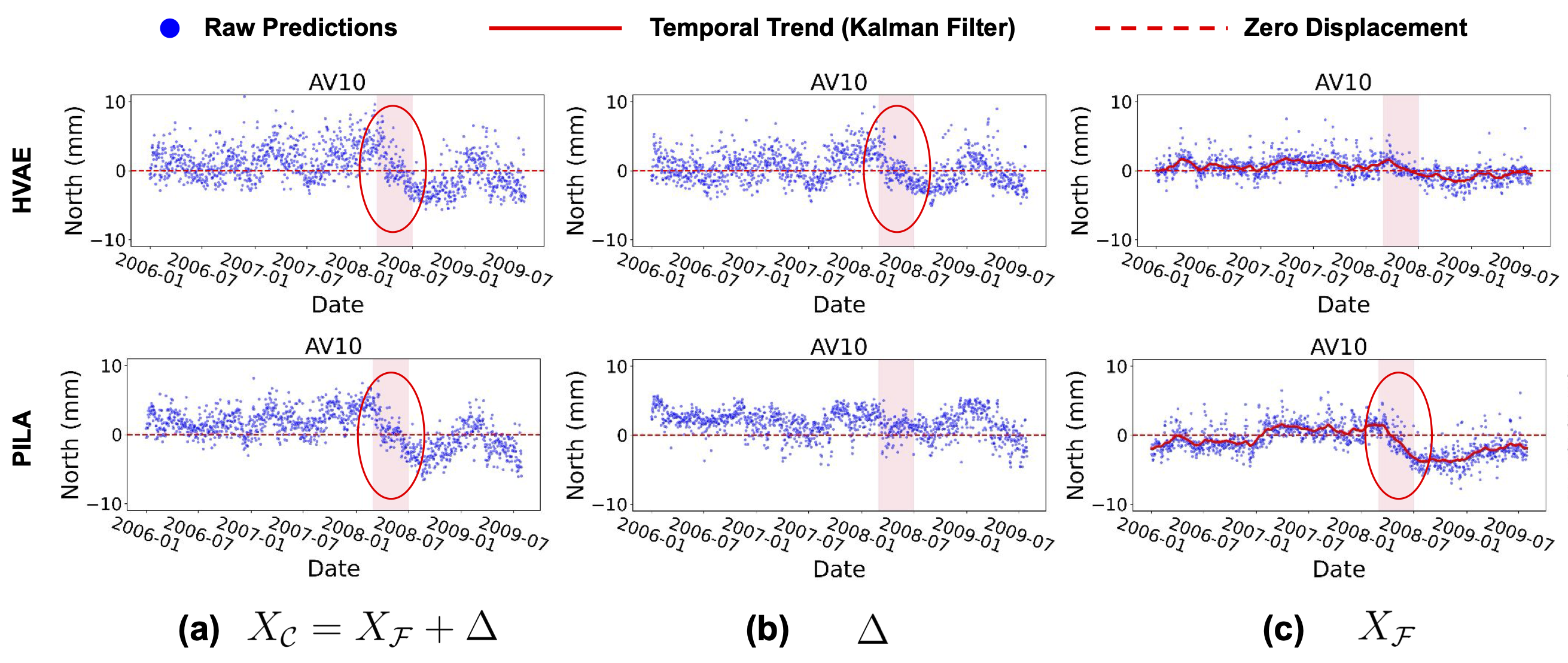}
    \caption{\textbf{PILA’s reconstruction of 2008 volcanic inflation signals clearly separate transient from seasonal effects.} 
    \textit{(a) The refined reconstruction $X_{\mathcal{C}}$ is similar for HVAE and PILA. However (b) in HVAE the transient signal is incorrectly absorbed into the residual ($\Delta$), whereas (c) PILA successfully captures the Mogi-source–induced transient, with an amplitude of about 5 mm over the inflation period.
    }
}
    \label{fig:mogi_inversion_gnss_transient_comparison_AV10}
\end{figure}

The steady location of the Mogi source and the estimated volume change from PILA clearly separate transient and seasonal effects. 
We examine reconstructions from both HVAE and PILA for the GNSS station AV10 (\cref{fig:mogi_inversion_gnss_transient_comparison_AV10}). Both models achieve comparable reconstructions after augmentations ($X_{\mathcal{C}}$). However HVAE incorrectly assigns transient effects to $\Delta$ from the auxiliary pathway, likely owing to its over-flexibility, whereas PILA assigns them to the physical reconstruction ($X_{\mathcal{F}}$) via the captured volume change ($Z_{\mathrm{\Delta V}}$). Its residual component ($\Delta$), learned via the PILA auxiliary pathway, mainly captures seasonal variations. 

\begin{figure}[htbp]
    \centering
    \includegraphics[width=0.9\textwidth]{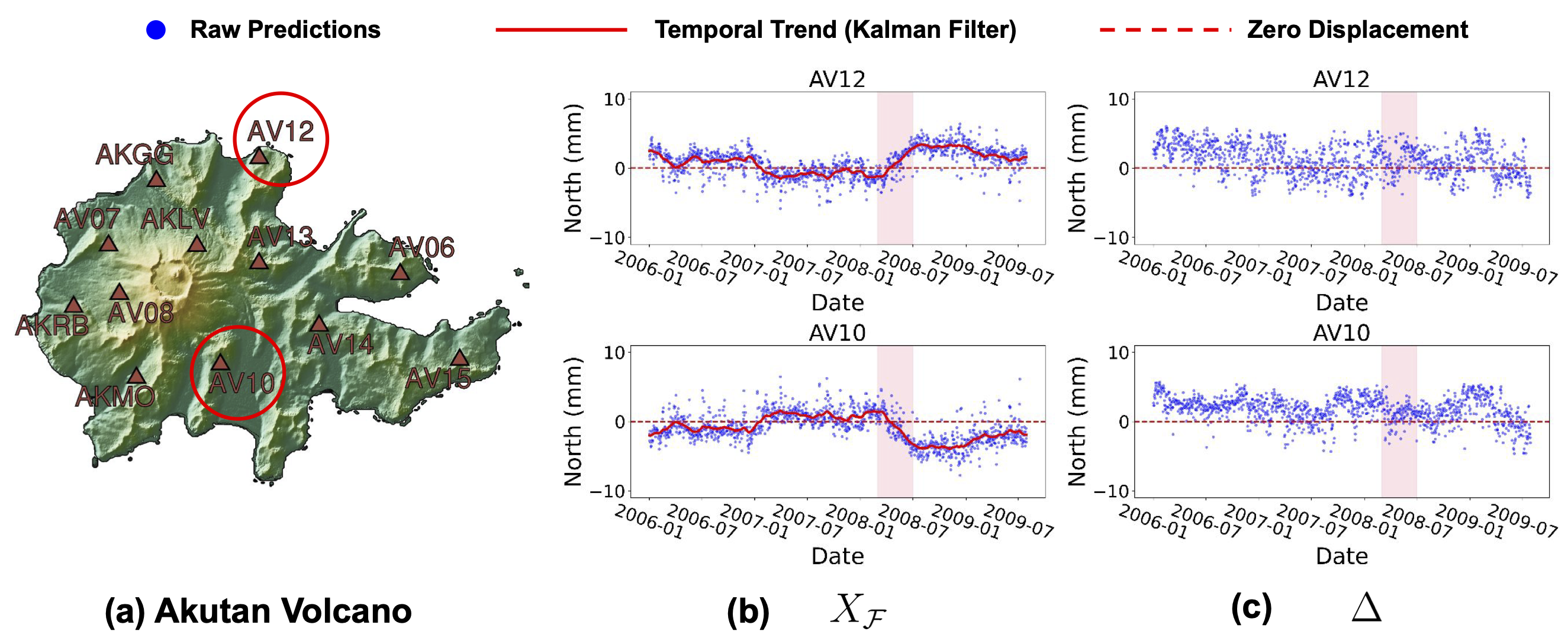}
    \caption{{
    \textbf{PILA reconstructions capture the 2008 volcanic inflation event, showing radial expansion of GNSS stations.} 
    \textit{For example, the two stations on opposite sides of the volcano show North/South displacements indicating radial expansion of about 5 mm over a ~4‑month period, starting in early 2008, followed by slow contraction due to volcanic deflation. This is consistent with previous findings \cite{walwer2016data}. Learned seasonal effects in $\Delta$ are broadly similar across both stations, as they should be.}}
}
    \label{fig:mogi_inversion_gnss_radial_expansion_AV10_AV12}
\end{figure}

Consequently PILA reconstruction characterises the radial expansion and uplift of the Earth's surface during the inflation. At GNSS stations AV10 and AV12, on opposite sides of the volcano, we compare the GNSS displacements reconstructed by $\mathcal{F}_{\mathrm{Mogi}}$ using the geophysical variables recovered from PILA (\cref{fig:mogi_inversion_gnss_radial_expansion_AV10_AV12}). The reconstructions show a radial expansion from March to July 2008, lasting about half a year, with displacements of ~5 mm, followed by a slow radial contraction, driven mainly by the volume change (\cref{fig:mogi_inversion_variables}), in agreement with previous studies \citep{ji2011transient,walwer2016data,shi2025fine}. In contrast to the transient signals represented by $X_{\mathcal{F}}$, the residual component ($\Delta$) learned by PILA remains similar across stations (\cref{fig:mogi_inversion_gnss_radial_expansion_AV10_AV12}, and \cref{fig:appx_mogi_inversion_gnss_transient_all_dirs_first_half}), as it should. When no inflation movement of the magma chamber occurs (e.g., in 2006 and in 2009), $X_{\mathcal{F}}$ is close to flat. After reconstruction with $\Delta$, $X_{\mathcal{C}}$ still captures seasonal displacements greater than 0 mm, reflecting the seasonal background signals present in the raw observation. 

Traditional inversion methods, like those in \citet{walwer2016data}, require GNSS signals to be specifically pre-processed to remove assumed seasonal components, followed by transient signal identification between stations, and numerical inversion based on grid search of the Mogi model. In contrast, in PILA these steps just emerge, jointly, from inference over the model. PILA simply captures transient signals, estimates Mogi parameters, and augments the physical reconstruction in a single model, without relying on additional filtering or strong assumptions about structures of the unmodeled processes. Inference is achieved by training on datasets that have no temporal overlap with the 2008 test period --- the learning-based approach proves to be capable of generalization.

\section{Discussion}
\subsection{Effect of residual and prior terms on model learning}\label{sec:ablations_residual_prior}
\begin{figure}[htbp]
    \centering
    \includegraphics[width=0.9\textwidth]{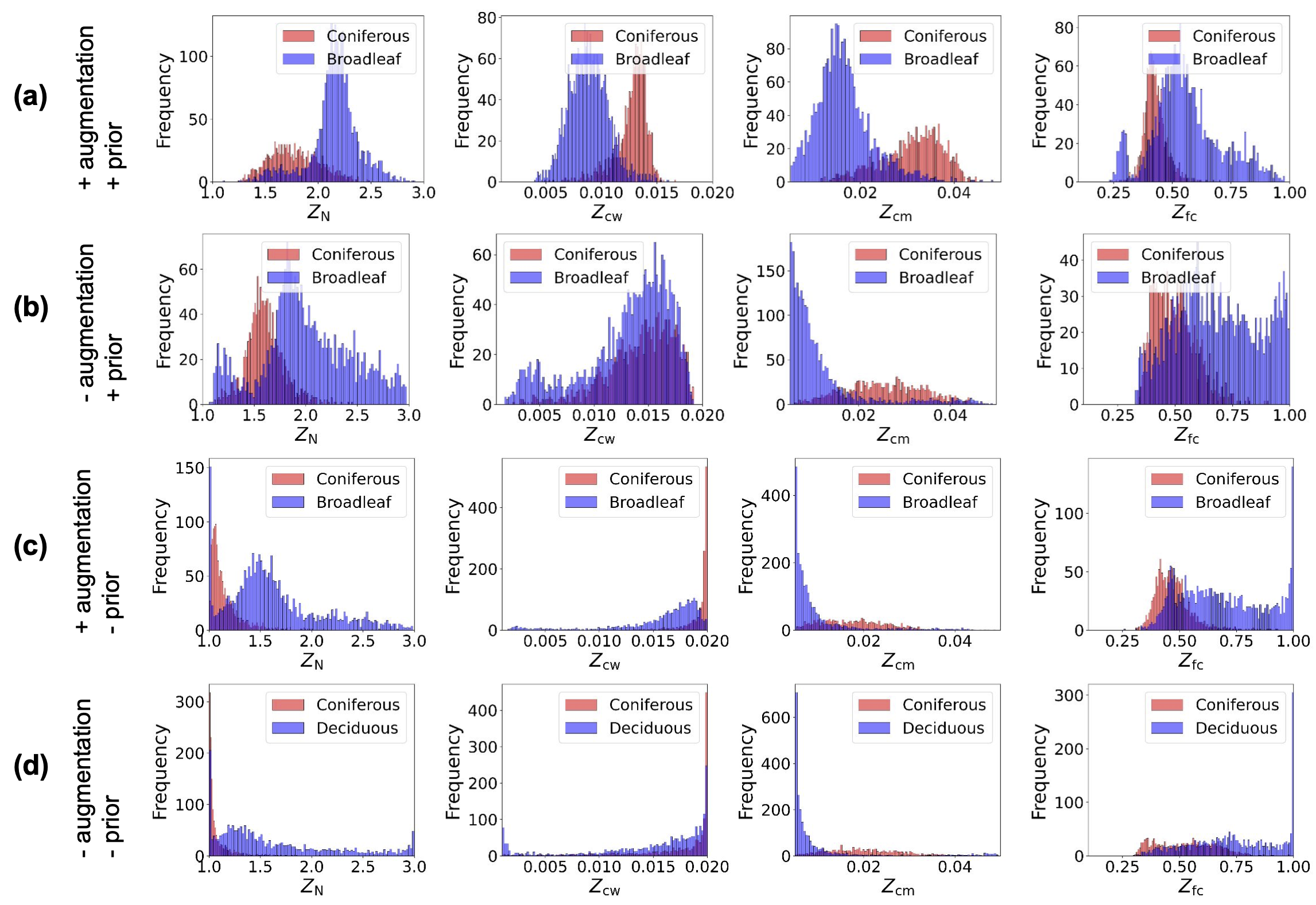}
    \caption{\textbf{In PILA, augmentation of the model, and application of a prior, are both important for plausible retrieval of the physical variables.} \textit{Ablations run on the RTM inversion of Austrian spectra: (a) PILA with both augmention and prior; (b) without augmentation, the prior alone does not ensure plausible retrieval of variables; (c) without the prior, augmentation alone also cannot plausibly retrieve variables; (d) removing also the end-stop prior, physical variables tend to be pushed to extreme values to minimise reconstruction loss.}}
    \label{fig:rtm_inversion_austria_ablations}
\end{figure}

In addition to reconstruction loss ($\mathcal{L}_{\mathrm{rec}}$, \cref{eq:loss_recontruction}), the objective function of PILA includes two key components: the residual term $\mathcal{L}_{\mathrm{res}}$ that augments the incomplete physical model (\cref{eq:loss_residual_basis}), and the prior term $\mathcal{L}_{\mathrm{prior}}$ that regularises the posterior distributions of $Z_{\mathrm{phy}}$ (\cref{eq:loss_prior_end_stop}). 
Taking the RTM inversion on the Austrian dataset as an example, we ablated each prior and residual loss term to see their impact on the learning outcome (\cref{fig:rtm_inversion_austria_ablations}). More details can be found in \cref{appx:ablations_residual_prior}. 

Without the residual term to compensate for model incompleteness, the distributions of learned variables tend to drift away from plausible ranges; frequently, predictions appear at boundary values, even though the prior is present (\cref{fig:rtm_inversion_austria_ablations} (b)). 
The prior alone is insufficient to recover plausible physical variables from an incomplete model.
When the prior term is also removed, there is no longer any regularisation in the space of $Z_{\mathrm{phy}}$ (\cref{fig:rtm_inversion_austria_ablations} (d)). All variables are distorted toward extremes as the model focuses exclusively on minimizing reconstruction loss.

Ablation of the prior term also highlights its role when just the residual term is present (\cref{fig:rtm_inversion_austria_ablations} (c)). Although the prior alone could not ensure plausible retrieval of the variables, in combination with the augmentation mechanism provided by the residual, it helps the model judge which part of the observation should be adjusted by the residual and which should be explained by the physical model. This balance supports a more stable and physically plausible retrieval. We further discuss the choice of prior in \cref{sec:prior}.

\subsection{Observability and model sensitivity of physical variables}\label{sec:sensitivity} 

Whether physical variables $Z_{\mathrm{phy}}$ can be successfully retrieved depends not only on the algorithm but also on their identifiability from raw observations $X$, and --- crucially in a physics-informed learning context --- on the sensitivity of the modeled observation $X_{\mathcal{F}}$ to changes in $Z_{\mathrm{phy}}$. More details can be found in \cref{appx:sensitivity}. 

\begin{figure}[htbp]
    \centering
    \includegraphics[width=0.9\textwidth]{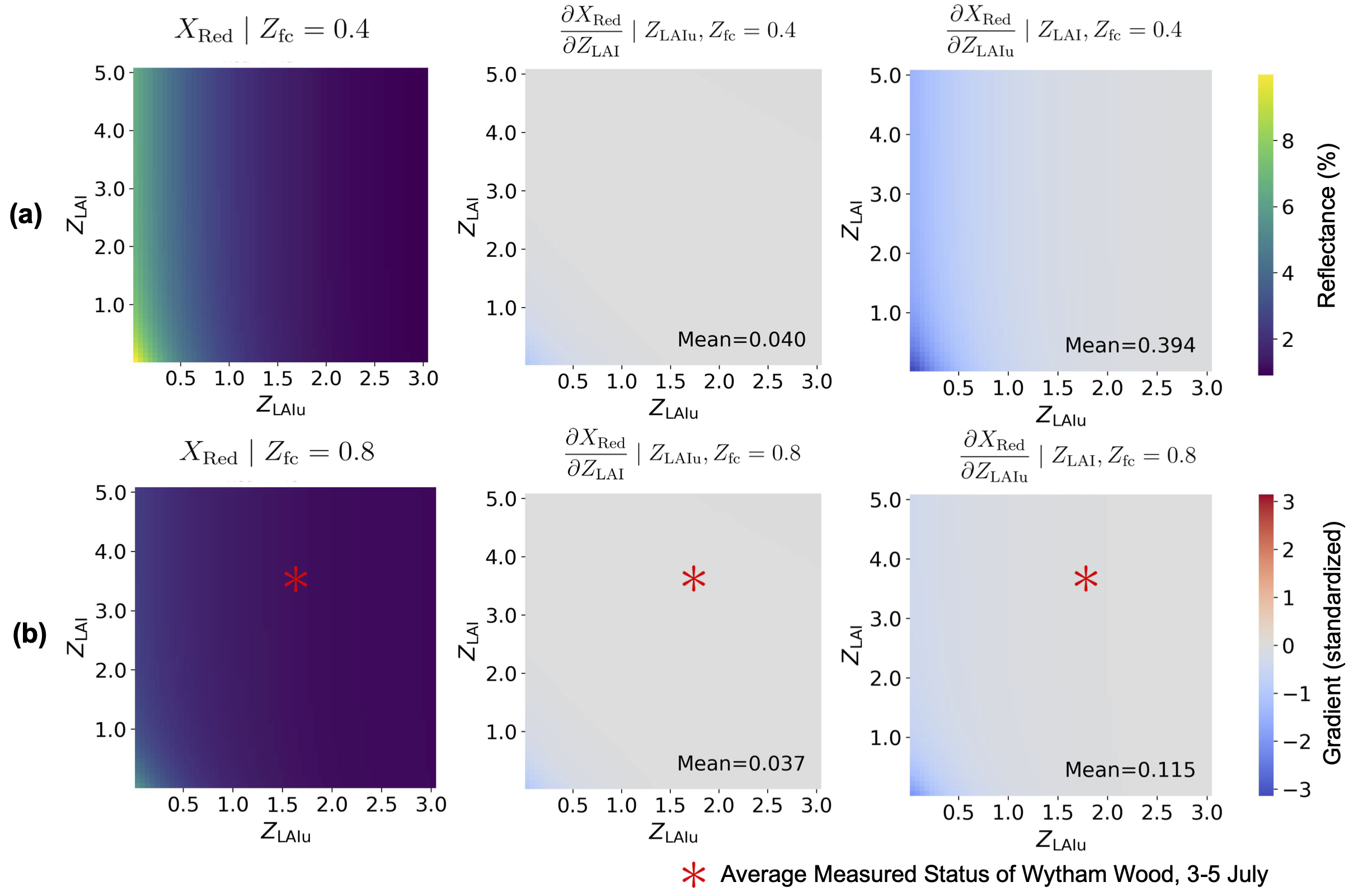}
    \caption{\textbf{Retrieval of physical variables depends not only on the learning framework, but also on their observability and the physical model’s sensitivity.} 
    \textit{This figure shows simulated red-band reflectance ($X_{\mathrm{Red}}$) as a function of canopy leaf area ($Z_{\mathrm{LAI}}$), understory leaf area ($Z_{\mathrm{LAIu}}$), and canopy cover ($Z_{\mathrm{fc}}$), along with standardized gradients for $Z_{\mathrm{LAI}}$ and $Z_{\mathrm{LAIu}}$. Both variables have similar gradient directions---greater leaf area lowers $X_{\mathrm{Red}}$---so their effects are hard to separate. In this RTM, (a) sensitivity to $Z_{\mathrm{LAIu}}$ is stronger than to $Z_{\mathrm{LAI}}$, but (b) it drops sharply as $Z_{\mathrm{fc}}$ increases from 0.4 to 0.8, and sensitivities of both variables approach zero under the high fractional coverage and dense canopy that characterized Wytham Wood during the measurement period (see \cref{tab:rtm_inversion_wytham_variables_insitu_july6}).}
    }
    \label{fig:rtm_inversion_variables_sensitivity_analysis_red_band}
\end{figure}

For the RTM, we study how $X_{\mathcal{F}}$ responds to changes in $Z_{\mathrm{LAI}}$ and $Z_{\mathrm{LAIu}}$ under different fractional coverages ($Z_{\mathrm{fc}}$) (\cref{fig:rtm_inversion_variables_sensitivity_analysis_red_band}). For each band, we compute the gradient of $X_{\mathcal{F}}$ with respect to one physical variable while keeping all others fixed. We then standardize these gradients so that their magnitudes match those encountered by the neural network during training. 
Taking the red band $X_{\mathrm{Red}}$ from the visible for example, the responses to $Z_{\mathrm{LAI}}$ and $Z_{\mathrm{LAIu}}$ show similar patterns (\cref{fig:rtm_inversion_variables_sensitivity_analysis_red_band}): increases in either variable lead to decreases in $X_{\mathrm{Red}}$. This behaviour is physically reasonable. Increasing leaf area results in greater absorption of visible light, thereby reducing reflectance in visible bands such as $X_{\mathrm{Red}}$. However, the similarity in response patterns for $Z_{\mathrm{LAI}}$ and $Z_{\mathrm{LAIu}}$ makes it difficult to disentangle these two variables—which represent leaf area at different canopy levels—from single-timestep optical spectra, as used in this study. 

Moreover, in the simulated outputs of $X_{\mathrm{Red}}$, they are more sensitive to the change of $Z_{\mathrm{LAIu}}$ than to $Z_{\mathrm{LAI}}$. As the fractional canopy coverage ($Z_{\mathrm{fc}}$) increases from a moderate (0.4) to a high level (0.8), the sensitivity of $X_{\mathrm{Red}}$ to both variables decreases—especially for $Z_{\mathrm{LAIu}}$. For instance, when $Z_{\mathrm{fc}}$ increases from 0.4 to 0.8, the mean of $\frac{\partial X_{\mathrm{Red}}}{\partial Z_{\mathrm{LAIu}}}$ decreases from 0.394 to 0.115. At $Z_{\mathrm{fc}}=0.4$, $X_{\mathrm{Red}}$ remains sensitive to $Z_{\mathrm{LAIu}}$ across the full range of $Z_{\mathrm{LAI}}$. However, when $Z_{\mathrm{fc}}=0.8$, this sensitivity becomes confined to regions where $Z_{\mathrm{LAI}}$ is relatively low. Patterns observed in other spectral bands are similar (see \cref{fig:appx_rtm_inversion_variables_sensitivity_analysis} in \cref{appx:sensitivity}). 

This behaviour is consistent with real-world canopy radiative transfer mechanisms. The reflectance of the understory vegetation reaches the satellite sensor mainly through canopy gaps or transmittance. When $Z_{\mathrm{fc}}=0.4$, more than half of the ground area consists of canopy gaps, so the understory contribution remains visible even under dense canopy layers (high $Z_{\mathrm{LAI}}$). However, when $Z_{\mathrm{fc}}$ increases to 0.8, only 20\% of the understory is visible directly. As canopy density increases (higher $Z_{\mathrm{LAI}}$), transmittance decreases, making it harder for the sensor to ``see through'' the canopy. Consequently, at high $Z_{\mathrm{fc}}$ and high $Z_{\mathrm{LAI}}$, the sensitivity of spectral bands to $Z_{\mathrm{LAIu}}$ approaches zero. 

This physical mechanism also explains the failure cases observed in variable retrieval. The similar response patterns of the spectral bands to both $Z_{\mathrm{LAI}}$ and $Z_{\mathrm{LAIu}}$ make disentanglement difficult. When $Z_{\mathrm{fc}}$ is moderate, the learning of $Z_{\mathrm{LAI}}$ collapses, while $Z_{\mathrm{LAIu}}$ still captures some variance due to its higher sensitivity. When fractional coverage becomes high and the canopy dense, the learning of both variables collapses due to lack of sensitivity---consistent with our ground validation in Wytham Woods, a broadleaf forest with high $Z_{\mathrm{fc}}$ and $Z_{\mathrm{LAI}}$ measured during summer. 

While the qualitative sensitivity patterns of our RTM align with physical intuition, the magnitude of these sensitivities—for example, the stronger response to $Z_{\mathrm{LAIu}}$ than to $Z_{\mathrm{LAI}}$, and the weaker sensitivity in near-infrared bands (\cref{appx:sensitivity})—depends on the accuracy of the physical model itself (\cref{appx:rtm_modelling}). Future work could consider more advanced RTMs with higher modeling capacity, such as 3D RTMs. Another avenue is to enhance variable retrieval by incorporating more informative observations, for instance through multi-sensor geo-embeddings that improve variable disentanglement and strengthen sensitivity.

\subsection{Choice of priors in physics-informed learning}\label{sec:prior}

In addition to an effective residual compensation mechanism, the choice of prior plays an important role. It imposes a preferred distribution for the physical variables and, during reconstruction, helps the algorithm judge which parts of the raw observation $X$ should be adjusted by the residual $\Delta$, and which parts should be explained by $X_{\mathcal{F}}$. 

As noted in \cref{sec:baseline_HVAE}, HVAE used a physical prior for all their study cases (pendulum, diffusion, etc.) with a mean centred inside the physical range and a somewhat arbitrary variance.
A similar situation occurs in \citet{zerah2024physics}, where an RTM is inverted using an autoencoder for mainly agricultural applications. Their setting assumes a relatively homogeneous landscape and does not require to compensate for model incompleteness; nonetheless they use a uniform prior over the physical range of $Z_{\mathrm{phy}}$. A uniform prior is not informative either, and they rely on multiple ablations to tune the KL term to enforce consistency between the posterior and this prior. 

\begin{figure}[htbp]
    \centering
    \includegraphics[width=0.9\textwidth]{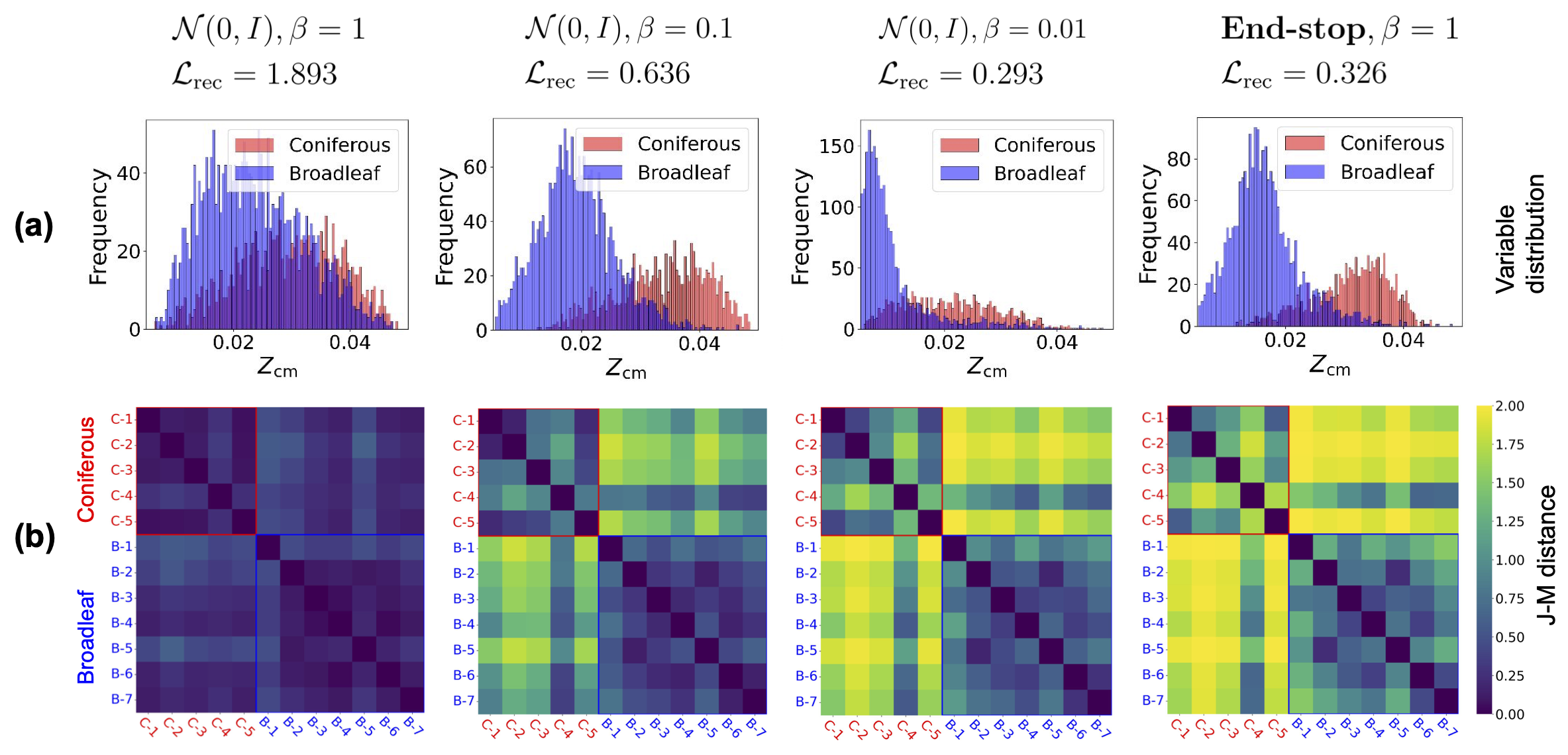}
    \caption{\textbf{An informative prior is often not available, and a simple but physically meaningful end-stop prior anyway requires less tuning.} 
    \textit{(a) Distributions of the retrieved Dry Matter ($Z_{\mathrm{cm}}$). (b) J-M distance between tree species using all retrieved variables. In the RTM inversion on the Austrian dataset, a standard normal prior that is too strong ($\beta=1$) prevents the model from learning physically meaningful patterns, while a very weak prior ($\beta=0.01$) produces implausible estimates, with many $Z_{\mathrm{cm}}$ retrievals at 0 (i.e., leaves without dry matter), in favour of minimizing reconstruction loss. In contrast, the simple end-stop prior achieves a good balance --- high reconstruction accuracy together with plausible physical variable retrievals.}
    }
    \label{fig:rtm_inversion_austria_prior_analysis}
\end{figure}

We also investigate how the choice of prior in PILA affects learning outcomes. A standard normal prior $\mathcal{N}(0,I)$ in the unbounded space $U_{\mathrm{phy}}$ is used, as it is more stable to train (see also \cref{appx:HVAE_formulations}). After applying a sigmoid function, this maps to the bounded space of $Z_{\mathrm{phy}}$. The corresponding KL regularisation term is defined as:
\begin{equation}\label{eq:kl_phy_standard_normal}
    \mathcal{L}_{\mathrm{prior}} = \mathcal{L}_\mathrm{KL}\!\left(q(U_{\mathrm{phy}}\mid X)\,\|\,\mathcal{N}(0,I)\right).
\end{equation}

In \cref{fig:rtm_inversion_austria_prior_analysis}, we change the weight $\beta$ applied to $\mathcal{L}_{\mathrm{prior}}$ as defined in \cref{eq:kl_phy_standard_normal}. Taking the learning of Dry Matter ($Z_{\mathrm{cm}}$) as an example, when $\beta=1$, the posterior estimation of $Z_{\mathrm{cm}}$ is strongly constrained by the prior $\mathcal{N}(0,I)$ (\cref{fig:rtm_inversion_austria_prior_analysis} (a)). This strong regularisation prevents the model from learning disentangled representations that align with the physical patterns of tree species and forest types (\cref{fig:rtm_inversion_austria_prior_analysis} (b)). 

As $\beta$ decreases from 1 to 0.1 and 0.01, the model achieves better reconstruction accuracy ($\mathcal{L}_{\mathrm{rec}}$) and more distinct physical patterns aligned with the relationships between tree species. However, this improvement comes at a cost: weaker regularisation makes it harder for the model to decide which parts of the observation should be adjusted by the residual and which should be explained by the physical variables. When $\beta=0.01$, some implausible estimates appear at the boundary values of $Z_{\mathrm{cm}}$, such as leaves predicted to exist without any substantial dry matter. 

In contrast, when we replace the Gaussian prior with a simple “end-stop” prior that merely encourages the model to avoid boundary values, the model achieves high reconstruction accuracy and plausible variable distributions with clear physical patterns. 

These experiments show that if the available prior is not physically informative but simply impose unit variance and independence across variables, as with $\mathcal{N}(0,I)$, $\beta$ requires sensitive tuning  to balance reconstruction accuracy with physical plausibility. A simple, physically interpretable prior (like the end-stop prior) can be more effective. As finer physical knowledge becomes available, the simple prior can readily be replaced by a more informative one as parameterized in \cref{eq:informative_prior}.

We further evaluate PILA's end-stop prior in HVAE and find that, although it improves the HVAE baseline, its performance remains inferior to that of PILA (\cref{appx:rtm_evaluation_greenhyperspectra}). Conversely, adding HVAE's synthetic-data regularisation (\cref{eq:pvae_loss_syn}) to PILA does not improve its performance (\cref{appx:syn_loss_on_pila}).

\subsection{Effects of rank on reconstruction accuracy and physical interpretability}\label{sec:rank}
\paragraph{Empirical analysis on real GNSS observations.} The residual rank ($r$) in PILA provides fine-grained control over the bias complexity, allowing augmentation of physical models of varying incompleteness. Here, we change the residual rank to evaluate its effect on reconstruction accuracy and the retrieval of geophysical variables. 

Since $\mathcal{F}_{\mathrm{Mogi}}$ is developed to explain only the transient signals caused by magma inflation events from a Mogi source, a large portion of the GNSS observations reflects other physical processes—such as tectonic motion and seasonal deformation driven by hydrological cycles (see \cref{appx:gnss_data_components} for the assumed GNSS components). It is therefore interesting to examine how changing the rank alters the capacity of $\Delta$ to model these non-volcanic signals, and how this affects the learning of Mogi parameters. 

\begin{figure}[htbp]
    \centering
    \includegraphics[width=0.9\textwidth]{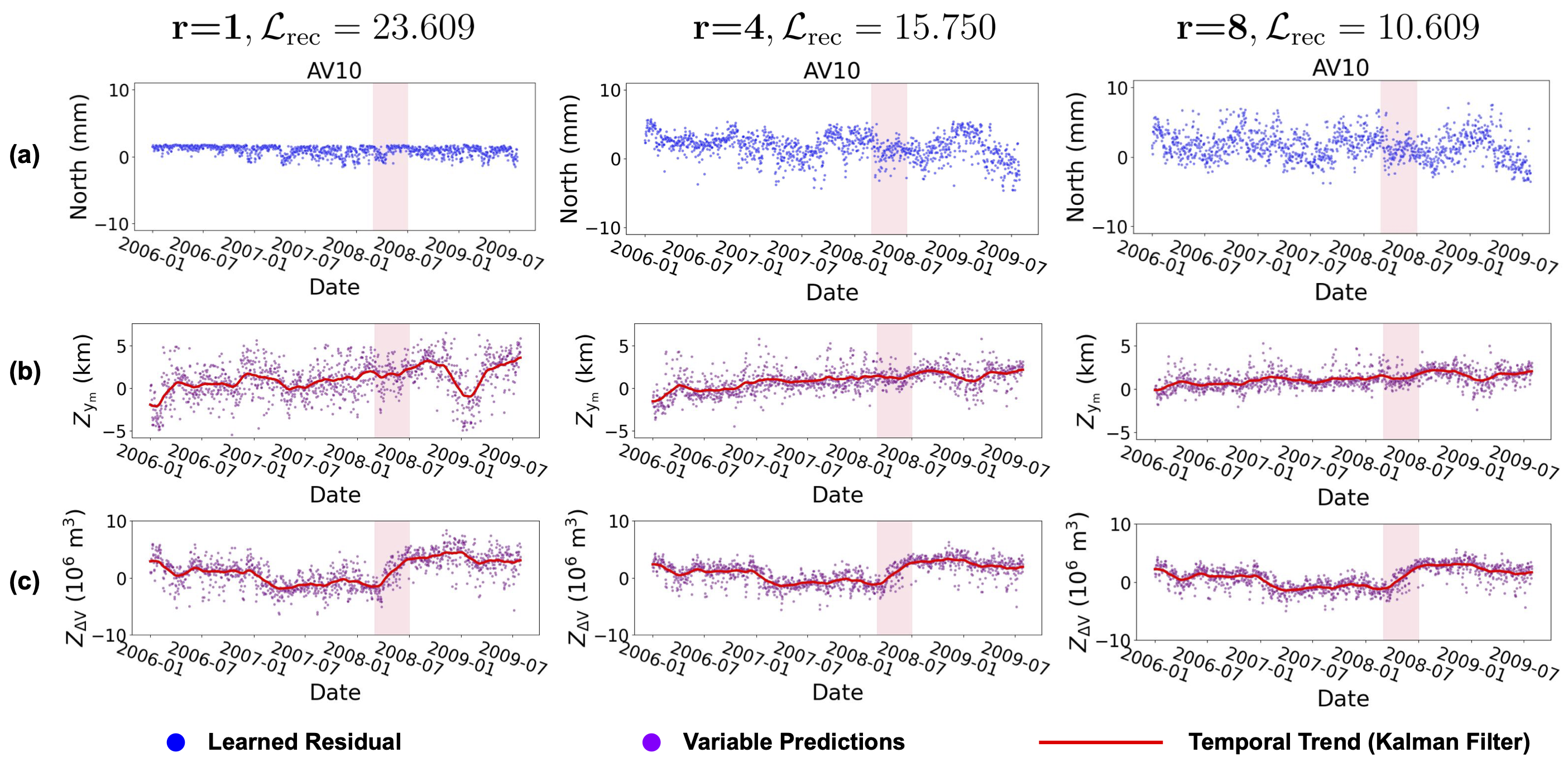}
    \caption{\textbf{Residual rank gives fine-grained control over the flexibility needed to accommodate the physical model's incompleteness.} 
    \textit{We change the rank of the residual in the Mogi inversion. Shown are (a) temporal variations of the learned residual $\Delta_{\mathrm{GNSS}}$ in the northward direction, (b) the retrieved coordinate of the horizontal location ($Z_{\mathrm{y_m}}$), which should remain stable over time, and (c) the retrieved volume change ($Z_{\Delta V}$), which should capture the 2008 inflation. When $r=1$, the model fails to capture seasonal components (a) not explained by the Mogi model; the reconstruction loss is high, and the retrieved horizontal location fluctuates heavily (b). As rank increases, the residual captures more of the seasonal variation ($r=4$) and even finer details ($r=8$). Consequently, the horizontal location becomes stable and $Z_{\Delta V}$ in (c) more clearly characterizes the 2008 inflation event.}
    }
    \label{fig:mogi_inversion_rank_analysis}
\end{figure}

In \cref{fig:mogi_inversion_rank_analysis}, we increase the rank used in Mogi inversion from the minimal setting ($r=1$) to higher values. When $r=1$, the learned $\Delta$ barely captures seasonal changes, and the reconstruction error is high ($\mathcal{L}_{\mathrm{rec}}=23.609$). Consequently, the predicted location (e.g., $Z_{\mathrm{y_m}}$) is noisy and exhibits an artificial seasonal trend. This suggests that when the residual capacity is too limited, the physical model overfits to explain the missing variability, leading to distorted physical variable estimates. 

As the rank increases from 1 to 4, the residual becomes capable of capturing seasonal components, and reconstruction accuracy improves substantially. At $r=8$, the residual captures additional nuances of the non-transient signals, further improving reconstruction. As a result, the predicted location of the Mogi source becomes smoother and more stable over time (\cref{fig:mogi_inversion_rank_analysis} (b)), while the inferred volume change continues to capture the 2008 inflation event but with less temporal noise (\cref{fig:mogi_inversion_rank_analysis} (c)). 

In the experiments presented above, we chose a basic rank $r=2$ for the relatively complete RTM and $r=4$ for the less complete Mogi model. These are empirical settings that work well in our inversions but are not necessarily optimal. Future studies could consider model selection metrics, like the Bayesian or Akaike Information Criterion, to identify the rank that best balances model complexity and reconstruction accuracy, for specific applications.

\paragraph{Analysis with synthetic data in a controlled setting.}
We further use synthetic data in a controlled setting to investigate the trade-off between reconstruction accuracy and physical interpretability.
To isolate the effect of PILA's low-rank residual from the case-specific structure of the real GNSS data, we construct synthetic observations, $X' = X_{\mathcal{F}} + \Delta'$, in which the generated residual $\Delta'$ has a precisely controlled intrinsic rank $r'$. 
The spatial components of $\Delta'$ are constructed to be orthogonal to the Mogi displacement direction, making the physical and residual components separable by construction, although this assumption excludes more challenging cases where residuals overlap with the physical signal.
We then apply PILA using different residual ranks $r$ to examine their effects on reconstruction and physical-variable retrieval. Further details of the data-generation process, including the construction of $\Delta'$, are provided in \cref{appx:gnss_synthetic_data}.

\begin{figure}[htbp]
    \centering
    \includegraphics[width=0.9\textwidth]{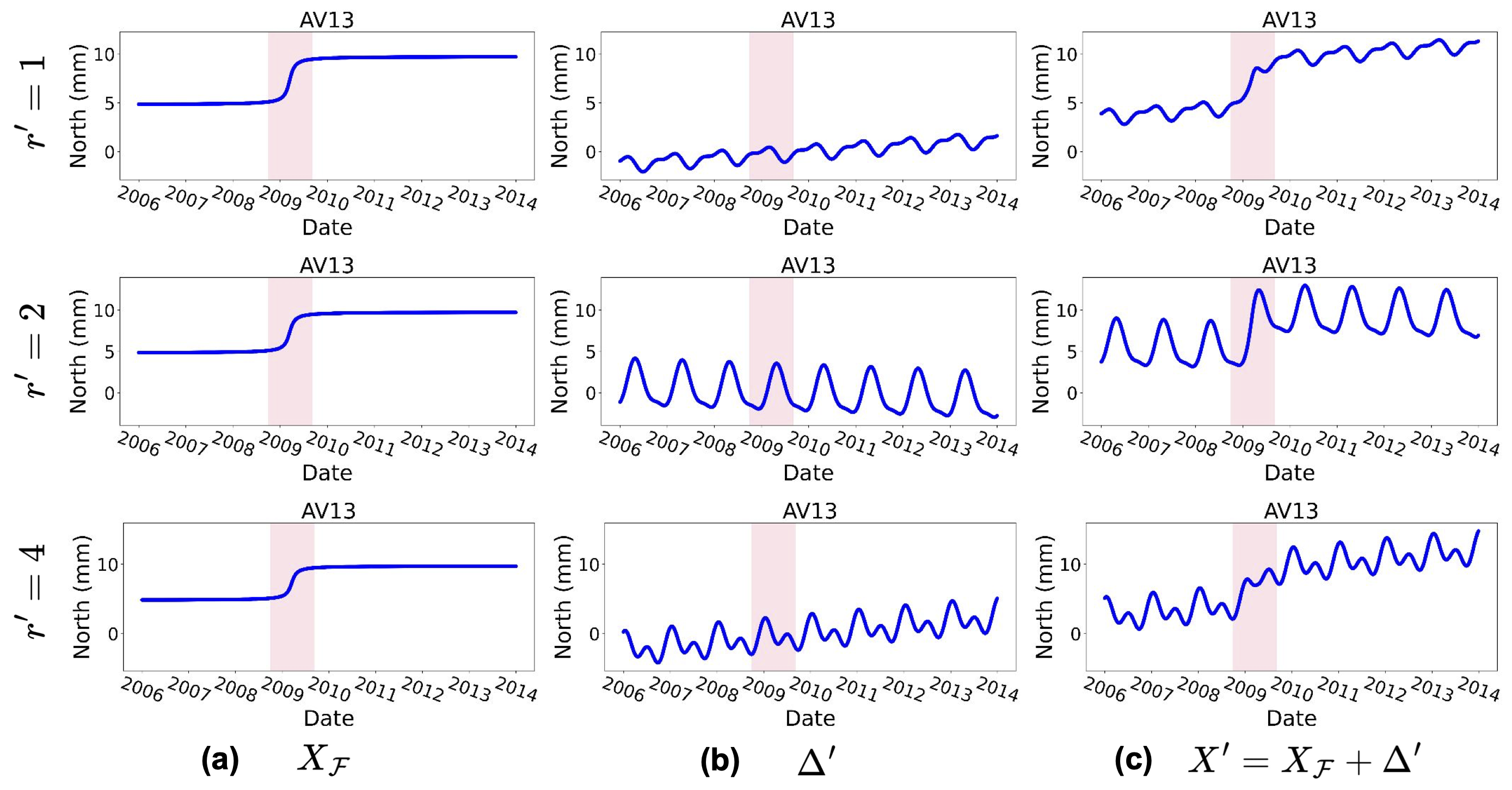}
    \caption{
    \textbf{Synthetic control enables the generation of GNSS observations by combining Mogi-induced displacements with systematic residuals of a prescribed intrinsic rank $r'$.}
    \textit{Synthetic residuals are generated with intrinsic ranks $r'\in\{1,2,4\}$. For illustration, we show the northward displacement time series at station AV13. The pink area highlights the main inflation period.}
    }
    \label{fig:rebuttal_synthetic_control_data}
\end{figure}

We generate a time series of physical-model outputs $X_{\mathcal{F}}$ and residuals $\Delta'$ with ground-truth ranks $r' \in {1,2,4}$.
This produces three synthetic datasets whose residuals have different known intrinsic ranks (\cref{fig:rebuttal_synthetic_control_data}).

For each dataset, we apply PILA while varying the rank of its data-driven residual $\Delta$ over $r\in\{1,2,4,8\}$. The cases $r<r'$, $r=r'$, and $r>r'$ correspond to under-parameterised, correctly specified, and over-parameterised residual models, respectively. By comparing different combinations of the intrinsic residual rank $r'$ and the residual rank $r$ used by PILA, we investigate how the capacity of the data-driven residual influences both reconstruction accuracy and physical interpretability. 
We jointly invert all four Mogi parameters, $(x_m,y_m,d,\Delta V)$, following the same configurations specified in \cref{tab:mogi_para_list}.

We evaluate reconstruction accuracy using the root mean squared error (RMSE) between the input $X'$ and the augmented reconstruction $X'_{\mathcal C}$. 
Physical interpretability is assessed using the prediction error of the retrieved volume change $\Delta V$. 
Since the magnitude of $\Delta V$ is on the order of $10^6$, we report its normalised RMSE (NRMSE), obtained by normalising with the range of the ground-truth $\Delta V$.

\begin{figure}[htbp]
    \centering
    \includegraphics[width=0.9\textwidth]{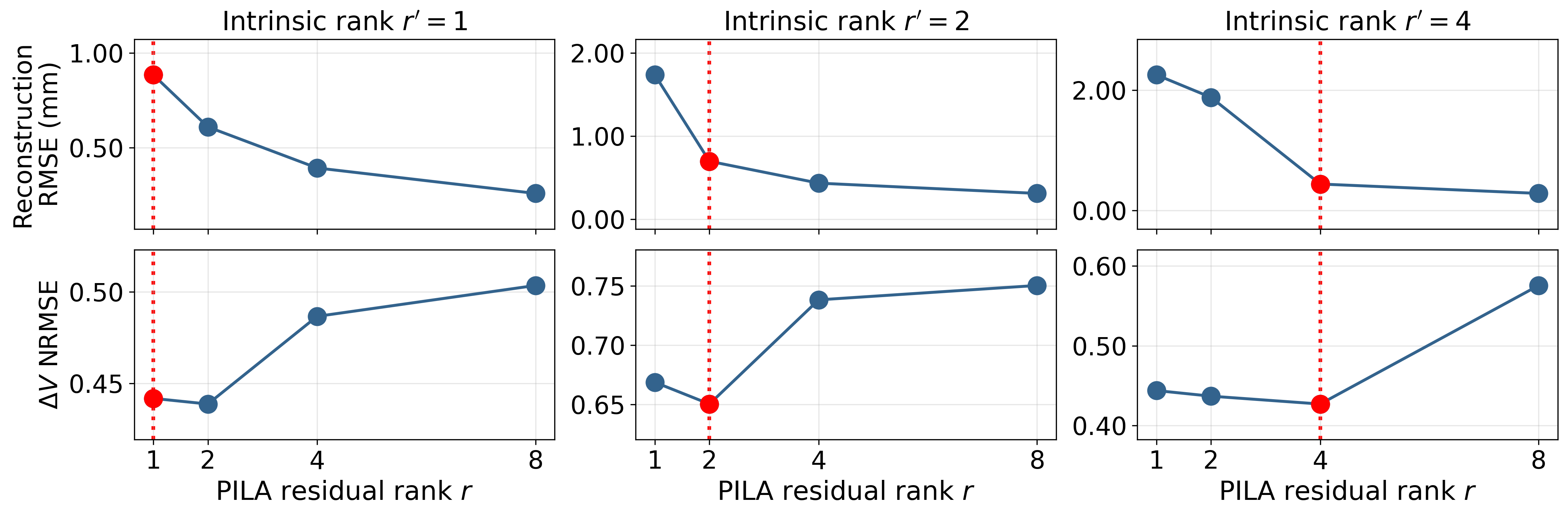}
    \caption{
    \textbf{Reconstruction accuracy continues to improve with PILA's increasing residual rank $r$, whereas physical interpretability is maximised when the residual rank $r$ is at or near the ground-truth intrinsic rank $r'$.}
    \textit{Synthetic datasets are generated with intrinsic residual ranks $r'\in\{1,2,4\}$ and inverted using PILA with residual ranks $r\in\{1,2,4,8\}$. Lower values indicate better performance. The red dotted line and point denote the matched-rank setting ($r=r'$).}
    }
    \label{fig:rebuttal_synthetic_control_rank_sweep}
\end{figure}

As shown in \cref{fig:rebuttal_synthetic_control_rank_sweep}, the reconstruction RMSE decreases monotonically as $r$ increases, regardless of the intrinsic rank $r'$ of $\Delta'$. This indicates that increasing the residual rank consistently improves reconstruction accuracy by providing greater flexibility to compensate for the incomplete physical model.
In contrast, the $\Delta V$ NRMSE exhibits an elbow-shaped trend. When $r<r'$, the data-driven residual lacks sufficient capacity to model all modes of the systematic mismatch, leading to poorer recovery of the underlying physical signal. As $r$ approaches the intrinsic rank $r'$, both metrics reach their lowest values or remain very close to the optimum. Increasing the residual rank beyond this point continues to improve reconstruction accuracy but can degrade physical interpretability, as the $\Delta V$ prediction error increase. This suggests that the additional residual capacity begins to model unnecessary degrees of freedom rather than the true systematic mismatch. 

These controlled experiments demonstrate that the residual rank $r$ acts as an effective tuning knob for balancing reconstruction accuracy and physical interpretability. While increasing $r$ consistently improves reconstruction accuracy, the most accurate recovery of the underlying physical process is achieved when the residual rank is selected at or near the intrinsic rank of the model mismatch. 

%% file: 5_conclusion.tex
\section{Conclusion}
\label{sec:conclusions}

We proposed PILA---physics-informed low-rank augmentation---to invert  physical models with different degrees of incompleteness. We tried models arising from representative inverse problems in EO: radiative transfer in optical remote sensing for inferring surface biophysical status, and volcanic deformation observed by GNSS for inferring subsurface geophysical status. Experiments on real-world data show that PILA retrieves physical variables more accurately and with better disentanglement than the state-of-the-art HVAE. 
We analyzed the effects of each regularisation term in the learning objective and discussed key aspects of inverse problems, including variable observability, model sensitivity, and choice of physical prior. 
Using synthetic data generated in a controlled setting, we further investigated the trade-off between reconstruction accuracy and physical interpretability and demonstrated the effectiveness of PILA's low-rank augmentation design.
We believe PILA is broadly applicable to inverse problems involving incomplete physical models, and that our EO demonstrations will promote physically meaningful interpretation of our planet in an era of abundant EO data.

%% file: 6_appendix.tex
\section{Appendix}\label{sec:appx}
\subsection{Making a fully differentiable RTM for scientific machine learning}\label{appx:making_diff_rtm}
Many legacy implementations of physical models in EO were originally designed for forward simulation, without considering differentiability or compatibility with backpropagation for machine learning. Although implementing a differentiable Mogi model is fairly straightforward due to its simplicity, the RTM modules we rely on ($\mathcal{F}_{\mathrm{RTM}}$) were taken from the \href{https://enmap-box-lmu-vegetation-apps.readthedocs.io/en/latest/tutorials/IVVRM_tut.html}{EnMap} application within the QGIS software. These modules were implemented using NumPy arrays and operations.  
To record the computational graph and support gradient backpropagation, we reimplemented the model using PyTorch operations. However, $\mathcal{F}_{\mathrm{RTM}}$ is not explicitly differentiable and represents a complex physical model—its original code base spans 1,742 lines across multiple scripts that simulate radiative transfer at various canopy levels.  
Converting standard NumPy operations into PyTorch is generally uncomplicated, but some operations are not directly differentiable and require custom handling:  
Below is an example code snippet from the original implementation:
\begin{verbatim}
n = PD_refractive
k = (np.outer(Cab, PD_k_Cab) + np.outer(Car, PD_k_Car) + np.outer(Anth, 
    PD_k_Anth) + np.outer(Cbrown, PD_k_Brown) 
    + np.outer(Cw, PD_k_Cw) + np.outer(Cm, PD_k_Cm)) / N[:, np.newaxis]

ind_k0_row, ind_k0_col = np.where(k == 0)

if len(ind_k0_row) > 0:
    k[ind_k0_row, ind_k0_col] = np.finfo(float).eps
trans = (1 - k) * np.exp(-k) + (k ** 2) * exp1(k)
\end{verbatim}

It involves the exponential integral function $E_1(Z)$ (referred to as \texttt{exp1} in the code example), which is given by the following integral:
\begin{equation}\label{eq:E1}
E_{1}(z) = \int_{z}^{\infty} \frac{e^{-t}}{t} dt
\end{equation}

and the derivative of $E_1(z)$ with respect to its argument $z$ can then be obtained as:
\begin{equation}\label{eq:E1_derivative}
\frac{d}{dz} E_{1}(z) = -\frac{e^{-z}}{z}
\end{equation}

Note that the derivative is not defined at $z=0$, and there is no built-in implementation of $E_{1}(z)$ in PyTorch. Therefore, we rely on \texttt{torch.from\_numpy} to access NumPy’s \texttt{exp1}, and we implement custom backward methods to specify the gradients.
The corresponding PyTorch version of the example code is shown below:
\begin{verbatim}
n = PD_refractive
k = (torch.outer(Cab, PD_k_Cab) + torch.outer(Car, PD_k_Car) 
    + torch.outer(Anth, PD_k_Anth) + torch.outer(Cbrown, PD_k_Brown) 
    + torch.outer(Cw, PD_k_Cw) + torch.outer(Cm, PD_k_Cm)) / N.unsqueeze(-1)

ind_k0_row, ind_k0_col = torch.where(k == 0)

if len(ind_k0_row) > 0:
    k[ind_k0_row, ind_k0_col] = torch.finfo(float).eps
trans = (1 - k) * torch.exp(-k) + (k ** 2) * torch.from_numpy(exp1(k.numpy()))
\end{verbatim}
Additionally, it should be emphasized that the structural parameter \texttt{N} appears in the denominator of the computation and therefore cannot be zero, which is consistent with its physically meaningful domain of strictly positive values.

To verify the PyTorch implementation of $\mathcal{F}_{\mathrm{RTM}}$ against its reference implementation in NumPy, we randomly generated 10,000 sets of input variables, evaluated both implementations on these inputs, and compared the resulting outputs (\cref{tab:conversion_unittest}). With an absolute tolerance of 1e-5, the output mismatch rate is 0.457\%. The maximum absolute deviation across all 130,000 output reflectance values between the two implementations is 3.050e-5, where the physical unit of reflectance is 1. These results demonstrate that the PyTorch implementation of $\mathcal{F}_{\mathrm{RTM}}$ is effectively equivalent to the original NumPy version and is thus suitable for use in subsequent tasks. Furthermore, the PyTorch-based RTM can be readily integrated into other deep-learning workflows. 
 
\begin{table*}[htbp] 
\caption{\textbf{Unit test comparing the outputs of our PyTorch implementation of $\mathcal{F}_\mathrm{RTM}$ with the original NumPy version.} \textit{We compare 130000 elements, obtained from 10000 simulated samples with 13 Sentinel-2 spectral bands in both implementations.}}
\label{tab:conversion_unittest}
\centering
\resizebox{\textwidth}{!}{
    \begin{tabular}{ccccc}
    \toprule
    Total Elements & Mismatched Elements & Absolute Tolerance & Mismatch Ratio & Max Absolute Difference \\
    \midrule
    130000 &  594 &  1e-5 &  0.457\% &  3.050e-5 \\
    \bottomrule
    \end{tabular}
}
\end{table*}

\subsection{Integrating the differentiable RTM into the autoencoder}\label{appx:bypassing_instability}
\paragraph{Forward pass}
By definition, the input variables for $\mathcal{F}_{\mathrm{RTM}}$ are non-negative quantities with physical interpretations (\cref{tab:inform_para_list}). In practice, however, $\mathcal{F}_{\mathrm{RTM}}$ may fail during the forward pass when variable values are outside their intended ranges (for example, the structure parameter \texttt{N} in the code snippet above). The same issue arises for $\mathcal{F}_{\text{Mogi}}$.  
Therefore, for both models, we first infer normalized variables $\eta \in [0, 1]$ and then map them to variables $Z \in [Z_{\text{min}}, Z_{\text{max}}]$ in their original physical units according to the specified value ranges (\cref{tab:inform_para_list,tab:mogi_para_list}).

\begin{equation}\label{eq:rescaling}
    Z = (Z_{\text{max}} - Z_{\text{min}}) \cdot \eta + Z_{\text{min}}
\end{equation}

\paragraph{Backpropagation}
Although we successfully encoded the forward pass, backpropagation through $\mathcal{F}_{\mathrm{RTM}}$ was unstable. For effective learning, gradients must pass through the physical model, but $\mathcal{F}_{\mathrm{RTM}}$ includes many operations (exponentials, logarithms, square roots) that, while differentiable in theory, often cause numerical instabilities when differentiating. These issues produced \texttt{NaN} gradients and consequently \texttt{NaN} losses.

When attempting to stabilize training, we found that the first \texttt{NaN} gradients, at the moment they first emerge, consistently coincide with specificsame groups of variables within $\mathcal{F}_{\mathrm{RTM}}$. In principle, one could constrain the specific operations in $\mathcal{F}_{\mathrm{RTM}}$ that generate these \texttt{NaN}s, but this would require deep knowledge of the physical formulation and its differentiability, which is a nontrivial undertaking.

Instead, we used a simple workaround: because the \texttt{NaN} gradients themselves disrupt learning, we intercept them at their first occurrence after backpropagation through $\mathcal{F}$ and replace them with small random values drawn from a uniform distribution on [0, 1] and scaled by 1e-7. The detailed algorithm is given in \cref{algo:grad_stab}.

\begin{algorithm}
\caption{Gradient stabilization}\label{algo:grad_stab}
    \begin{algorithmic}
    \STATE Calculate gradients    
    \STATE Initialize list $grads$, containing gradients of all model parameters where gradients exist and contain any NaN values
    
    \IF{$grads$ is not empty}
        \STATE Set a small constant epsilon equal to 1e-7

        \FOR{each v in $grads$}
            \STATE Generate random values of the same shape as v, scaled by epsilon
            \STATE Create a mask where v is NaN or equals 0
            \STATE Replace values in v where the mask is True with corresponding random values
        \ENDFOR
    \ENDIF
    \STATE Update gradients    
    \end{algorithmic}
\end{algorithm}

This simple fix stabilised learning in the autoencoder when integrating the RTM, despite the instability in $\mathcal{F}_{\mathrm{RTM}}$ during backpropagation. By blocking \texttt{NaN} gradients from reaching earlier layers while keeping the forward pass intact, the stabiliser lets training bypass instability points during optimisation and continues searching for optimal solutions.

\subsection{Biophysical variables of the RTM}\label{appx:rtm_vars}
The input to $\mathcal{F}_{\mathrm{RTM}}$ comprises biophysical variables at three hierarchical levels (\cref{tab:para_overview}). The original INFORM RTM does not use fractional coverage $Z_{\mathrm{fc}}$ as a direct input. We add $Z_{\mathrm{fc}}$ as one of seven variables predicted by the encoder and then use it to infer crown diameter $Z_{\mathrm{cd}}$ and height $Z_{\mathrm{h}}$ via derived equations, reducing the ill-posedness of the inverse problem.
Because fractional coverage is jointly determined by stem density and crown diameter within each hectare (10,000 $m^2$), $Z_{\mathrm{cd}}$ can be derived from $Z_{\mathrm{fc}}$ and stem density using \cref{eq:fc2cd}. To derive $Z_{\mathrm{h}}$, we fit an allometric equation (\cref{eq:cd2h}) using temperate forest samples from the global allometric database \citep{jucker2017allometric}. The resulting relationship between $Z_{\mathrm{h}}$ and $Z_{\mathrm{cd}}$ has $R^2 = 0.383$ (\cref{eq:cd2h}).

\begin{equation}\label{eq:fc2cd}
    Z_{\mathrm{cd}} = 2 \cdot \sqrt{\frac{Z_{\mathrm{fc}} \cdot 10000}{\pi \cdot Z_{\mathrm{sd}}}}
\end{equation}

\begin{equation}\label{eq:cd2h}
    Z_{\mathrm{h}} = \exp\left(2.117 + 0.507 \cdot \ln(Z_{\mathrm{cd}})\right)
\end{equation}

\begin{table*}[t] 
\caption{\textbf{Overview of the biophysical variables of $\mathcal{F}_{\mathrm{RTM}}$.} \textit{These variables can be attributed to three hierarchical levels. 7 variables are learned directly. \textbf{*} $Z_{\mathrm{cd}}$ and $Z_{\mathrm{h}}$ will be inferred from $fc$ using \cref{eq:fc2cd} and \cref{eq:cd2h}, respectively.}}
\label{tab:para_overview}
\centering
\resizebox{\textwidth}{!}{
    \begin{tabular}{lcccccc}
    \toprule
    \multirow{2}*{Group} & \multirow{2}*{Variable} & \multirow{2}*{Acronym} & \multirow{2}*{To Learn} & \multirow{2}*{Default Value} & \multicolumn{2}{c}{Sample Range} \\
    \cmidrule(lr){6-7}
    ~ & ~ & ~ & ~ & ~ & Min & Max \\
    \midrule
    Background & Soil brightness factor & psoil & \ding{55} & 0.8 & - & - \\
    \midrule
    \multirow{9}*{Leaf Model} & Structure Parameter & N & \ding{51} & - & 1 & 3 \\
    ~ & Chlorophyll A+B & cab & \ding{51} & - & 10 & 80 \\
    ~ & Water Content & cw & \ding{51} & - & 0.001 & 0.02 \\
    ~ & Dry Matter & cm & \ding{51} & - & 0.005 & 0.05 \\
    ~ & Carotenoids & car & \ding{55} & 10 & - & - \\
    ~ & Brown Pigments & cbrown & \ding{55} & 0.25 & - & - \\
    ~ & Anthocyanins & anth & \ding{55} & 2 & - & - \\
    ~ & Proteins & cp & \ding{55} & 0.0015 & - & - \\
    ~ & Cabon-based Constituents & cbc & \ding{55} & 0.01 & - & - \\
    \midrule
    \multirow{7}*{Canopy Model} & Leaf Area Index & LAI & \ding{51} & - & 0.01 & 5 \\
    ~ & Leaf Angle Distribution & typeLIDF & \ding{55} & Beta Distribution & - & - \\
    ~ & Hot Spot Size & hspot & \ding{55} & 0.01 & - & - \\
    ~ & Observation Zenith Angle & tto & \ding{55} & 0 & - & - \\
    ~ & Sun Zenith Angle & tts & \ding{55} & 30 & - & - \\
    ~ & Relative Azimuth Angle & psi & \ding{55} & 0 & - & - \\
    \midrule
    \multirow{4}*{Forest Model} & Undergrowth LAI & LAIu & \ding{51} & - & 0.01 & 3 \\
    ~ & Stem Density & sd & \ding{55} & 500 & - & - \\
    ~ & Fractional Coverage & fc & \ding{51} & - & 0.1 & 1 \\
    ~ & Tree Height & h & \ding{51} & * & * & * \\
    ~ & Crown Diameter & cd & \ding{51} & * & * & * \\
    \bottomrule
    \end{tabular}
}
\end{table*}

\subsection{Sentinel-2 observation: spectral bands, temporal, and species information}\label{appx:s2_data}
Detailed information on the spectral bands of $X_{\mathrm{S2}}$ can be viewed in \cref{tab:s2_bands}. Statistics of $X_{\mathrm{S2}}$ can be viewed in \cref{tab:real_data_appx}. These spectra were sampled from individual sites and covered a time sequence of 14 timestamps. 
Note that our current approach to inverting $\mathcal{F}_{\mathrm{RTM}}$ does not integrate temporal information into the learning process, although temporal variations can be evaluated. 
Such temporal information could serve as useful prior knowledge to boost the model's performance, especially to ensure the consistency of temporal variations, which can be considered in future works. 
\begin{table*}[htbp]
\caption{\textbf{Sentinel-2 bands} to use. \textit{VNIR stands for Visible and Near Infrared. SWIR stands for Short Wave Infrared.}}
\label{tab:s2_bands}
\centering
\resizebox{\textwidth}{!}{
    \begin{tabular}{lccccccccccc}
    \toprule
    Band & B2 & B3 & B4 & B5 & B6 & B7 & B8 & B8a & B9 & B11 & B12 \\
    \midrule
    Resolution & 10 m & 10 m & 10 m & 20 m & 20 m & 20 m & 10 m & 20 m & 60 m & 20 m & 20 m \\
    Central Wavelength & 490 nm & 560 nm & 665 nm & 705 nm & 740 nm & 783 nm & 842 nm & 865 nm & 940 nm & 1610 nm & 2190 nm \\
    Description & Blue & Green & Red & VNIR & VNIR & VNIR & VNIR & VNIR & SWIR & SWIR & SWIR \\
    \bottomrule
    \end{tabular}
}
\end{table*}

All samples from $X_{\mathrm{S2}}$ cover both coniferous and broadleaf forests consisting of 5 and 7 species, respectively:
\begin{itemize}
    \item \textbf{Coniferous species}: Pseudotsuga Menziesii (C1), Picea Abies (C2), Pinus Nigra (C3), Larix Decidua (C4), Pinus Sylvestris (C5).
    \item \textbf{Broadleaf species}: Prunus Spp (D1), Fagus Sylvatica (D2), Carpinus Betulus  (D3), Quercus Spp (D4), Acer Pseudoplatanus (D5), Fraxinus Excelsior (D6), Alnus Glutinosa (D7).
\end{itemize}
Larix decidua (C4) is one of the few coniferous species that loses its leaves in autumn, causing its spectral characteristics to resemble those of broadleaf species.

\begin{table*}[htbp]
\caption{\textbf{Information of the Sentinel-2 dataset $X_{\mathrm{S2}}$}}
\label{tab:real_data_appx}
\centering
\resizebox{\textwidth}{!}{
    \begin{tabular}{cccc}
    \toprule
    Total Number of Spectra & Number of Individual Sites & Number of Dates & Number of Species\\
    \midrule
    17962 & 1283 &  14 &  12 \\
    \bottomrule
    \end{tabular}
}
\end{table*}

\subsection{Sensitivity and saturation of leaf area indices from a modelling perspective}\label{appx:rtm_modelling}
We now examine in more detail why both PILA and HVAE have difficulties retrieving $Z_{\text{LAI}}$ and $Z_{\text{LAIu}}$. 
Understanding how $Z_{\text{LAI}}$ and $Z_{\text{LAIu}}$ enter the radiative transfer modelling offer helpful insights for our discussion in \cref{sec:sensitivity}.

\paragraph{Mixing of the spectral reflectance from the canopy and the understory} 
The RTM applied in this study \citep{atzberger2000development} calculates a series of intermediate reflectances using SAIL ($\mathcal{F}_{\text{SAIL}}$) \citep{verhoef1984light}, a canopy reflectance model in which leaf optical properties are described by the PROSPECT model \citep{feret2017prospect}. These intermediate reflectances are then combined according to the forest reflectance model FLIM \citep{rosema1992new}. 

Restricting the set of variables in the FLIM model to $Z_{\text{LAI}}$ and $Z_{\text{LAIu}}$, and treating all other parameters as constant at the forest stand level, the resulting mixed spectral reflectance ($\mathcal{R}$) can be expressed as:
\begin{equation}
    \mathcal{R}(Z_{\text{LAI}}, Z_{\text{LAIu}}) = \lambda_{\text{over}}(Z_{\text{LAI}}) \cdot \mathcal{R}_{\text{over}} \;+\; \lambda_{\text{under}}(Z_{\text{LAI}}) \cdot \mathcal{R}_{\text{under}}(Z_{\text{LAIu}}),
\end{equation}
where
\begin{itemize}
    \item $\mathcal{R}_{\text{over}}$ denotes the reflectance of a fully closed canopy (obtained by running $\mathcal{F}_{\text{SAIL}}$ with $Z_{\text{LAI}}=15$). This reflectance is almost constant and corresponds to the conditions of maximum absorption and scattering by the crown foliage. 
    \item $\mathcal{R}_{\text{under}}(Z_{\text{LAIu}})$ denotes the understory reflectance computed by $\mathcal{F}_{\text{SAIL}}(Z_{\text{LAIu}})$. The quantities $Z_{\text{LAI}}$ and $Z_{\text{LAIu}}$ are both supplied to the same leaf area argument in $\mathcal{F}_{\text{SAIL}}$ via exponential terms. In practice, the dependence follows $\mathcal{R}_{\text{under}} \propto \exp(- Z_{\text{LAIu}})$, approaching saturation once the understory layer becomes sufficiently dense.
    \item $\lambda_{\text{over}}(Z_{\text{LAI}})$ and $\lambda_{\text{under}}(Z_{\text{LAI}})$ are weighting factors determined by forest structural properties, in particular by the fractional cover ($Z_{\mathrm{fc}}$, i.e. the proportion to which the satellite’s optical sensor has a direct view of the ground). Furthermore, the overstory LAI ($Z_{\mathrm{LAI}}$) affects these weights via the canopy transmittances computed with $\mathcal{F}_{\text{SAIL}}(Z_{\text{LAI}})$ as input, which describes how much of the ground the sensor can see through the canopy.
\end{itemize}

\paragraph{Disentanglement}  
Both $Z_{\text{LAI}}$ and $Z_{\text{LAIu}}$ pass through the same $\mathcal{F}_{\text{SAIL}}$ operator and have similar spectral signatures in the Sentinel-2 bands. With single-date, single-view multispectral data, their separate contributions are therefore very hard to disentangle. This explains failure modes where the inversion collapses onto one variable (e.g. $Z_{\text{LAI}}$ remains in a narrow range while $Z_{\text{LAIu}}$ varies), or where both parameters saturate in high-cover, dense forest canopies, as in the Wytham dataset.

\paragraph{Sensitivity and saturation}  
The dependence of $\mathcal{R}$ on $Z_{\text{LAI}}$ and $Z_{\text{LAIu}}$ can be expressed approximately by the following derivatives:
\begin{equation}
    \frac{\partial \mathcal{R}}{\partial Z_{\text{LAI}}} \approx 
    \mathcal{R}_{\text{over}}\cdot\frac{\partial \lambda_{\text{over}}}{\partial Z_{\text{LAI}}}
    + \mathcal{R}_{\text{under}} \cdot \frac{\partial \lambda_{\text{under}}}{\partial Z_{\text{LAI}}},
\end{equation}

\begin{equation}
    \frac{\partial \mathcal{R}}{\partial Z_{\text{LAIu}}} \approx 
    \lambda_{\text{under}} \frac{\partial \mathcal{R}_{\text{under}}}{\partial Z_{\text{LAIu}}}.
\end{equation}

Because both $\lambda_{\text{over}}$ and $\lambda_{\text{under}}$ involve exponentials of $Z_{\text{LAI}}$, their derivatives decay exponentially. Consequently:
\begin{itemize}
    \item For large $Z_{\text{LAI}}$ values, $\exp(-Z_{\text{LAI}})$ becomes negligible. This corresponds to a dense canopy at certain fractional cover. In this regime, forest reflectance is effectively insensitive to additional increases in $Z_{\text{LAI}}$, leading to ``LAI saturation.'' 
    \item $\mathcal{R}_{\text{under}}$ also saturates as $Z_{\text{LAIu}}$ grows. Furthermore, when the fractional cover of the overstory is high and the canopy is dense, $\lambda_{\text{under}}$ becomes very small, strongly reducing the understory signal. Thus, $Z_{\text{LAIu}}$ can only be retrieved reliably in relatively open canopies. A ``LAIu saturation'' occurs when the understory canopy itself becomes dense. 
\end{itemize}

At Wytham Wood, both canopy cover is both high and dense, as verified by ground measurements. This largely accounts for the saturation of both $Z_{\text{LAI}}$ and $Z_{\text{LAIu}}$ when satellite spectra from these locations are used as inputs.

\subsection{Robustness of species separability across multiple runs}\label{appx:robustness_species_separability_JM_distance}
To assess the robustness of the species-separability results on the Austrian dataset (\cref{fig:rtm_inversion_austria_variables_JM_distance} of \cref{sec:results_rtm}), we repeated our experiment using three random seeds (123, 240, and 318) (\cref{tab:species_pairwise_distances}). The original experiment was conducted using random seed 123. The random seed is set before model training and controls the parameter initialisation together with the random shuffling of the training data.

For each run, each tree species is represented by the distribution of its retrieved biophysical variables. We then compute pairwise distances between the 12 species and group the resulting species pairs into three categories: pairs within the five coniferous species, pairs within the seven broadleaf species, and pairs containing one coniferous and one broadleaf species. The reported values are the mean and standard deviation of the average pairwise distance in each category across the three runs.

We report both the Jeffreys-Matusita distance ($d_{\mathrm{JM}}\in[0,2)$) and its underlying Bhattacharyya distance ($d_{\mathrm{B}}\in[0,+\infty)$). These two distances are related by
\begin{equation}
    d_{\mathrm{JM}}
    =
    2\left[1-\exp(-d_{\mathrm{B}})\right].
\end{equation}

This transformation preserves the ordering of all pairwise distances while mapping them to the bounded interval $[0,2)$. However, because the J-M distance saturates as $d_{\mathrm{B}}$ increases, differences between well-separated species pairs, particularly those between coniferous and broadleaf species, become compressed. Consequently, the improvement achieved by PILA appears less pronounced in the J-M distance than in the underlying Bhattacharyya distance. 
For example, for HVAE the average between-group distance increases 20.3\% from 1.581 under the J-M distance to 1.983 under the Bhattacharyya distance, whereas for PILA it increases 44.0\% from 1.719 to 3.070.

Across all three runs, PILA consistently achieves larger within-group and between-group separability than HVAE under both distance measures, supporting the robustness of our findings reported in \cref{fig:rtm_inversion_austria_variables_JM_distance} of \cref{sec:results_rtm}.

\begin{table}[htbp]
\centering
\caption{
\textbf{PILA consistently increases species separation across repeated runs.}
\textit{Pairwise distances between species are computed from their retrieved biophysical variables and averaged over conifer–conifer, broadleaf–broadleaf, and conifer–broadleaf pairs. Results are reported as mean $\pm$ standard deviation across three random seeds (123, 240, 318). Because the Jeffreys–Matusita distance is bounded by 2 and compresses differences between well-separated pairs, the corresponding unsaturated Bhattacharyya distance more clearly shows the increased between-group separation produced by PILA.}
}
\resizebox{0.9\textwidth}{!}{
\begin{tabular}{c l c c c}
\toprule
\textbf{Distance}
& \textbf{Model}
& \textbf{Within conifers}
& \textbf{Within broadleaves}
& \textbf{Between groups} \\
\midrule

\multirow{2}{*}{Jeffreys-Matusita}
& HVAE
& $1.078 \pm 0.020$
& $0.793 \pm 0.014$
& $1.581 \pm 0.036$ \\

& \textbf{PILA}
& $\mathbf{1.225 \pm 0.034}$
& $\mathbf{0.868 \pm 0.030}$
& $\mathbf{1.719 \pm 0.011}$ \\
\midrule

\multirow{2}{*}{Bhattacharyya}
& HVAE
& $0.886 \pm 0.041$
& $0.533 \pm 0.018$
& $1.983 \pm 0.187$ \\

& \textbf{PILA}
& $\mathbf{1.130 \pm 0.043}$
& $\mathbf{0.622 \pm 0.028}$
& $\mathbf{3.070 \pm 0.111}$ \\
\bottomrule
\end{tabular}
}
\label{tab:species_pairwise_distances}
\end{table}

\subsection{Evaluation on the GreenHyperSpectra dataset}\label{appx:rtm_evaluation_greenhyperspectra}

For the RTM inversion, we further evaluate both PILA and the HVAE baseline on the GreenHyperSpectra (GHS) dataset \citep{cherif2026greenhyperspectra}.
The GHS dataset itself is a hyperspectral dataset without ground measurements of vegetation traits. To enable quantitative evaluation, the authors additionally integrate a trait-annotated hyperspectral dataset \citep{cherif2023spectra}.
In their experimental protocol, the trait-annotated dataset is split into training and test sets, with the test set reserved exclusively for evaluation, while the training set is combined with the GHS spectra for model training. 
We follow the same experimental protocol. 
However, unlike their supervised setting, our method performs entirely self-supervised learning, using only the physical model as prior and without leveraging any trait labels from the training set.

The GHS dataset contains 139,295 spectra and is partitioned into 20 stratified subsets, each having a similar data distribution. 
To reduce computational cost, we use only the first two subsets (13,948 spectra), following the observation of \citet{cherif2026greenhyperspectra} that training on a subset achieves comparable performance to using the full dataset due to the similar distributions across subsets. 
Together with the 4,508 spectra from the trait-annotated training set, this results in a total of 18,456 training samples. 
The evaluation is performed on the remaining 4,508 spectra from the held-out test set. 
The spectra contain 1,721 bands spanning the wavelength ranges of 400-1350 nm, 1431-1800 nm, and 2051-2450 nm.

Unlike the Wytham and Austrian datasets \citep{origo2020fiducial,immitzer2019optimal}, which primarily consist of forest vegetation, both the spectra and trait measurements in the GHS benchmark cover a much broader range of vegetation types, including croplands and grasslands in addition to forests. 
We therefore adapt the RTM by removing the forest model and retaining only the leaf model PROSPECT-D and the canopy model SAIL (see \cref{appx:rtm_modelling} for the RTM components), resulting in the widely used PROSAIL model. 
Consequently, forest-specific variables, such as fractional cover and the separation of LAI into canopy and understory components, are no longer modelled, leaving a single LAI variable.

Accordingly, the retrieval variables are reduced to four biophysical variables: chlorophyll content ($Z_{\mathrm{cab}}$), leaf water content ($Z_{\mathrm{cw}}$), leaf dry matter ($Z_{\mathrm{cm}}$), and leaf area index ($Z_{\mathrm{LAI}}$), all of which have corresponding ground-truth measurements in the trait-annotated dataset \citep{cherif2023spectra}. 
The dataset also includes four additional measured traits that are not retrieved due to the ill-posedness of the inverse problem. 
Protein and carbon-based constituents (e.g., starch) are sub-components of leaf dry matter and sum to $Z_{\mathrm{cm}}$, making them difficult to distinguish from dry matter without additional supervision or prior information. 
Accordingly, the PROSPECT-D model also parameterises only $Z_{\mathrm{cm}}$. Similarly, carotenoids and anthocyanins exhibit spectral responses that largely overlap with chlorophyll, but with substantially smaller magnitudes in green vegetation. 
Following the protocol adopted in the main paper, these variables are therefore fixed during training. 
Since the GHS dataset covers a much wider range of vegetation types than the forest datasets, we recalibrate the ranges of the retrieved variables using the training set.

\begin{table*}[htbp] 
\caption{
\textbf{Overview of the biophysical variables of the PROSAIL RTM used to invert the GreenHyperSpectra dataset.} \textit{These variables can be attributed to two hierarchical levels. 4 variables are learnt directly.}
}
\label{tab:para_overview_prosail}
\centering
\resizebox{\textwidth}{!}{
    \begin{tabular}{lcccccc}
    \toprule
    \multirow{2}*{Group} & \multirow{2}*{Variable} & \multirow{2}*{Acronym} & \multirow{2}*{To Learn} & \multirow{2}*{Default Value} & \multicolumn{2}{c}{Sample Range} \\
    \cmidrule(lr){6-7}
    ~ & ~ & ~ & ~ & ~ & Min & Max \\
    \midrule
    Background & Soil brightness factor & psoil & \ding{55} & 0.8 & - & - \\
    \midrule
    \multirow{9}*{Leaf Model} & Structure Parameter & N & \ding{55} & 1.5 & - & - \\
    ~ & Chlorophyll A+B & cab & \ding{51} & - & 0 & 126 \\
    ~ & Water Content & cw & \ding{51} & - & 0.001 & 0.085 \\
    ~ & Dry Matter & cm & \ding{51} & - & 0.001 & 0.070 \\
    ~ & Carotenoids & car & \ding{55} & 8 & - & - \\
    ~ & Brown Pigments & cbrown & \ding{55} & 0.25 & - & - \\
    ~ & Anthocyanins & anth & \ding{55} & 1.25 & - & - \\
    ~ & Proteins & cp & \ding{55} & 0.0015 & - & - \\
    ~ & Cabon-based Constituents & cbc & \ding{55} & 0.01 & - & - \\
    \midrule
    \multirow{7}*{Canopy Model} & Leaf Area Index & LAI & \ding{51} & - & 0.01 & 9.20 \\
    ~ & Leaf Angle Distribution & typeLIDF & \ding{55} & Beta Distribution & - & - \\
    ~ & Hot Spot Size & hspot & \ding{55} & 0.01 & - & - \\
    ~ & Observation Zenith Angle & tto & \ding{55} & 0 & - & - \\
    ~ & Sun Zenith Angle & tts & \ding{55} & 30 & - & - \\
    ~ & Relative Azimuth Angle & psi & \ding{55} & 0 & - & - \\
    \bottomrule
    \end{tabular}
}
\end{table*}

\cref{tab:para_overview_prosail} summarises the RTM components used in PROSAIL together with the corresponding parameter ranges. The same configuration is applied to both PILA and the HVAE baseline.

\begin{figure}[htbp]
    \centering
    \includegraphics[width=\textwidth]{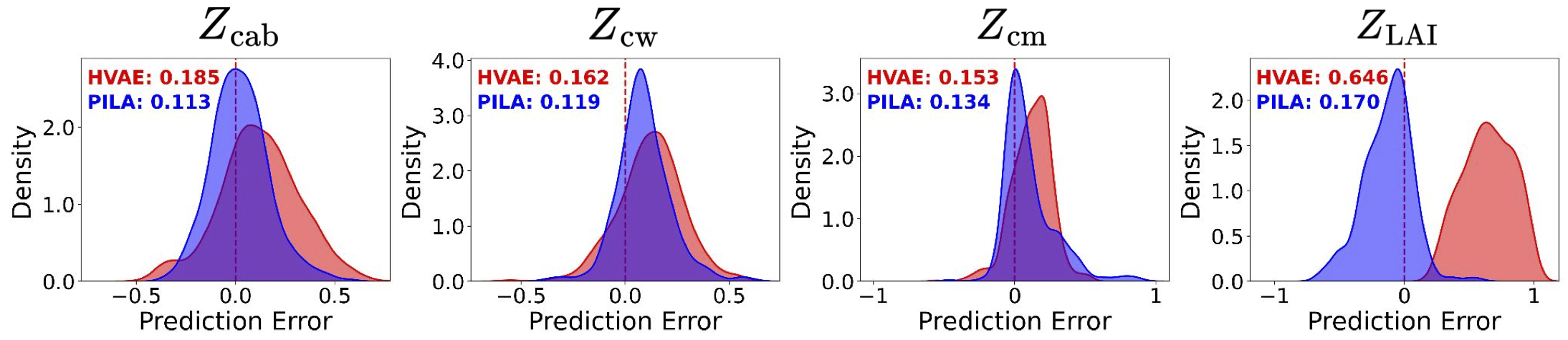}
    \caption{
    \textbf{PILA reduces the prediction error by 12.4--73.7\% compared with the HVAE baseline on the GreenHyperSpectra benchmark.}
    \textit{Results are evaluated on the 4,508 spectra in the held-out test set of the GreenHyperSpectra dataset.}
    }
    \label{fig:rebuttal_greenhyperspectra_inversion_vars}
\end{figure}

\Cref{fig:rebuttal_greenhyperspectra_inversion_vars} visually compares the inversion results of PILA and HVAE by showing the distributions of the prediction errors (MAE). On the GHS dataset, PILA reduces the prediction error by 12.4--73.7\% across the four retrieved variables relative to the HVAE baseline.
\Cref{tab:rtm_inversion_greenhyperspectra} reports the quantitative inversion results for PILA and HVAE. We additionally evaluate a variant of HVAE in which the KL-divergence regularisation is replaced by the end-stop prior used in PILA. 
We follow the same dynamic weighting strategy described in \cref{appx:implementation_detals} when applying the end-stop loss to HVAE, ensuring its meaningful contribution despite the presence of multiple competing loss terms.

\begin{table}[htbp]
\centering
\caption{
    \textbf{PILA achieves the most accurate retrieval of biophysical variables on the GreenHyperSpectra benchmark.}
    \textit{Results are evaluated on the held-out test set of the GreenHyperSpectra dataset. Compared with both the original HVAE and HVAE with the end-stop prior, PILA consistently achieves the lowest standardised mean absolute error (MAE) across all four retrieved biophysical variables.}
}
\resizebox{0.9\textwidth}{!}{
\begin{tabular}{l *{4}{cc}}
\toprule
\multirow{2}{*}{\textbf{Model}}
& \multicolumn{2}{c}{$\mathbf{Z_{\mathrm{cab}}}$}
& \multicolumn{2}{c}{$\mathbf{Z_{\mathrm{cw}}}$}
& \multicolumn{2}{c}{$\mathbf{Z_{\mathrm{cm}}}$}
& \multicolumn{2}{c}{$\mathbf{Z_{\mathrm{LAI}}}$} \\
\cmidrule(lr){2-3}
\cmidrule(lr){4-5}
\cmidrule(lr){6-7}
\cmidrule(lr){8-9}
& \textbf{Mean} & \textbf{MAE}
& \textbf{Mean} & \textbf{MAE}
& \textbf{Mean} & \textbf{MAE}
& \textbf{Mean} & \textbf{MAE} \\
\midrule
HVAE
& 53.054 & 0.185
& 0.0242 & 0.162
& 0.0170 & 0.153
& 9.186 & 0.646 \\

HVAE (end-stop)
& 53.053 & 0.157
& 0.0215 & 0.131
& 0.0182 & 0.158
& 1.833 & 0.185 \\

\textbf{PILA}
& 40.736 & \textbf{0.113}
& 0.0215 & \textbf{0.119}
& 0.0157 & \textbf{0.134}
& 2.075 & \textbf{0.170} \\

Ground Truth
& 37.838 & N/A
& 0.0139 & N/A
& 0.0087 & N/A
& 3.247 & N/A \\
\bottomrule
\end{tabular}
}
\label{tab:rtm_inversion_greenhyperspectra}
\end{table}

PILA achieves the highest retrieval accuracy in both comparisons, highlighting the effectiveness of its simple end-stop prior as well as its low-rank augmentation.
In particular, replacing the KL-divergence regularisation with the end-stop prior reduces the mean absolute error (MAE) compared with the original HVAE. However, its retrieval accuracy remains inferior to that of PILA, suggesting that simply constraining the physical variables at their boundaries is insufficient.

\subsection{Ablation of the physical consistency loss on PILA}\label{appx:syn_loss_on_pila}
We further investigate whether adding the physical consistency loss ($\mathcal{L}_{\mathrm{syn}}$) improves PILA. This follows the same experimental procedure used in HVAE. For each batch of real data (batch size: 256), we randomly sample a batch of synthetic pairs $(Z'_{\mathrm{phy}}, X'_{\mathcal{F}})$ to regularise the physics encoder $\mathcal{E}_{\mathrm{phy}}$ (see \cref{eq:pvae_loss_syn} in \cref{sec:baseline_HVAE} for details). Incorporating this term into the PILA objective (\cref{eq:pila_loss}) gives
\begin{equation}
    \mathcal{L}
    =
    \mathcal{L}_{\mathrm{rec}}
    +
    \beta\mathcal{L}_{\mathrm{prior}}
    +
    \lambda\mathcal{L}_{\mathrm{res}}
    +
    \gamma\mathcal{L}_{\mathrm{syn}}.
\end{equation}

We evaluate three weights for the additional loss, $\gamma\in\{0.1,1,10\}$, while keeping the original RTM inversion settings unchanged ($\beta=1$, $\lambda=0.1$). The experiment is conducted on the GreenHyperSpectra dataset described in \cref{appx:rtm_evaluation_greenhyperspectra}, where four retrieved biophysical variables have corresponding ground-truth trait measurements. We evaluate whether introducing $\mathcal{L}_{\mathrm{syn}}$ reduces the retrieval error measured by the mean absolute error (MAE).

As shown in \cref{tab:pila_syn_loss_ablation}, introducing the physical consistency loss provides no substantial improvement. Across all four biophysical variables, none of the MAEs changes by more than 0.003 after adding $\mathcal{L}_{\mathrm{syn}}$. Furthermore, the average MAE across the four variables exhibits no consistent trend as $\gamma$ increases, and is slightly worse at the largest weight ($\gamma=10$). Overall, these results suggest that incorporating the physical consistency loss does not provide a meaningful benefit for PILA.

\begin{table}[htbp]
\centering
\caption{
\textbf{Introducing the physical consistency loss based on the synthetic data does not help improve PILA's performance.}
\textit{Standardised MAE for PILA trained without physical consistency regularisation introduced by $\mathcal{L}_\mathrm{syn}$ and with different weights $\gamma$. The percentage in the last column denotes the relative change in mean MAE compared with the baseline model without consistency regularisation. Lower values are better.}
}
\resizebox{0.9\textwidth}{!}{
\begin{tabular}{lccccl}
\toprule
\textbf{Model}
& $\mathbf{Z_{\mathrm{cab}}}$ MAE 
& $\mathbf{Z_{\mathrm{cw}}}$ MAE
& $\mathbf{Z_{\mathrm{cm}}}$ MAE
& $\mathbf{Z_{\mathrm{LAI}}}$ MAE
& \textbf{Mean MAE} \\
\midrule
\textbf{PILA}
& 0.113
& 0.119
& 0.134
& 0.170
& 0.134 \\

+ $\mathcal{L}_{\mathrm{syn}}$ ($\gamma=0.1$)
& 0.110
& 0.119
& 0.132
& 0.170
& 0.133 ($-0.7\%$) \\

+ $\mathcal{L}_{\mathrm{syn}}$ ($\gamma=1.0$)
& 0.112
& 0.122
& 0.132
& 0.170
& 0.134 ($-0.0\%$) \\

+ $\mathcal{L}_{\mathrm{syn}}$ ($\gamma=10$)
& 0.113
& 0.121
& 0.134
& 0.173
& 0.135 ($+0.9\%$) \\
\bottomrule
\end{tabular}
}
\label{tab:pila_syn_loss_ablation}
\end{table}

As discussed in \cref{sec:baseline_HVAE}, when the physical model $\mathcal{F}$ is itself incomplete, the synthetic data generated by $\mathcal{F}$ provide only an imperfect supervisory signal. Consequently, the physical consistency loss introduces an additional objective whose reliability depends on the fidelity of the physical model. In contrast, PILA already incorporates the physical prior directly by using $\mathcal{F}$ as the decoder while allowing the low-rank residual to model systematic discrepancies between the physical model and the observations. 
Our experiments suggest that adding $\mathcal{L}_{\mathrm{syn}}$ provides little additional benefit.

\subsection{Formulations of the volcanic deformation model of a Mogi source}\label{appx:mogi_descriptions}
$\mathcal{F}_{\mathrm{Mogi}}$ describes the displacement field on the surface that results from a spherical pressure source, typically a magma chamber, at depth. The inverse problem of $\mathcal{F}_{\mathrm{Mogi}}$ deals with estimating the location and volume change of magma chamber activities given the observed displacements at the ground stations. In this study, $\mathcal{F}_{\mathrm{Mogi}}$ to use is further simplified by assuming only a point pressure source. Specifically, for $N$ ground stations located at $\{x_i, y_i, z_i, i=1,\cdots,N\}$, the displacement field $X_{\mathrm{GNSS}}=\{x_{e,i}, x_{n,i}, x_{v,i}, i=1,\cdots,N\}$ on the surface along the east, north and vertical directions due to the subsurface volumetric change of a Mogi source is estimated by:
\begin{align}
    \hat{x}_{e,i} &= \frac{\alpha (x_i-Z_{\mathrm{x_m}})}{R_i^3}, \\
    \hat{x}_{n,i} &= \frac{\alpha (y_i-Z_{\mathrm{y_m}})}{R_i^3}, \\
    \hat{x}_{v,i} &= \frac{\alpha (z_i-Z_{\mathrm{d}})}{R_i^3} 
\end{align}
where $Z_{\mathrm{x_m}}$ and $Z_{\mathrm{y_m}}$ are the horizontal coordinates of the Mogi source, and $Z_{\mathrm{d}}$ is its depth. $\alpha$ is the scaling factor that includes the volume change $Z_{\mathrm{\Delta V}}$ and the material properties (Poisson's ratio of the medium), given by:
\begin{equation}
     \alpha = \frac{(1 - \nu) Z_{\mathrm{\Delta V}}}{\pi}
\end{equation}
and $R_i$ is the radial distance from the Mogi source point to the station at ($x_i$, $y_i$, $0$) (the equations are valid for a half-space, so elevation is neglected):
\begin{equation}
    R_i = \sqrt{(x_i - x_m)^2 + (y_i - y_m)^2 + Z_{\mathrm{d}}^2}
\end{equation}
For the inversion problem, we assume that the following parameters are known:
\begin{itemize}
    \item $\nu$, Poisson's ratio, usually set at 0.25, a typical value for the Earth's crust.
    \item $\{x_i, y_i\, i=1,\cdots, N\}$, location of the GNSS stations.
\end{itemize}
Thus, our goal is to estimate the location $(Z_{\mathrm{x_m}}, Z_{\mathrm{y_m}}, Z_{\mathrm{d}})$ and volume change $Z_{\mathrm{\Delta V}}$ of the Mogi source, given the measured displacements recorded by the GNSS stations $X_{\mathrm{GNSS}}=\{u_{e,i}, u_{n,i}, u_{v,i}, i=1,\cdots,N\}$.

\begin{figure}[htbp]
    \centering
    \includegraphics[width=0.7\textwidth]{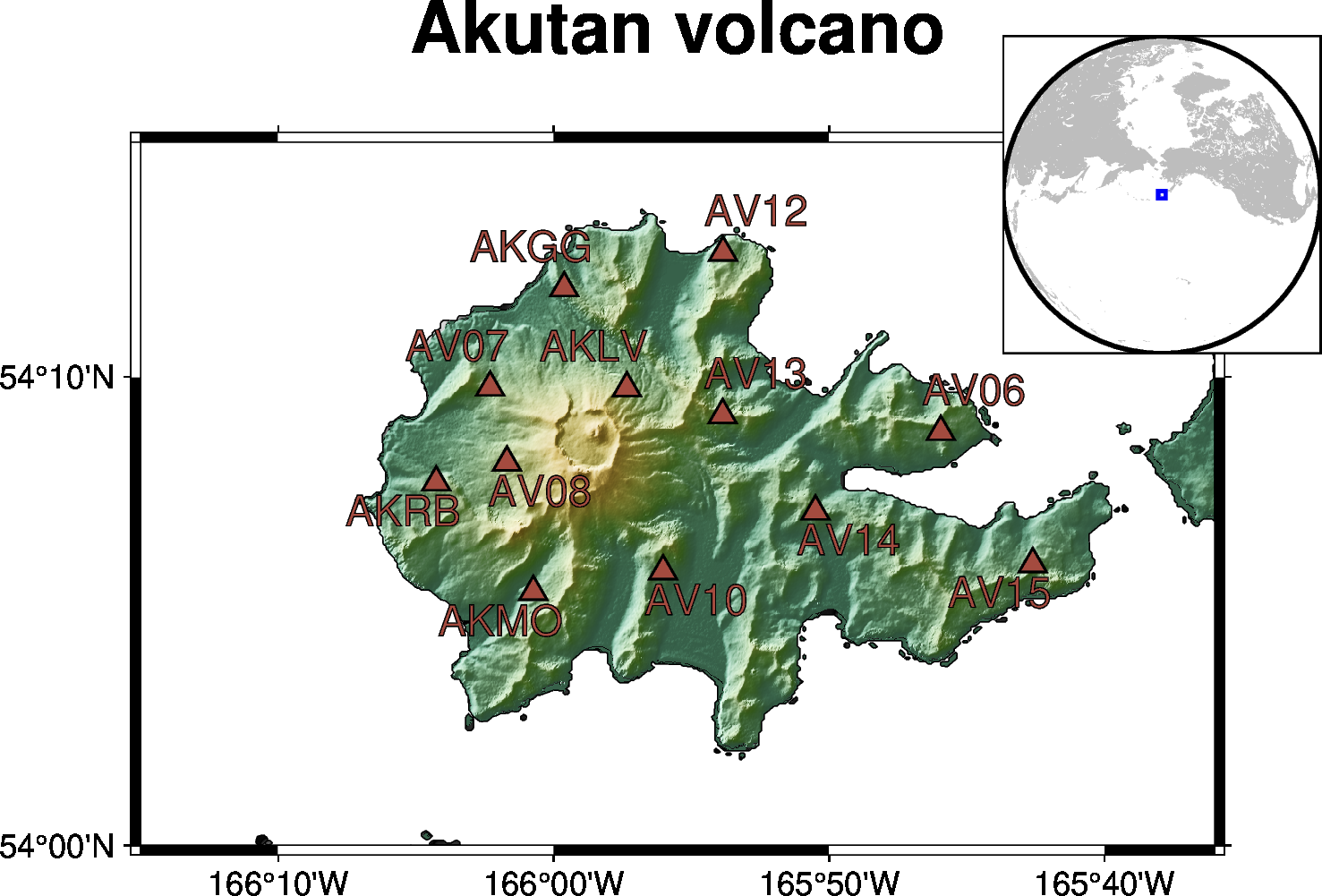}
    \caption{\textbf{Locations of the 12 GNSS stations around the Akutan Volcano.}}
    \label{fig:map_akutan_volcano_appx}
\end{figure}

\subsection{GNSS observation}\label{appx:gnss_data}
Surface displacement is monitored through continuous GNSS (Global Navigation Satellite System) stations that produce daily position time series (\url{http://geodesy.unr.edu}), \cite{blewitt2018harnessing}. 
For the Akutan volcano, we have 12 GNSS stations within the region (\cref{fig:map_akutan_volcano_appx}). From these stations, we have compiled daily time-series GNSS displacement data spanning from 2006 to 2024, which we will use to test our inversion approach. GNSS stations are often missing data due to various issues such as power outages or instrument replacements. Approximately 20\% of our raw time series data is missing, which, although significant, is still considered good in terms of GNSS observations. We have used a variational Bayesian Independent Component Analysis \citep{gualandi2016blind} to fill these data gaps using existing observations to ensure that we have a complete daily time series.
Our model takes the displacement field observed on each day as a training sample. Similarly to the RTM case, future work could consider integrating temporal priors. 

\subsection{Assumed sources of the GNSS-observed displacement fields}\label{appx:gnss_data_components}
In addition to deformations due to a Mogi source, the surface displacement observed by a GNSS station can be influenced by other physical processes and noise sources. In simplified form, a displacement $u(t)$ of a station along a certain direction at time $t$ can be modelled as follows \citep{bevis2014trajectory}:
\begin{equation}
    u(t) = S_{\mathrm{linear}}(t) + S_{\mathrm{annual}}(t) + S_{\mathrm{semiannual}}(t) + S_{\mathrm{volcanic}}(t) + \epsilon_{\mathrm{pink}} + \epsilon_{\mathrm{white}} \label{eq:gps_total}
\end{equation}
Where:
\begin{align}
    & S_{\mathrm{linear}}(t) = q + m \cdot t \label{eq:gps_linear}\\
    & S_{\mathrm{annual}}(t) = A_1 \cdot \sin\left(\frac{2\pi t}{T_1}\right) + B_1 \cdot \cos\left(\frac{2\pi t}{T_1}\right) \label{eq:gps_annual}\\
    & S_{\mathrm{semiannual}}(t) = A_2 \cdot \sin\left(\frac{2\pi t}{T_2}\right) + B_2 \cdot \cos\left(\frac{2\pi t}{T_2}\right) \label{eq:gps_semi}\\
    & S_{\mathrm{volcanic}}(t) = C \cdot Z_{\mathrm{\Delta V, t}}\label{eq:gps_eruption}
\end{align}
$S_{\mathrm{linear}}$ represents the linear trend, often due to tectonic plate movement; $S_{\mathrm{annual}}$ and $S_{\mathrm{semiannual}}$ represent the annual and semiannual seasonality of the signal, potentially due to hydrological processes; and $S_{\mathrm{volcanic}}$ accounts for the deformation due to the Mogi source since the inflation event happened at $t_s$ and depends on the volume change $Z_{\mathrm{\Delta V, t}}$ at time step $t$. $\epsilon_{\mathrm{pink}}$ and $\epsilon_{\mathrm{white}}$ are pink and white noises, respectively, commonly observed in natural phenomena.

Volcanic deformation $S_{\mathrm{volcanic}}(t)$ can be small (usually at the $\mathrm{mm}$ level) compared to other sources of deformation in the \cref{eq:gps_total} level ($\mathrm{mm}$ to $\mathrm{cm}$), making the data-driven compensation both an essential ingredient for accurate results and a challenging task due to incompleteness of $\mathcal{F}_{\mathrm{Mogi}}$. 

\subsection{Synthetic GNSS observations with controlled residual rank}\label{appx:gnss_synthetic_data}
\paragraph{Data generation}
To isolate the effect of PILA's low-rank residual from the case-specific structure of the real GNSS data, we construct synthetic observations in which the unmodelled component has a precisely controlled intrinsic rank $r'$. We then apply PILA with different residual ranks $r$ to examine their effects on reconstruction and physical-variable retrieval.

As introduced in \cref{eq:forward_observation}, for a real observation $X$, $X_{\mathcal F}$ is the component described by the physical forward model, $\Delta$ is the systematic component not represented by that model, and $\epsilon$ is observational noise.

For GNSS observations around a volcano, $X_{\mathcal F}$ can be modelled using a Mogi source driven by a volume change $\Delta V$. The unmodelled component $\Delta$ may contain long-term and seasonal signals, while $\epsilon$ may contain white and pink noise. Further details of the Mogi model and the assumed composition of the GNSS signals are provided in \cref{appx:mogi_descriptions} and \cref{appx:gnss_data_components}.

In this controlled experiment, we consider only the physically modelled component and the systematic residual, while ignoring the noise term:
\begin{equation}\label{eq:mogi_synthetic_observation}
X' = X_{\mathcal F} + \Delta'.
\end{equation}

We use the geographic locations of the 12 GNSS stations around Akutan Volcano. Each observation contains three displacement components at each station, giving $d=36$ dimensions. We generate $n=2922$ daily observations, corresponding to approximately eight years beginning on 1 January 2006. Therefore, $X'$, $X_{\mathcal F}$, and $\Delta'$ are all in $\mathbb{R}^{n\times d}$.

\paragraph{Mogi source with transient inflation}

We simplify the Mogi-source model by fixing the horizontal source location to $(x_m,y_m)=(2,0),\mathrm{km}$, approximately at the centre of the GNSS network, and its depth to $d_m=5,\mathrm{km}$. We vary only the source volume change:
\begin{equation}
\Delta V(t)
=
\Delta V_{\mathrm{base}}
+
s_{\Delta V}\rho(t),
\end{equation}
where $\Delta V_{\mathrm{base}}$ is a constant offset and $s_{\Delta V}$ controls the magnitude of the transient inflation. These values are selected so that $\Delta V(t)$ lies within a realistic range.

The transient inflation is represented using an arctangent function:
\begin{equation}
\rho(t)
=
\arctan\left(\frac{t-t_0}{w}\right),
\end{equation}
where $t_0$ determines the timing of the inflation event and $w$ controls how rapidly it develops. We normalise $\rho(t)$ using its minimum and maximum values over the complete time span so that $\rho(t)\in[0,1]$. We set $t_0$ to day 1168 and $w$ to 30 days, such that around $90\%$ of the total volume change occurs over approximately nine months.

Using a constant unit volume change, $\Delta V=1$, we obtain the unit spatial direction of the Mogi-induced deformation:
\begin{equation}\label{eq:mogi_unit_direction}
\mathbf{m}
=
\frac{
\mathcal{F}(\Delta V=1)
}{
\Vert
\mathcal{F}(\Delta V=1)
\Vert
},
\end{equation}
where $\mathbf{m}\in\mathbb{R}^{d}$ represents the three displacement components across the 12 stations.

Because $(x_m,y_m,d_m)$ are fixed, the Mogi displacement is linear in $\Delta V(t)$. The physical component at time $t$ can therefore be written as
\begin{equation}
X_{\mathcal F}(t)
=
\Delta V(t)\mathbf{m}.
\end{equation}

Stacking the displacement fields over all $n$ time steps gives $X_{\mathcal F}\in\mathbb{R}^{n\times d}$.

\paragraph{Modelling the unmodelled component with controlled rank}

The signals not represented by the Mogi source may arise from processes such as residual tectonic motion and hydrological loading. These signals commonly contain long-term trends and seasonal variations, which can be approximated using linear and sinusoidal functions.

Our objective is to construct a synthetic residual $\Delta'\in\mathbb{R}^{n\times d}$ with an assigned rank $r'$:
\begin{equation}\label{eq:mogi_synthetic_delta}
\Delta'
=
s_{\Delta'}\mathbf{S}\mathbf{V}^{\top},
\end{equation}
where $\mathbf{S}\in\mathbb{R}^{n\times r'}$ represents the temporal signals, $\mathbf{V}\in\mathbb{R}^{d\times r'}$ represents their spatial distributions across the 12 stations, and $s_{\Delta'}$ controls the magnitude of the residual. We next describe how $\mathbf{S}$ and $\mathbf{V}$ are constructed.

\paragraph{Sampling temporal trajectories}

We first generate $r'$ temporal trajectories $\mathbf{s}_i\in\mathbb{R}^{n}$, for $i\in\{1,\ldots,r'\}$, using components commonly employed to describe GNSS station motion \citep{bevis2014trajectory}:
\begin{equation}
\mathbf{s}_i(t)
=
a_{\mathrm{linear}}\phi_{i,1}\frac{t}{T}
+
a_{\mathrm{annual}}
\cos\left(
\frac{2\pi t}{T}+\phi_{i,2}
\right)
+
a_{\mathrm{semi}}
\cos\left(
\frac{2\pi t}{T/2}+\phi_{i,3}
\right).
\end{equation}

Here, $T=365.25$ days represents one annual cycle. The sign of the linear trend is randomly sampled as $\phi_{i,1}\in\{-1,1\}$, while the annual and semi-annual phases are randomly sampled as $\phi_{i,2},\phi_{i,3}\sim \mathcal{U}(0,2\pi)$. The relative amplitudes are set to
$
(a_{\mathrm{linear}}
,
a_{\mathrm{annual}}
,
a_{\mathrm{semi}})
=
(1,2,1).
$

All three components are included in each trajectory. Randomly sampling $\phi_{i,1}$, $\phi_{i,2}$, and $\phi_{i,3}$ produces different temporal trajectories with the same general GNSS-inspired composition. Stacking them gives
\begin{equation}
[\mathbf{s}_1,\ldots,\mathbf{s}_{r'}]
\in
\mathbb{R}^{n\times r'}.
\end{equation}

\paragraph{Sampling spatial distributions}

The temporal trajectories are distributed across the $d=36$ displacement dimensions. We first sample $r'$ spatial vectors independently from a standard normal distribution:
\begin{equation}
\mathbf{G}
=
[\mathbf{g}_1,\ldots,\mathbf{g}_{r'}]
\sim
\mathcal{N}(0,1)^{d\times r'}.
\end{equation}

The vector $\mathbf{m}$ in \cref{eq:mogi_unit_direction} describes the spatial direction of the deformation induced by the Mogi source. Therefore, $\mathbf{m}^{\top}\mathbf{g}_i$ measures the projection of each sampled spatial vector $\mathbf{g}_i$ onto the Mogi displacement direction. We remove this projected component by defining
\begin{equation}
\mathbf{v}_i
=
\mathbf{g}_i
-
(\mathbf{m}^{\top}\mathbf{g}_i)\mathbf{m}.
\end{equation}

This gives $\mathbf{v}_i\perp\mathbf{m}$, meaning that the sampled spatial distribution does not overlap with the Mogi-induced displacement direction. This helps separate the physical component $X_{\mathcal F}$ from the residual component $\Delta'$.

Stacking the spatial vectors gives
\begin{equation}
[\mathbf{v}_1,\ldots,\mathbf{v}_{r'}]
\in
\mathbb{R}^{d\times r'}.
\end{equation}

\paragraph{Orthonormalisation}

We orthonormalise the sampled temporal trajectories and spatial distributions by applying QR decomposition separately:
\begin{equation}
\mathbf{S}
=
\mathcal{F}_{\mathrm{QR}}
\left(
[\mathbf{s}_1,\ldots,\mathbf{s}_{r'}]
\right),
\qquad
\mathbf{V}
=
\mathcal{F}_{\mathrm{QR}}
\left(
[\mathbf{v}_1,\ldots,\mathbf{v}_{r'}]
\right).
\end{equation}

The resulting matrices satisfy
\begin{equation}
\mathbf{S}^{\top}\mathbf{S}
=
\mathbf{I},
\qquad
\mathbf{V}^{\top}\mathbf{V}
=
\mathbf{I}.
\end{equation}

Thus, the columns of $\mathbf{S}$ represent independent temporal directions, while the columns of $\mathbf{V}$ represent independent spatial directions. Substituting these matrices into the residual construction in \cref{eq:mogi_synthetic_delta} gives
\begin{equation}
\operatorname{rank}(\Delta')=r'.
\end{equation}

\vspace{-5mm}
\subsection{Formulation of the latent space of physical variables}\label{appx:HVAE_formulations}
\paragraph{Issues in the HVAE configuration}
In a standard VAE, a standard normal prior $\mathcal{N}(0,I)$ regularizes the latent space of the posterior $q(U \mid X)$, and the KL term is computed as:
\begin{equation}
    \mathcal{L}_{\text{KL}, U} = \mathcal{L}_\mathrm{KL}\!\left(q(U\mid X)\,\|\,\mathcal{N}(0,I)\right).
\end{equation}

When the decoder is replaced by a forward physical model $\mathcal{F}$, it is common to learn $Z_{\text{phy}}$ in bounded ranges, e.g. $Z_{\text{phy}} \in [0, 1]$, then rescale them to their original ranges before using them as input to $\mathcal{F}$. This contrasts with the unbounded variables $U$ in a standard VAE.

To learn $Z_{\text{phy}} \sim \mathcal{N}\!\big(\mu_{\text{phy}},\,\sigma^2_{\text{phy}}\big)$, HVAE uses the following formulation in the encoder $\mathcal{E}_{\text{phy}}$:
\begin{itemize}
    \item The output of the last linear layer of $\mathcal{E}_{\mathrm{phy}}$ (denoted as $\mu$), used to predict $\mu_{\text{phy}}$, is passed through a SoftPlus activation so that $\mu_{\text{phy}} > 0$:
    \begin{equation}
        \mu_{\text{phy}} = f_\mathrm{softplus}(\mu) = \log(1 + e^\mu).
    \end{equation}
    The prediction for $\sigma_{\text{phy}}$ remains unbounded. 
    \item The sampled variable $Z_{\text{phy}} \sim \mathcal{N}\!\big(\mu_{\text{phy}},\,\sigma^2_{\text{phy}}\big)$ is clamped to the bounded range $[0,1]$ before being passed to the physical model $\mathcal{F}$. 
    \item The prior distribution is no longer standard normal. Instead, to match the bounded range $[0,1]$, they essentially used the following prior for all their study cases after scaling the physical variables to the common range: 
    \[
        p(Z_{\text{phy}}) \sim \mathcal{N}\!\big(0.5,\,(0.866)^2\mathbf{I}),
    \]
\end{itemize}

In practice, this setup could be problematic for several reasons:
\begin{itemize}
    \item The SoftPlus activation makes the gradient unstable, often resulting in extremely large KL terms. 
    \item Clamping uses only the part of $q(Z_{\text{phy}}\mid X)$ within $[0,1]$, but the full distribution is compared to $\mathcal{N}\!\big(0.5,\,0.866^2\big)$ when computing the losses, creating a problematic mismatch.
    \item The prior distribution deviates from a standard normal, and the prior $p(Z_{\mathrm{phy}})$ is set somewhat arbitrarily (e.g., variance as $0.866^2$). 
\end{itemize}

In HVAE implementations, the KL term weight is typically set very small (e.g., $\beta = e^{-11}$ in the galaxy experiment and $\beta = e^{-9}$ in the pendulum experiment, likely to reduce the impact of the prior. We adopt a similar weight ($\beta = e^{-9}$) when reproducing this baseline.

\paragraph{Prior integration in \cref{sec:prior}}
We discuss the use of different physical priors in \cref{sec:prior}, where we place a standard normal prior on the unbounded latent space $U_{\text{phy}}$ to compute the KL term:
\begin{equation}
    \mathcal{L}_{\mathrm{KL}} = \mathrm{KL}\!\left(q(U_{\mathrm{phy}}\mid X)\,\|\,\mathcal{N}(0,I)\right).
\end{equation}

The sampled variable $U_{\text{phy}} \sim \mathcal{N}(\mu,\sigma^2)$ is then passed through a sigmoid function to obtain $Z_{\text{phy}} \in (0,1)$:
\begin{equation}
    Z_{\mathrm{phy}} = f_{\text{sigmoid}}(U_{\mathrm{phy}}) = \frac{1}{1 + e^{-U_{\mathrm{phy}}}}.
\end{equation}

This approach offers the following benefits when applying a KL term to integrate a physical prior:
\begin{itemize}
    \item The latent distribution $U_{\text{phy}} \sim \mathcal{N}\!\big(\mu,\,\sigma^2\big)$ follows standard VAE practice with unbounded $\mu$ and $\sigma$, which is well-behaved and easier to control. 
    \item The Sigmoid maps any unbounded $U_{\text{phy}}$ to $Z_{\text{phy}} \in (0,1)$, avoiding clamping. 
    \item The same design applies when a more informative physical prior as formulated in \cref{eq:informative_prior} is available and a KL term regularizes the latent space before mapping to bounded physical variables. 
\end{itemize}

\subsection{Implementation details}
\label{appx:implementation_detals}
For a fair comparison, we use the same depth and width for the MLP layers in all modules where the corresponding components in HVAE and PILA share the same notation (\cref{fig:methods_phylora} and \cref{fig:methods_physvae}). The feature extractor $\mathcal{E}_{\mathrm{R}}$ is implemented here as a shallow MLP that maps $X$ to a 128-dimensional feature representation $R$ for encoding.

All datasets are split into training, validation, and test subsets, and all reported results use final evaluations on the test sets. Ground measurements in Wytham Wood were collected at 42 sampling sites between 3 and 5 July 2018. Each site is a $20~m\times20~m$ plot where multiple samples were averaged to obtain the ground measurements. Further details on these measurements and the sampling protocol are provided in \citep{origo2020fiducial}. To validate against the ground measurements, we extract the forest spectra at the 42 site locations from Sentinel-2 observations on the four closest evaluation dates. 
For the GNSS data, the 2006–2009 time series are held out as a test set to ensure coverage of the 2008 inflation event. All models are trained with the Adam optimizer, an initial learning rate of $0.0003$, weight decay of $0.0001$, and 150 training epochs. 

\begin{figure}[htbp]
    \centering
    \includegraphics[width=0.9\textwidth]{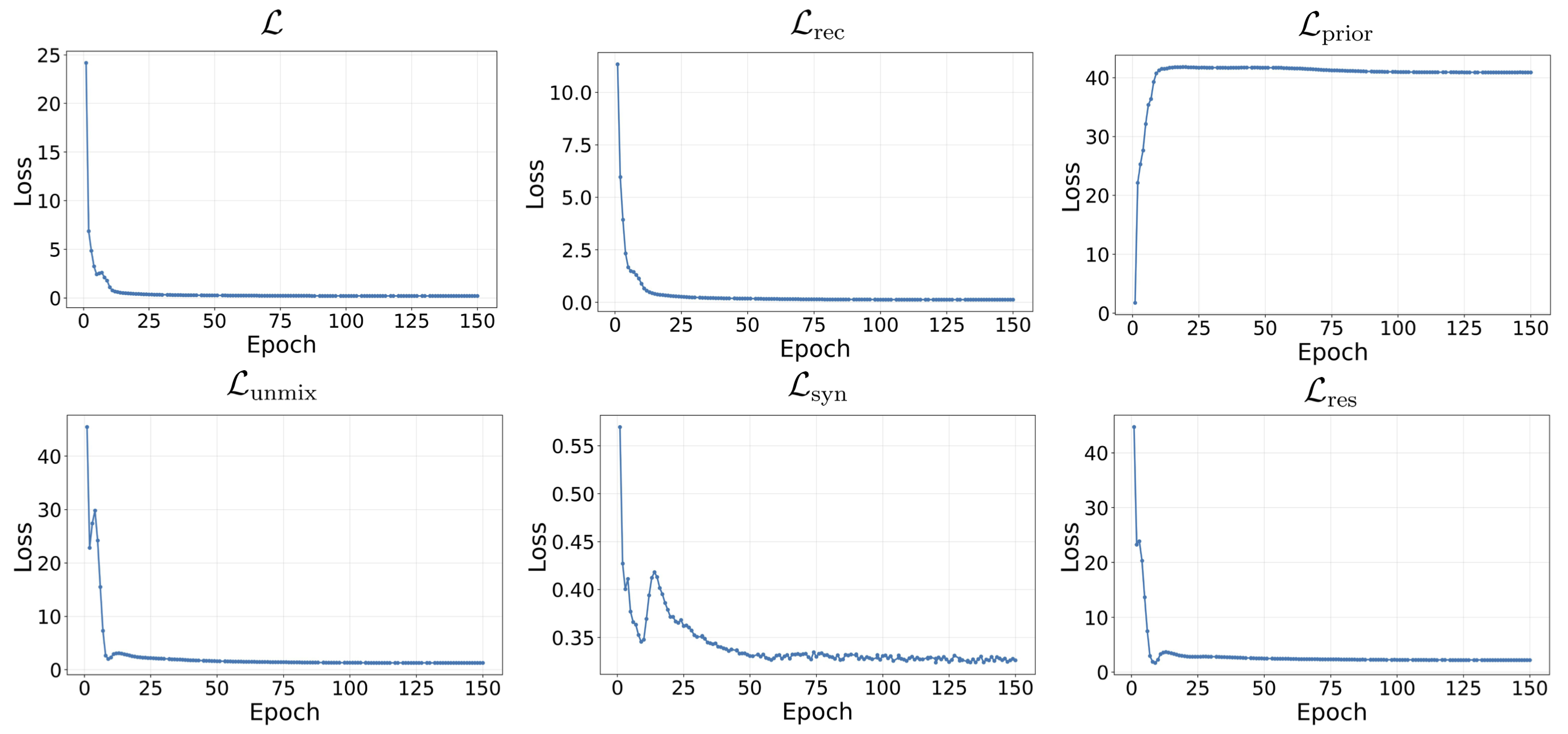}
    \caption{\textbf{The loss terms in HVAE converge and achieve a balance during training.}}
    \label{fig:appx_convergence_plot_HVAE}
\end{figure}

For HVAE, following \citet{takeishi2021physics}, we set the KL coefficient in the ELBO to a small value ($e^{-9}$), matching their implementation. For $\lambda_{1}$, $\lambda_{2}$, and $\lambda_{3}$, we similarly tune them relative to the ELBO (primarily the reconstruction loss): we adjust $\lambda_{1-3}$ based on the observed loss magnitudes so that, after scaling, each regularisation term is roughly one order of magnitude smaller than the reconstruction loss. 
Empirically, these weights yield a stable balance among loss terms during training (\cref{fig:appx_convergence_plot_HVAE}), ensuring that each HVAE regularisation term contributes to the objective without any single term dominating.

\begin{figure}[htbp]
    \centering
    \includegraphics[width=0.6\textwidth]{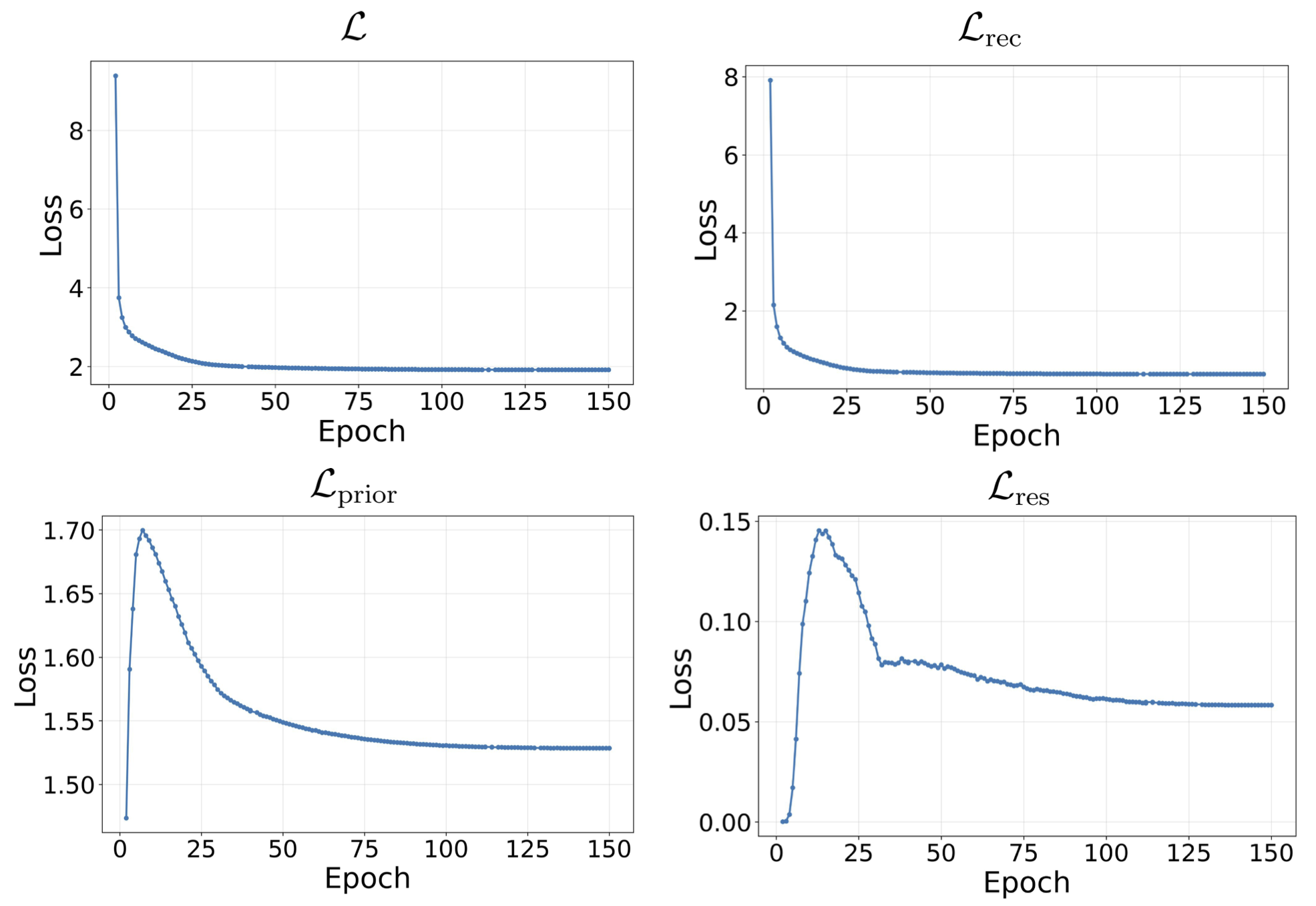}
    \caption{\textbf{The loss terms in PILA also converge during training.}}
    \label{fig:appx_convergence_plot_PILA}
\end{figure}

The hyperparameters for PILA are likewise chosen so that their contributions match the reconstruction loss. For RTM inversion, we set $\beta=1$ for $\mathcal{L}_{\mathrm{prior}}$ and $\lambda=0.1$ for $\mathcal{L}_{\mathrm{res}}$; for Mogi inversion, we use $\beta=10$ and $\lambda=0.1$. For PILA inversion, we apply 30 epochs of annealing on the residual $\Delta$, gradually increasing its weight from 0 to 1 to progressively introduce the data-driven augmentation. All PILA loss terms converge by the end of training (\cref{fig:appx_convergence_plot_PILA}).
All experiments are run on a system equipped with an AMD EPYC 7702 64-Core Processor and 1 TB of RAM. 

\vspace{-5pt}
\FloatBarrier
\clearpage
\subsection{Extended RTM inversion results for Austrian data in \cref{sec:results_rtm}}\label{appx:results_rtm_austria}
\vspace{-5pt}
\begin{figure}[htbp]
    \centering
    \includegraphics[width=0.9\textwidth]{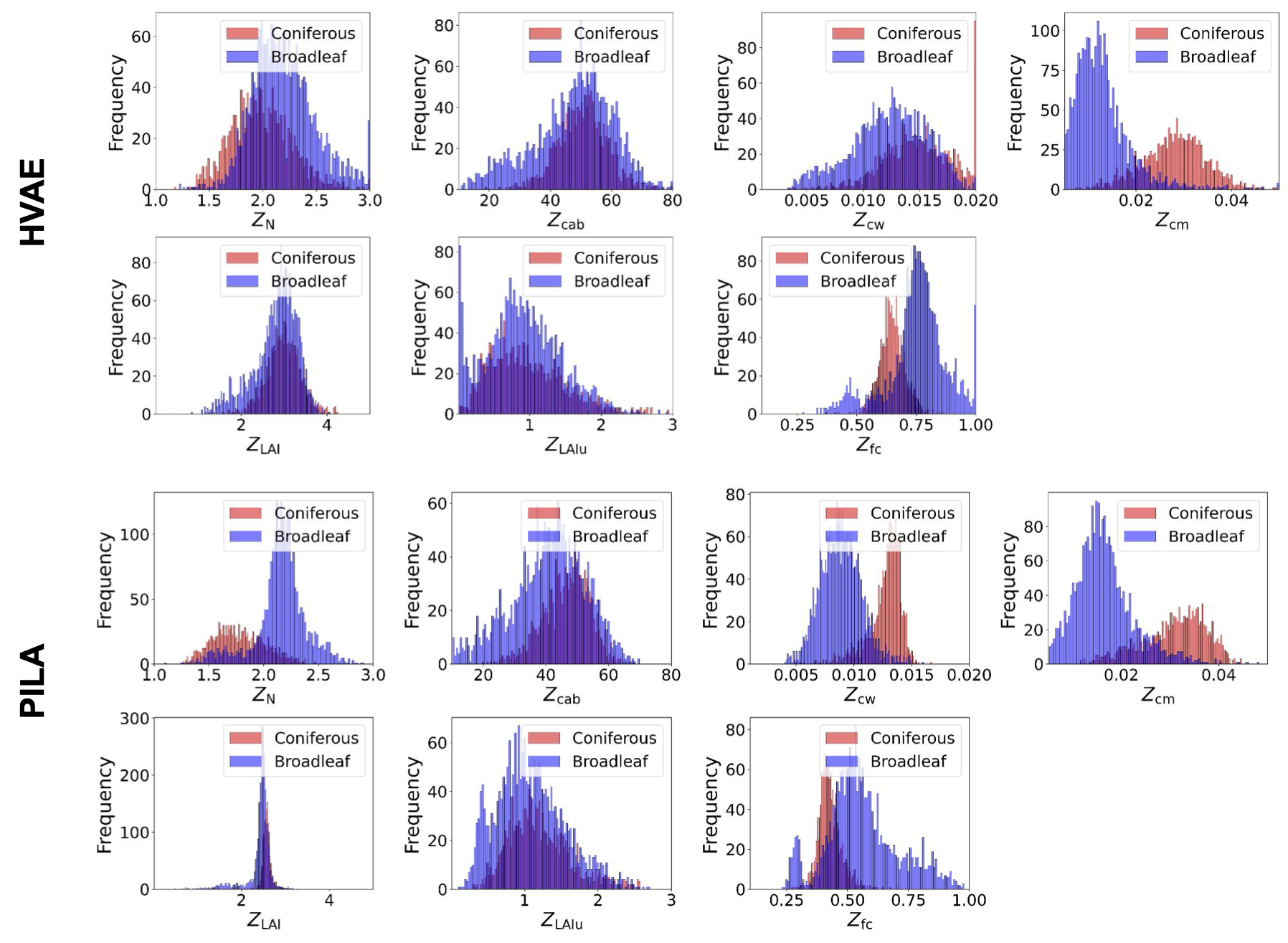}
    \caption{\textbf{Extended results complementing \cref{fig:rtm_inversion_austria_variables} in the main text}, including the distributions of all retrieved variables.}
    \label{fig:appx_rtm_inversion_austria_variables}
\end{figure}

\vspace{-5pt}

\begin{figure}[htbp]
    \centering
    \includegraphics[width=0.9\textwidth]{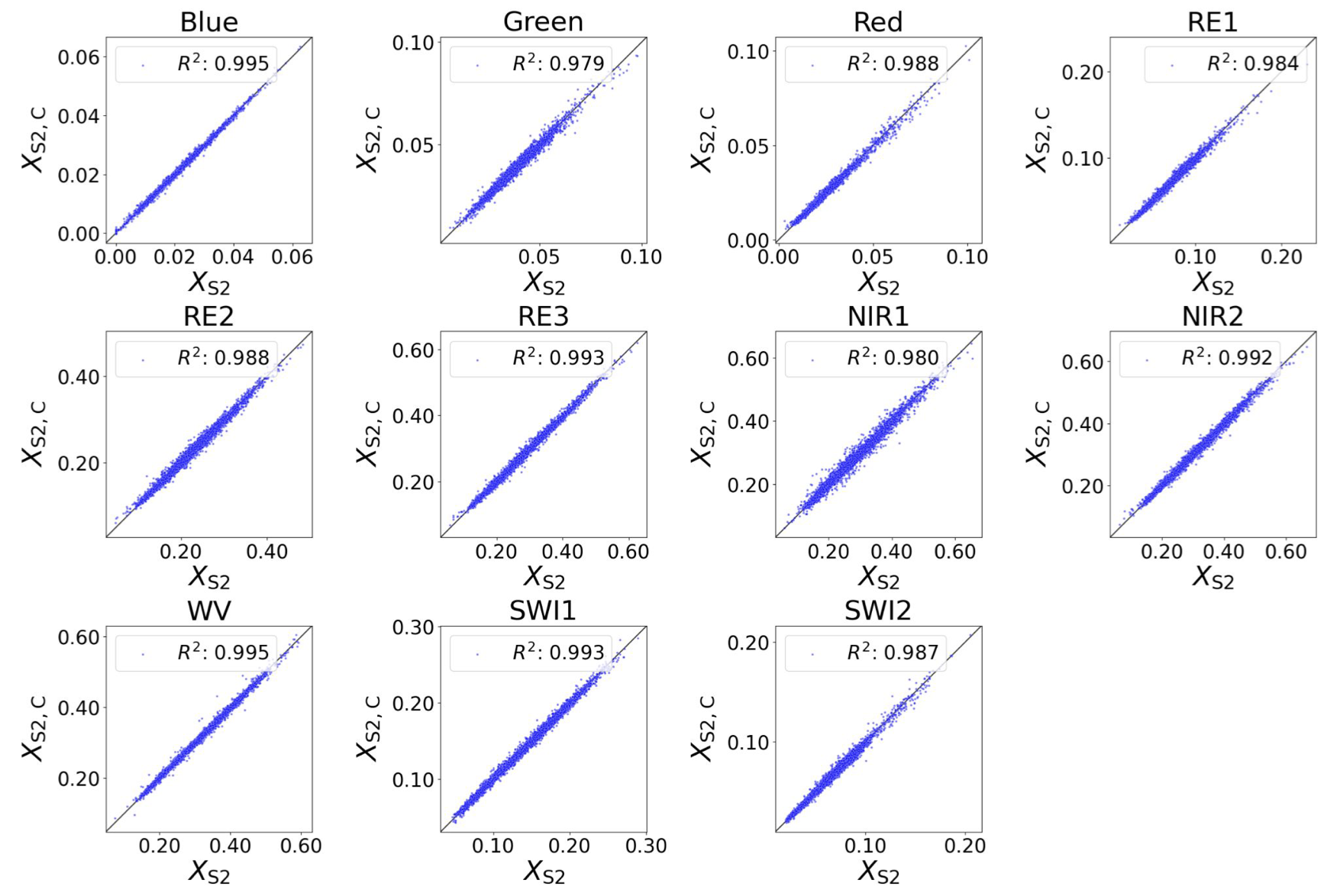}
    \caption{\textbf{Extended results complementing \cref{fig:rtm_inversion_austria_spectra} in the main text}, including HVAE's spectral reconstructions for all Sentinel-2 bands.}
    \label{fig:appx_rtm_inversion_austria_spectra_HVAE}
\end{figure}

\begin{figure}[htbp]
    \centering
    \includegraphics[width=0.9\textwidth]{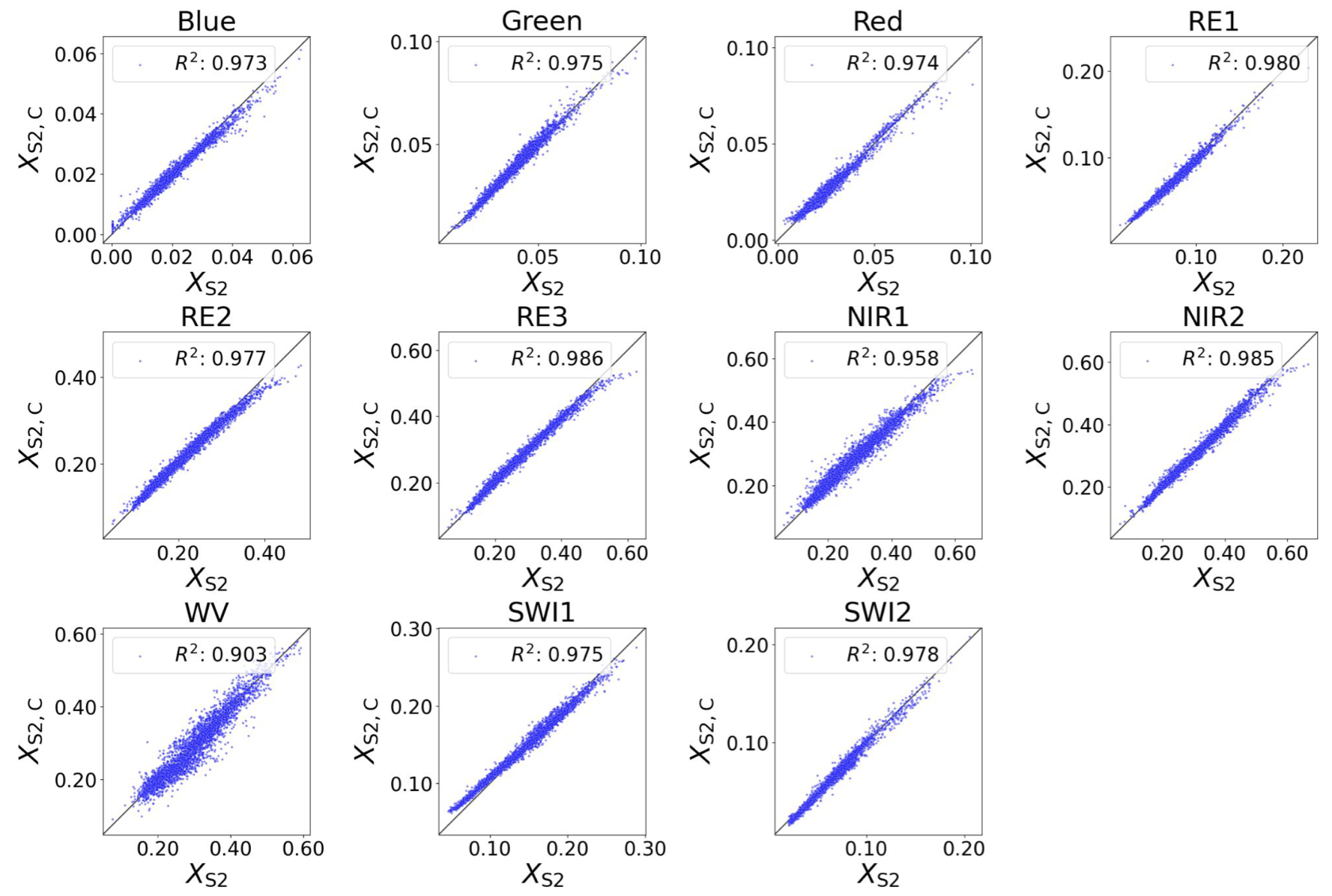}
    \caption{\textbf{Extended results complementing \cref{fig:rtm_inversion_austria_spectra} in the main text}, including PILA's spectral reconstructions for all Sentinel-2 bands.}
    \label{fig:appx_rtm_inversion_austria_spectra_PILA}
\end{figure}

\FloatBarrier
\subsection{Extended RTM inversion results for Wytham data in \cref{sec:results_rtm}}\label{appx:results_rtm_wytham}

\vspace{-5pt}
\begin{table}[htbp]
\centering
\caption{\textbf{Extended results complementing \cref{tab:rtm_inversion_wytham_variables_insitu_july6} in the main text}, including metrics for all dates. \textit{PILA substantially reduces prediction errors for all dates with available Sentinel-2 observations, especially July 6, which is closest to the ground measurement period.}}
\resizebox{0.9\textwidth}{!}{
\begin{tabular}{l l *{4}{cc}}
\toprule
\multirow{2}{*}{\textbf{Date}} & \multirow{2}{*}{\textbf{Model}} & \multicolumn{2}{c}{$\mathbf{Z_{\mathrm{cab}}}$} & \multicolumn{2}{c}{$\mathbf{Z_{\mathrm{fc}}}$} & \multicolumn{2}{c}{$\mathbf{Z_{\mathrm{LAI}}}$} & \multicolumn{2}{c}{$\mathbf{Z_{\mathrm{LAIu}}}$} \\
\cmidrule(lr){3-4} \cmidrule(lr){5-6} \cmidrule(lr){7-8} \cmidrule(lr){9-10}
& & \textbf{Mean} & \textbf{MAE} & \textbf{Mean} & \textbf{MAE} & \textbf{Mean} & \textbf{MAE} & \textbf{Mean} & \textbf{MAE} \\
\midrule
\multirow{2}{*}{26 June} & HVAE & 35.888 & 0.159 & 0.988 & 0.199 & 2.511 & 0.189 & 2.063 & 0.339 \\
& \textbf{PILA} & 44.992 & 0.150 & 0.806 & 0.101 & 2.666 & 0.164 & 2.022 & 0.334 \\
\midrule
\multirow{2}{*}{29 June} & HVAE & 36.814 & 0.147 & 1.000 & 0.207 & 1.969 & 0.292 & 2.953 & 0.511 \\
& \textbf{PILA} & 36.839 & 0.098 & 0.820 & 0.121 & 2.606 & 0.177 & 1.750 & 0.317 \\
\midrule
\multirow{2}{*}{\textbf{06 July}} & HVAE & 27.784 & 0.160 & 1.000 & 0.207 & 2.200 & 0.249 & 2.949 & 0.515 \\
& \textbf{PILA} & 34.354 & 0.097 & 0.888 & 0.121 & 2.718 & 0.159 & 2.069 & 0.366 \\
\midrule
\multirow{2}{*}{11 July} & HVAE & 37.852 & 0.163 & 1.000 & 0.207 & 1.905 & 0.308 & 2.948 & 0.516 \\
& \textbf{PILA} & 36.984 & 0.096 & 0.853 & 0.116 & 2.640 & 0.170 & 1.838 & 0.348 \\
\midrule
\multicolumn{2}{l}{\textbf{Ground Truth}} & 36.640 & N/A & 0.814 & N/A & 3.421 & N/A & 1.615 & N/A \\
\bottomrule
\end{tabular}
}
\label{tab:rtm_inversion_wytham_variables_insitu_extended_data}
\end{table}

\vspace{-5pt}
\FloatBarrier
\subsection{Extended results for Mogi inversion at Akutan Volcano in \cref{sec:results_mogi}}\label{appx:results_mogi}
\vspace{-5pt}

\begin{figure}[htbp]
    \centering
    \includegraphics[width=0.8\textwidth]{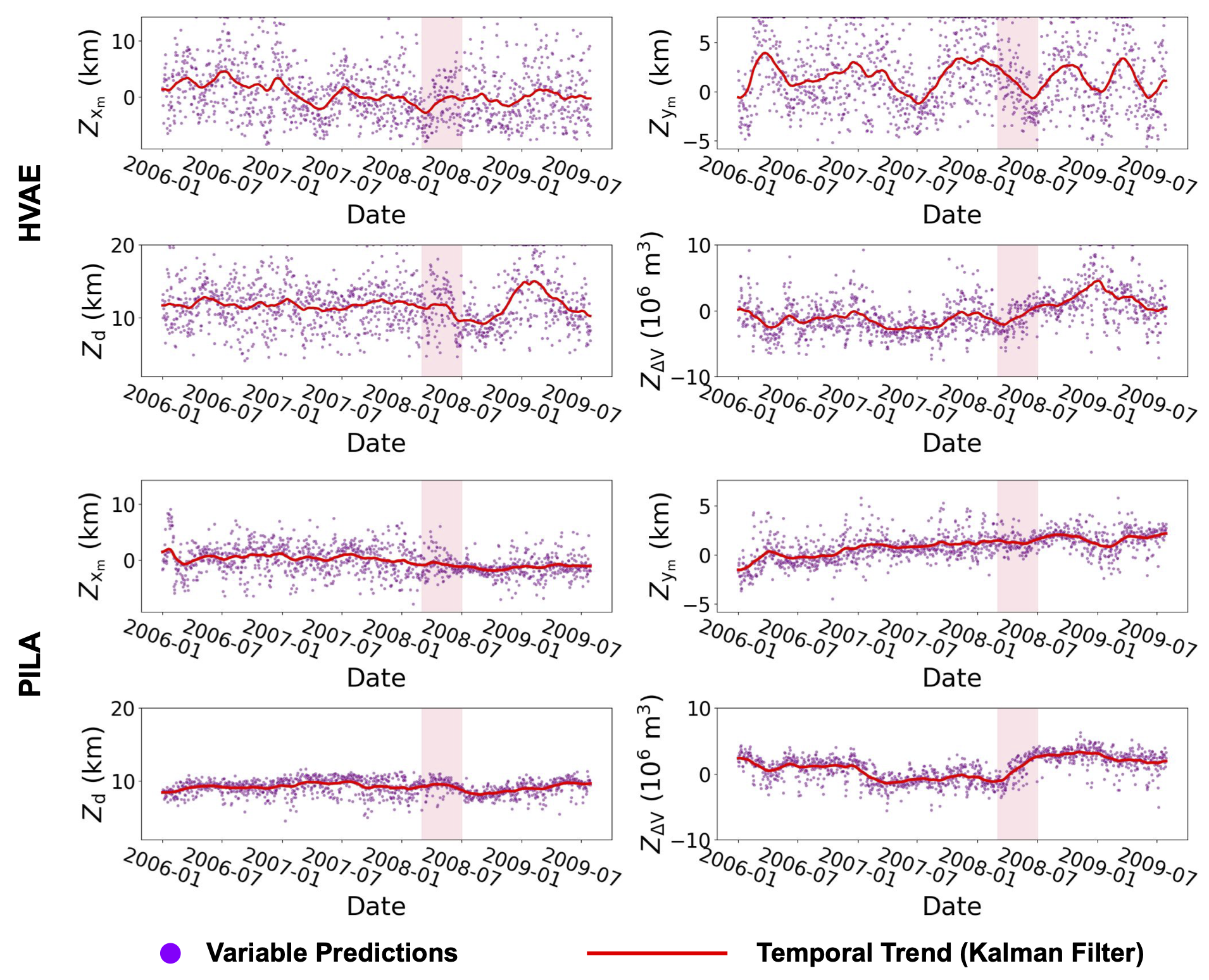}
    \caption{\textbf{Extended results complementing \cref{fig:mogi_inversion_variables} in the main text}, including raw variable predictions from both HVAE and PILA.}
    \label{fig:appx_mogi_inversion_variables}
\end{figure}

\begin{figure}[htbp]
    \centering
    \includegraphics[width=0.9\textwidth]{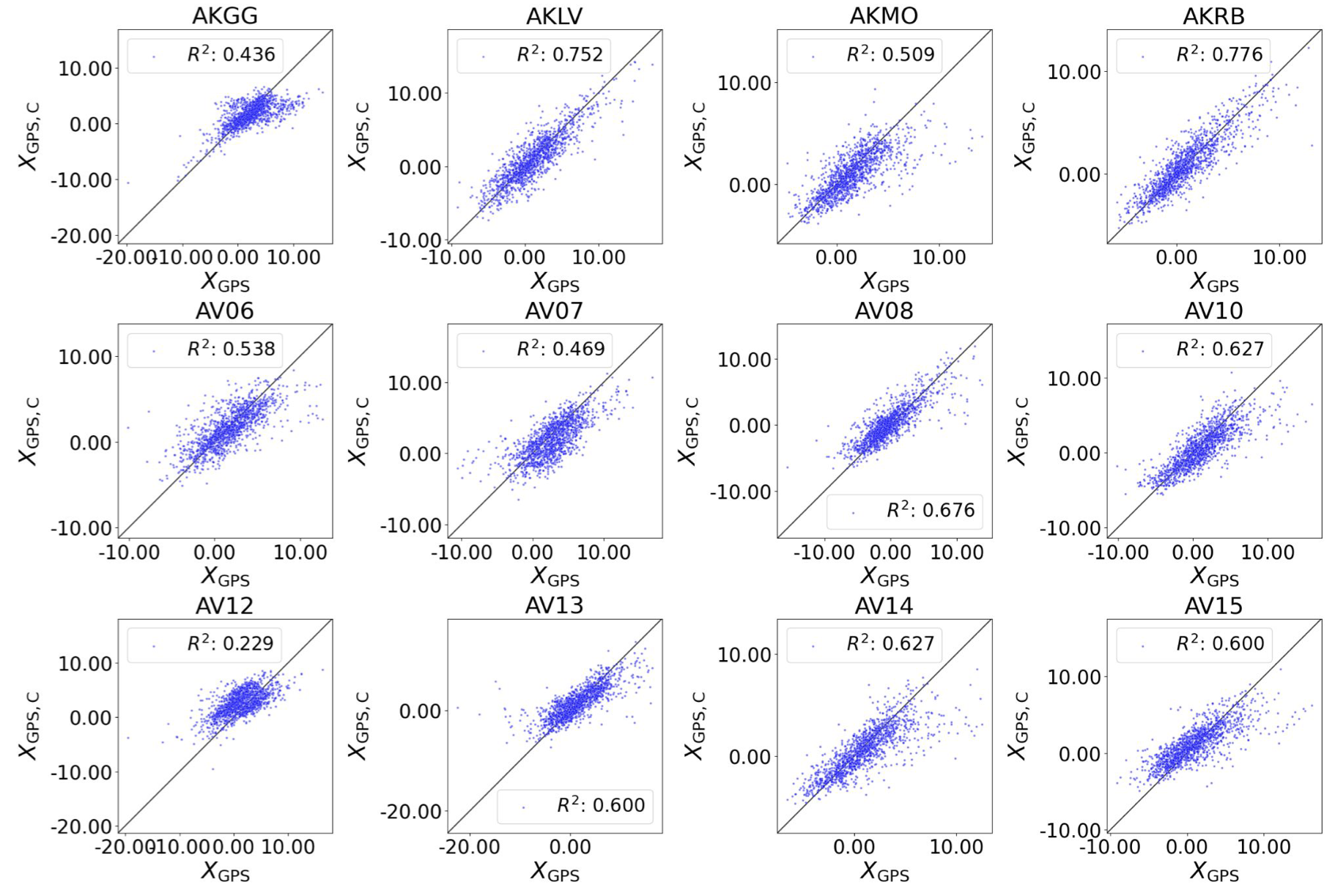}
    \caption{\textbf{Extended results complementing \cref{fig:mogi_inversion_gnss} in the main text}, including HVAE's reconstructions of GNSS signals for all stations.}
    \label{fig:appx_mogi_inversion_gnss_HVAE}
\end{figure}

\begin{figure}[htbp]
    \centering
    \includegraphics[width=0.9\textwidth]{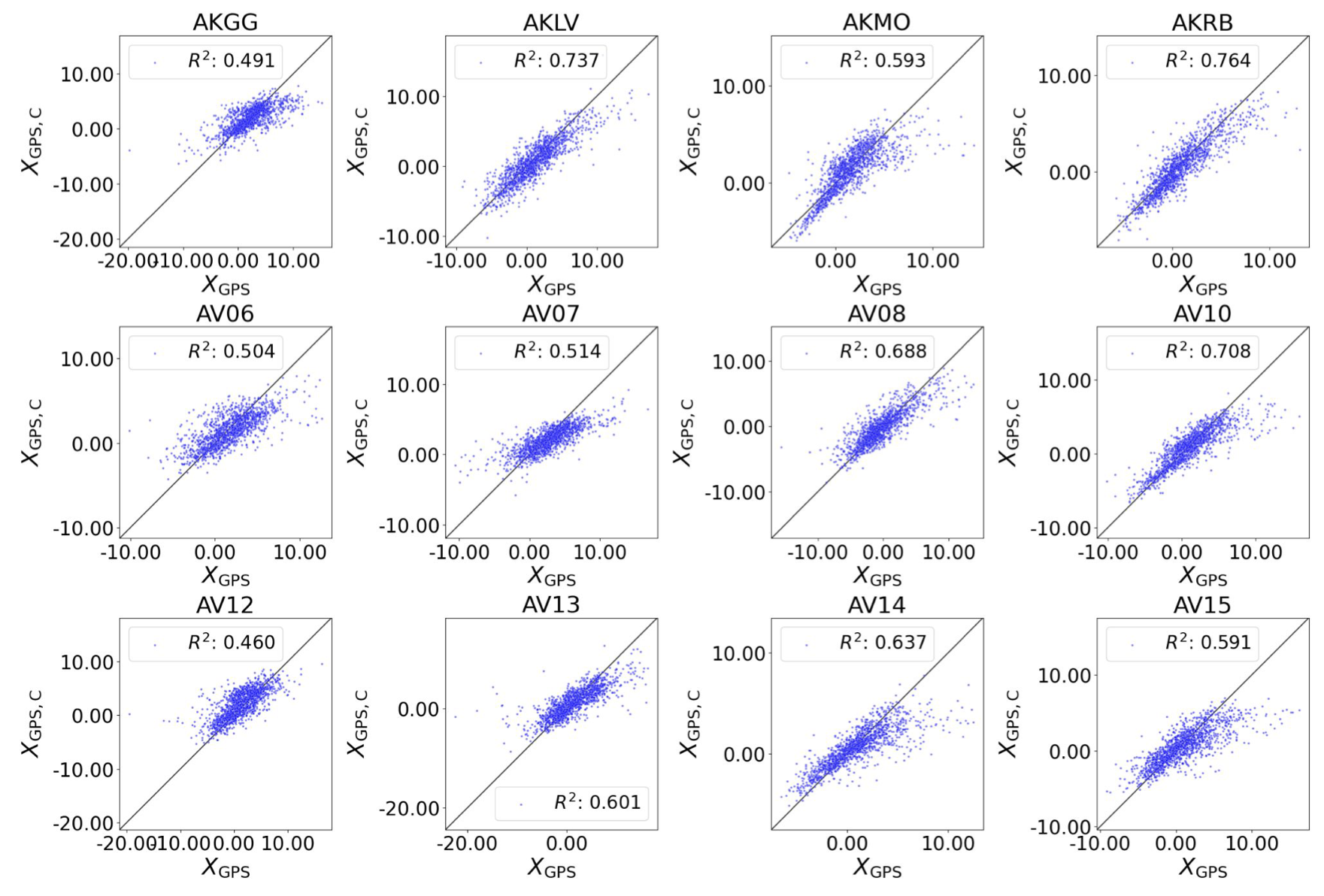}
    \caption{\textbf{Extended results complementing \cref{fig:mogi_inversion_gnss} in the main text}, including PILA's reconstructions of GNSS signals for all stations.}
    \label{fig:appx_mogi_inversion_gnss_PILA}
\end{figure}

\begin{figure}[htbp]
    \centering
    \includegraphics[width=\textwidth]{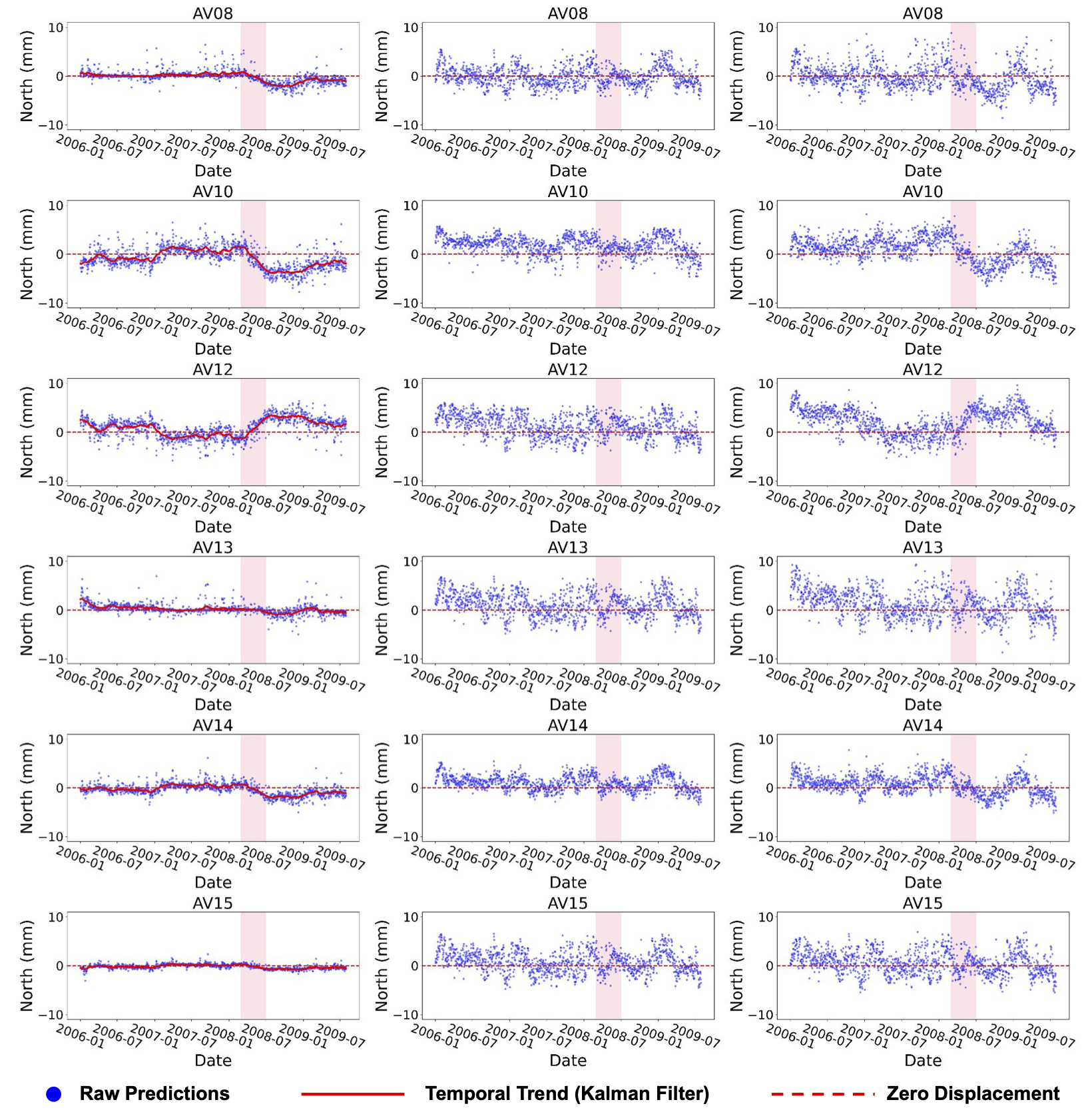}
    \caption{\textbf{Extended results complementing \cref{fig:mogi_inversion_gnss_transient_comparison_AV10} in the main text}, showing PILA's physical reconstruction (first column, $X_{\mathcal{F}}$), learned residual (second column, $\Delta$), and refined reconstruction (third column, $X_{\mathcal{C}}=X_{\mathcal{F}}+\Delta$) for six additional GNSS stations in the North direction around Akutan volcano.}
    \label{fig:appx_mogi_inversion_gnss_transient_v_residual_first_half}
\end{figure}

\begin{figure}[htbp]
    \centering
    \includegraphics[width=\textwidth]{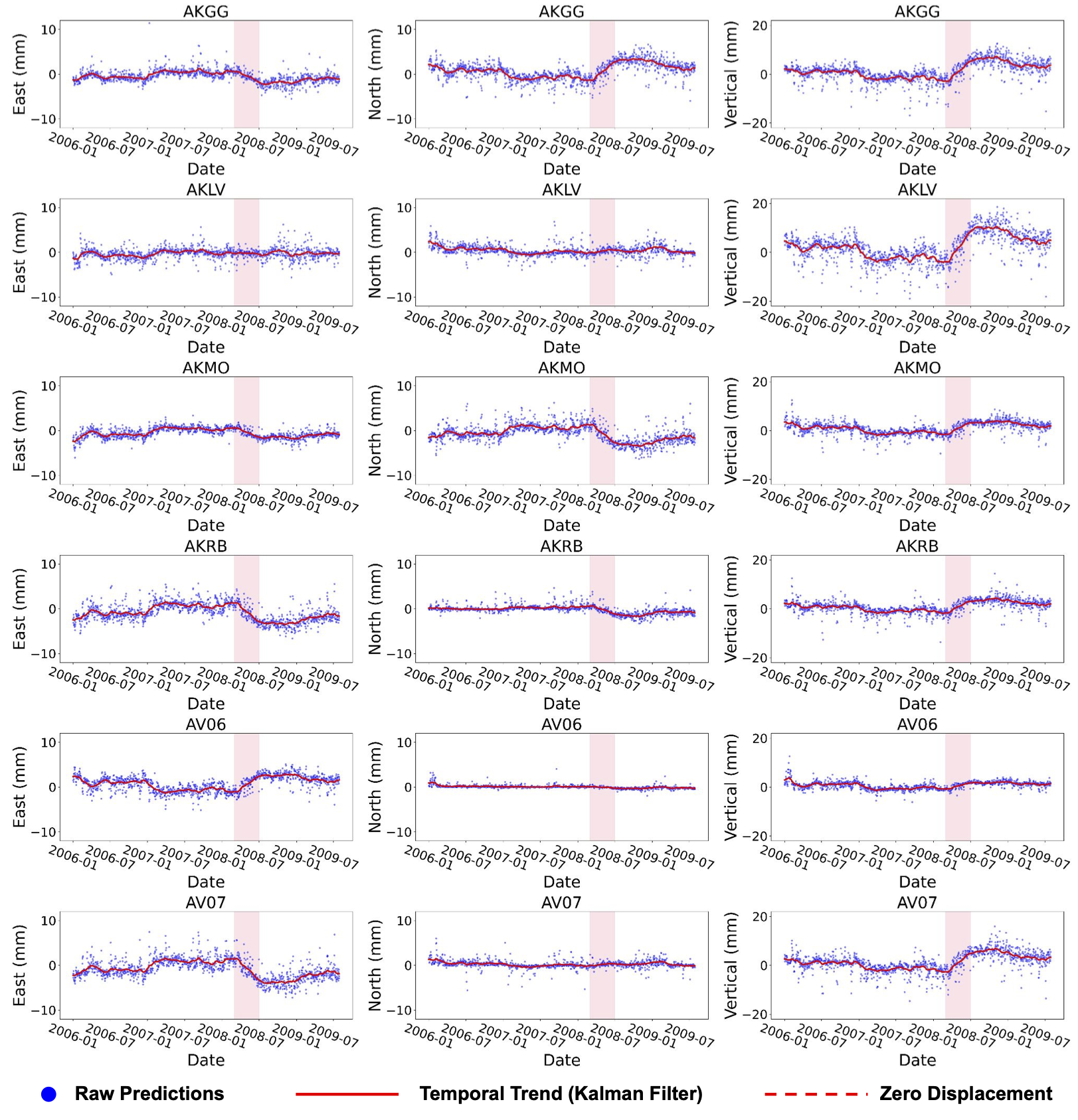}
  \caption{\textbf{Extended results complementing \cref{fig:mogi_inversion_gnss_radial_expansion_AV10_AV12} in the main text}, showing PILA's physical reconstruction for six additional GNSS stations around Akutan volcano along the three directions (East, North, Vertical). \textit{It reveals horizontal radial expansion and vertical uplift during the early-2008 inflation, followed by slow contraction.}}
\label{fig:appx_mogi_inversion_gnss_transient_all_dirs_first_half}
\end{figure}

\FloatBarrier
\subsection{Extended results for ablation studies in \cref{sec:ablations_residual_prior}}\label{appx:ablations_residual_prior}
\vspace{-5pt}
\begin{figure}[htbp]
    \centering
    \includegraphics[width=0.9\textwidth]{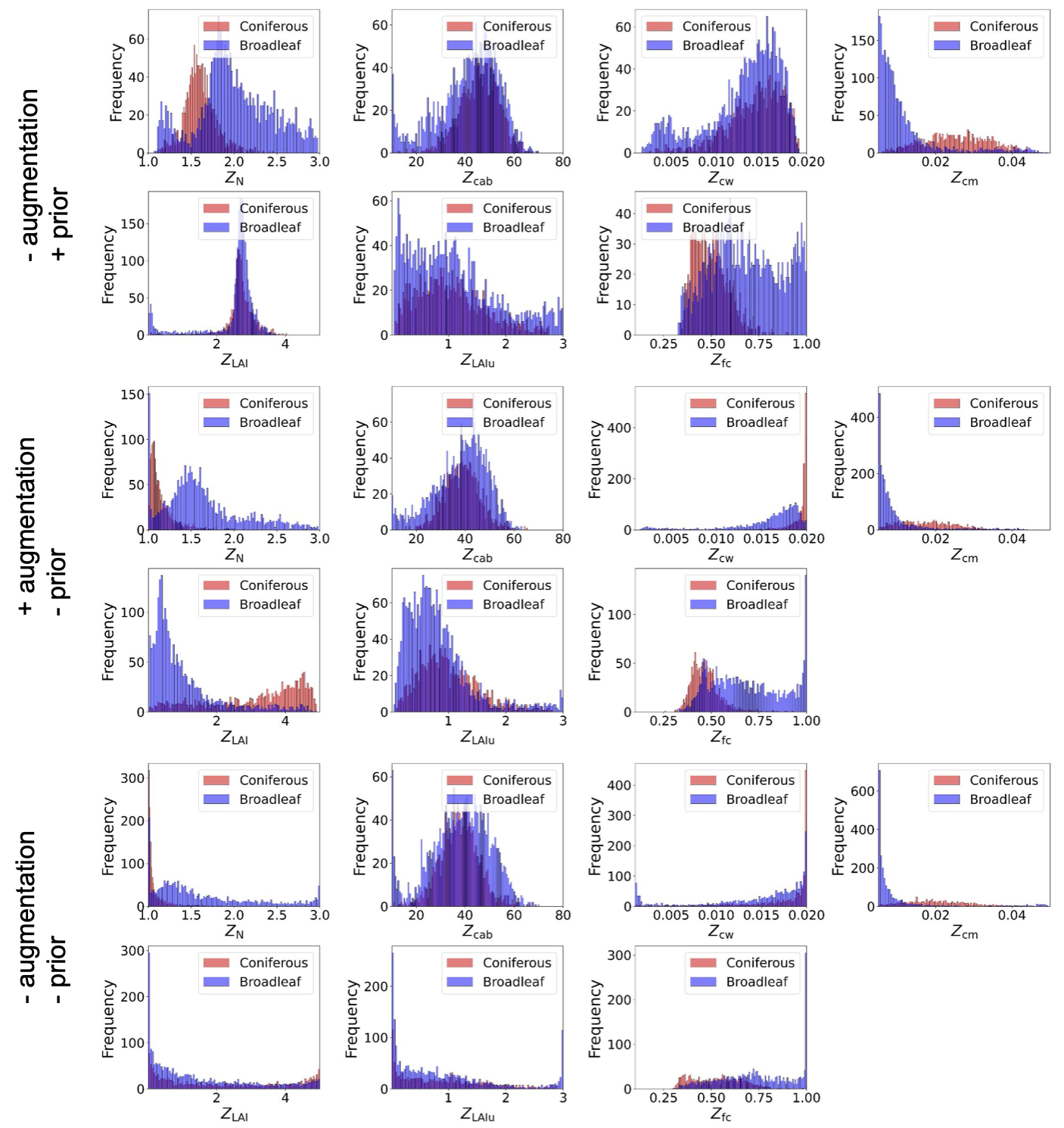}
    \caption{\textbf{Extended results complementing \cref{fig:rtm_inversion_austria_ablations}} in the main text, showing PILA ablations for all retrieved RTM inversion variables on the Austrian dataset.}
    \label{fig:appx_rtm_inversion_austria_ablations_vars}
\end{figure}
\vspace{-5pt}
\begin{figure}[htbp]
    \centering
    \includegraphics[width=0.75\textwidth]{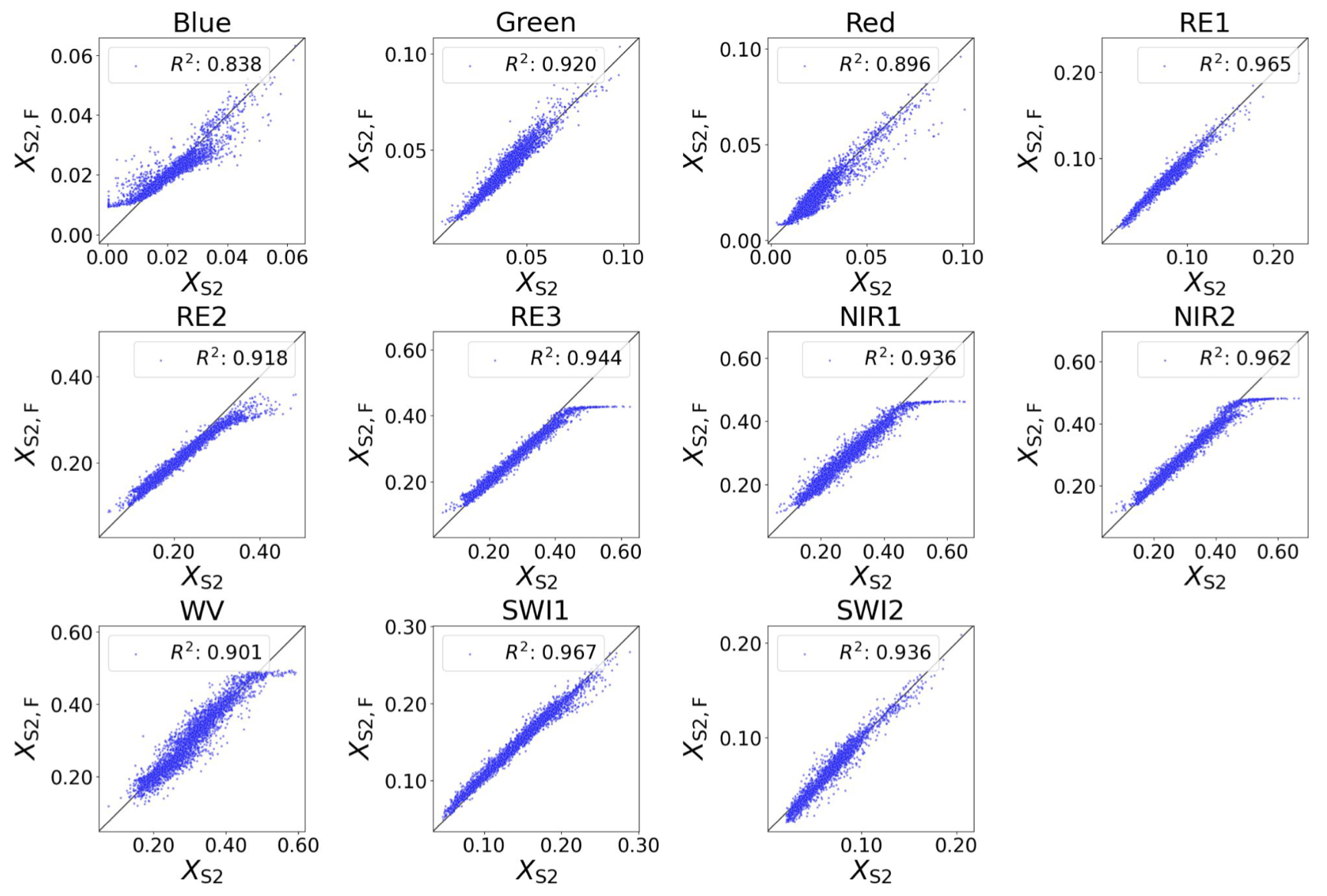}
    \caption{\textbf{With the RTM as decoder and no residual augmentation or prior, the physical reconstruction clearly deviates from the measured spectra}, overestimating visible bands at lower ranges and underestimating near-infrared bands at higher ranges.}
    \label{fig:appx_rtm_inversion_austria_ablations_spectra_no_adaptation_no_prior}
\end{figure}

\FloatBarrier
\subsection{Extended results for sensitivity analysis in \cref{sec:sensitivity}}\label{appx:sensitivity}
\begin{figure}[htbp]
    \centering
    \includegraphics[width=0.73\textwidth]{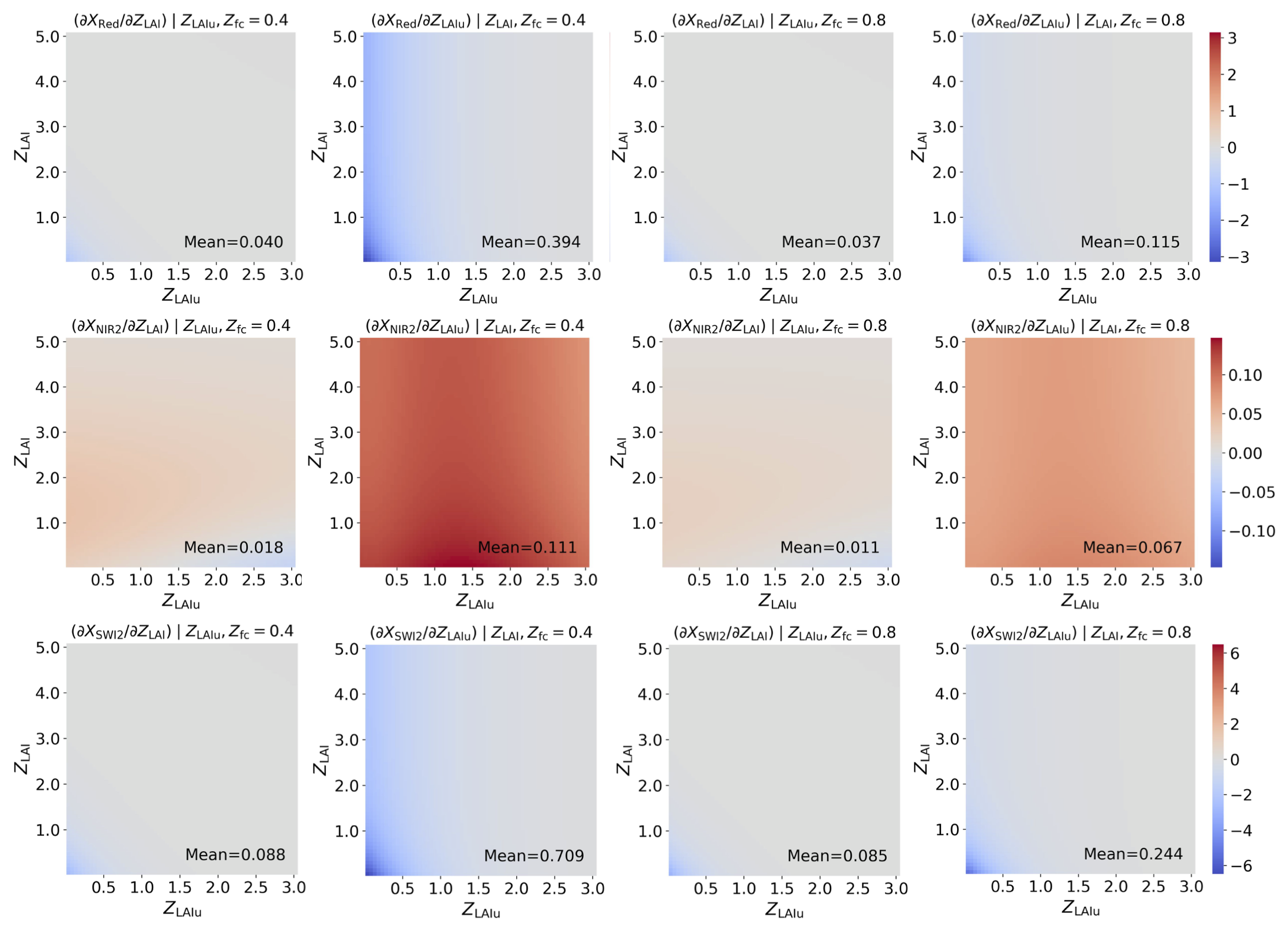}
    \caption{\textbf{Extended results complementing \cref{fig:rtm_inversion_variables_sensitivity_analysis_red_band} in the main text}, including sensitivities from NIR2 and SWI2 bands. \textit{Sensitivities to leaf area indices show consistent patterns across all bands, as discussed in \cref{sec:sensitivity}.}}
    \label{fig:appx_rtm_inversion_variables_sensitivity_analysis}
\end{figure}